\documentclass[11pt,a4paper]{article} 

\usepackage{hyperref}
\usepackage{url}

\usepackage{listings, xcolor}
\definecolor{codegreen}{rgb}{0,0.6,0}
\definecolor{codegray}{rgb}{0.5,0.5,0.5}
\definecolor{codepurple}{rgb}{0.58,0,0.82}
\definecolor{backcolour}{rgb}{0.95,0.95,0.92}

\lstdefinestyle{mystyle}{
    backgroundcolor=\color{backcolour},   
    commentstyle=\color{codegreen},
    keywordstyle=\color{magenta},
    numberstyle=\tiny\color{codegray},
    stringstyle=\color{codepurple},
    basicstyle=\ttfamily\footnotesize,
    breakatwhitespace=false,         
    breaklines=true,                 
    captionpos=b,                    
    keepspaces=true,                 
    numbers=left,                    
    numbersep=5pt,                  
    showspaces=false,                
    showstringspaces=false,
    showtabs=false,                  
    tabsize=2
}

\usepackage[utf8]{inputenc} 
\usepackage[T1]{fontenc}    
\usepackage{hyperref}       
\usepackage{url}            
\usepackage{booktabs}       
\usepackage{amsfonts}       
\usepackage{nicefrac}       
\usepackage{microtype}      
\usepackage{xcolor}         

\usepackage{cmap}
\usepackage[english]{babel}
\usepackage{amssymb,amsmath,amsthm}
\usepackage{amsthm}
\usepackage{bbm}
\usepackage{microtype}
\usepackage{xcolor}

\usepackage{multirow}
\usepackage{enumitem}
\setitemize{noitemsep,topsep=0pt,parsep=0pt,partopsep=0pt}

\usepackage{xspace}

\usepackage{enumitem}
\usepackage[olditem,oldenum]{paralist}
\usepackage{tikz}
\usetikzlibrary{shapes.geometric}
\usepackage{mathrsfs}
\makeatletter
\newcommand*\bigcdot{\mathpalette\bigcdot@{.6}}
\newcommand*\bigcdot@[2]{\mathbin{\vcenter{\hbox{\scalebox{#2}{$\m@th#1\bullet$}}}}}
\makeatother

\usepackage{xcolor, soul}
\definecolor{babyblueeyes}{rgb}{0.63, 0.79, 0.95}
\sethlcolor{babyblueeyes}

\usepackage{algorithm,algorithmic} 
\usepackage{subfigure}
\usepackage{booktabs}
\usepackage{hyperref}

\usepackage{graphicx}
\graphicspath{ {./paper_figs/} }

\newcommand{\augmentelargeur}[1]{
\addtolength{\evensidemargin}{-#1}
\addtolength{\oddsidemargin}{-#1}
\addtolength{\textwidth}{#1}
\addtolength{\textwidth}{#1}
}
\augmentelargeur{1cm}
\newcommand{\augmentehauteur}[1]{
\addtolength{\topmargin}{-#1}
\addtolength{\textheight}{#1}
\addtolength{\textheight}{#1}
}
\augmentehauteur{2.1cm}

\newcommand{\TODO}[1]{{\color{red} TODO: {#1}}}
\newcommand{\todo}[1]{\TODO{#1}}
\newcommand{\option}[1]{{\color[rgb]{.4,0,.8}[#1]}} 
\newcommand{\authorcomment}[2]{{\color[rgb]{#1}#2}}
\newcommand{\NDY}[1]{\authorcomment{0.0,0.8,0.4}{[NdY: #1]}}

\renewcommand{\TODO}[1]{}
\renewcommand{\todo}[1]{}
\renewcommand{\option}[1]{#1}  
\renewcommand{\authorcomment}[2]{}
\renewcommand{\NDY}[1]{}

\newtheorem{thm}{Theorem} 

\newtheorem*{notation*}{Notation}

\DeclareMathOperator*{\argmax}{arg\,max}
\DeclareMathOperator*{\argmin}{arg\,min}
\newcommand{\abs}[1]{\left\lvert#1\right\rvert}
\newcommand{\norm}[1]{\left\lVert#1\right\rVert}

\newcommand{\1}{\mathbbm{1}}  
\def\d{\operatorname{d}\!{}}  

\def\R{{\mathbb{R}}}

\def\D{{\mathcal{D}}}
\newcommand{\deq}{\mathrel{\mathop{:}}=}  
\newcommand{\from}{\colon} 
\def\eps{\varepsilon}
\renewcommand{\epsilon}{\varepsilon}
\renewcommand{\phi}{\varphi}
\let\oldPr\Pr
\renewcommand{\Pr}{\oldPr\nolimits}
\DeclareMathOperator{\E}{\mathbb{E}}  

\DeclareMathOperator{\Cov}{Cov}

\DeclareMathOperator{\diag}{diag}
\DeclareMathOperator{\Id}{Id}

\newcommand{\transp}[1]{#1^{\!\top}} 

\newcommand{\tsum}{{\textstyle \sum}}

\newcommand{\loss}{\mathcal{L}}

\hypersetup{
	plainpages=false,
	colorlinks=true,              
	linkcolor=blue,               
	anchorcolor=blue,             
	citecolor=blue,               
	filecolor=blue,               
	pagecolor=blue,               
	urlcolor=blue,                
	pdfview=FitH,                 
	pdfstartview=FitH,            
	pdfpagelayout=SinglePage      
}

\def\CL{\textbf{\texttt{CL}}\xspace}
\def\Rand{\textbf{\texttt{Rand}}\xspace}
\def\FB{\textbf{\texttt{FB}}\xspace}
\def\SF{\textbf{\texttt{SF}}\xspace}
\def\LRAP{\textbf{\texttt{LRA-P}}\xspace}
\def\LRASR{\textbf{\texttt{LRA-SR}}\xspace}
\def\ICM{\textbf{\texttt{ICM}}\xspace}
\def\lap{\textbf{\texttt{Lap}}\xspace}
\def\Latent{\textbf{\texttt{Latent}}\xspace}
\def\Transition{\textbf{\texttt{Trans}}\xspace}
\def\APS{\textbf{\texttt{APS}}\xspace}
\def\AEnc{\textbf{\texttt{AEnc}}\xspace}

\title{Does Zero-Shot Reinforcement Learning Exist?}

\author{Ahmed Touati, Jérémy Rapin, Yann Ollivier \footnote{Meta AI
Research, Paris}}
\date{
\today
}

\begin{document}

\maketitle

\begin{abstract}
\NDY{
Titles:\\
Does Zero-Shot RL Exist?\\
Towards Zero-Shot Reinforcement Learning\\
Getting Closer to Zero-Shot Reinforcement Learning\\
Towards Full Zero-Shot Reinforcement Learning\\
Rethinking Zero-Shot Reinforcement Learning\\
Zero-Shot RL: Towards Controllable Agents\\
Zero-Shot Reinforcement Learning?\\
Successor Representations for Offline Unsupervised Reinforcement
Learning\\
Do Successor Representations Provide Universal RL Agents?\\
Do Successor Representations Solve Zero-Shot RL?  \\
From RL to zero-shot RL\\
RL with No Rewards and No Planning\\
Zero-Shot RL Revisited\\
Towards Controllable Agents\\
The Zero-Shot RL Revolution\\
}

A \emph{zero-shot RL agent} is an agent that can solve \emph{any} RL task in a
given environment, instantly with no additional planning or learning, after an
initial reward-free learning phase.
%
This marks a shift from
the reward-centric RL paradigm towards
``controllable'' agents that can follow arbitrary instructions in an
environment. Current RL agents can solve families of related tasks
at best, or require planning anew for each task.
Strategies for approximate zero-shot RL have been suggested using
successor features (SFs) \cite{borsa2018universal} or forward-backward
(FB) representations
\cite{touati2021learning}, but testing has been
limited. 

After clarifying the relationships between these schemes, we 
introduce improved losses and new SF models,
and test the viability of zero-shot RL schemes
systematically on tasks
from the Unsupervised RL benchmark
\cite{laskin2021urlb}.
To disentangle universal representation learning from exploration, we
work in an offline setting and repeat the tests 
on several existing replay buffers.

SFs appear to suffer from the choice of the
elementary state features. SFs with Laplacian eigenfunctions do well,
while SFs based on auto-encoders,
inverse curiosity, transition models, low-rank
transition matrix,
contrastive learning, or diversity (APS), 
perform unconsistently.
In contrast, FB representations jointly
learn the elementary and successor features from a single, principled
criterion. They
perform best and consistently across the board, reaching $85\%$ of
supervised RL performance with a good replay buffer, in a zero-shot
manner.
\end{abstract}

\section{Introduction}
\label{sec:intro}

For breadth of applications, reinforcement learning (RL) lags behind
other fields of machine learning, such as vision or natural language
processing, which have effectively adapted to a wide range of tasks,
often in almost zero-shot manner, using pretraining on large, unlabelled
datasets~\cite{brown2020language}. The RL paradigm itself may be in part to blame:
RL agents are usually trained for only one reward function
or a small family of related rewards. Instead, we would like to train
\option{``controllable''} agents that can be given a description of any
task (reward function) in their environment, and then immediately know what to
do, reacting instantly
to such commands as “fetch this object while
avoiding that area”.

The promise of \emph{zero-shot RL} is to train without rewards or tasks, yet
immediately perform well on any reward function given at test time, with
no extra training, planning, or finetuning, and only a minimal amount of extra computation to
process a task description (Section~\ref{sec:notation}
gives the precise definition we use for \emph{zero-shot RL}). 
%
How far away are such zero-shot agents?
In the RL paradigm,
a new task (reward
function) means re-training the agent from scratch, and providing many
reward samples. Model-based RL trains a reward-free, task-independent world model, but
still requires heavy planning when a new reward function is specified
(e.g., \cite{chua2018deep,moerland2020model}).
Model-free
RL is reward-centric from start, and produces specialized agents.
Multi-task agents generalize within a family of related tasks only.
Reward-free, unsupervised skill pre-training (e.g., 
\cite{eysenbach2018diversity}) still requires substantial
downstream task adaptation, such as training a hierarchical
controller.

Is zero-shot RL possible?
If one ignores
practicality, zero-shot RL is easy: make a list of all possible
rewards up to precision $\eps$, then pre-learn all the associated optimal
policies.
Scalable zero-shot RL must somehow exploit
the relationships between policies for all tasks. 
Learning to go from $a$ to $c$ is not independent from going from $a$ to
$b$ and $b$ to $c$, and this produces rich, exploitable algebraic
relationships \cite{successorstates,pmlr-v37-schaul15}.

Suggested strategies for \option{generic} zero-shot RL so far have used 
successor representations \cite{dayan1993improving}, under
two forms: \emph{successor features} (SFs)
\cite{barreto2017successor} as in
\cite{borsa2018universal,hansen2019fast,liu2021aps}; and
\emph{forward-backward} (FB) representations \cite{touati2021learning}.
Both SFs and FB lie in between model-free and model-based RL,
by predicting features of future states, or summarizing long-term
state-state relationships.
Like model-based approaches, they decouple the dynamics of the
environment from the
reward function.
Contrary to world models, they require neither planning at
test time nor a generative model of states or trajectories. 

Yet SFs heavily depend on a choice of basic state features. 
To get a full
zero-shot RL algorithm, a representation
learning method must provide those. While SFs have been successively applied to transfer
between tasks, most of the time, the basic features were handcrafted or
learned using prior task class knowledge. Meanwhile, FB is a
standalone method with no task prior and good theoretical backing, but testing has been
limited to goal-reaching in a few environments%
.
Here:


\begin{itemize}[noitemsep]
\item We systematically assess SFs and FB for
zero-shot RL, including many new models of SF basic features\option{, and
improved FB loss functions}. We
use 13 tasks from the Unsupervised RL benchmark \cite{laskin2021urlb},
repeated on
several ExORL training replay buffers \cite{yarats2021reinforcement} to
assess robustness to the exploration method.
\item We systematically study the influence of basic features for
SFs, by testing
SFs on features from ten RL representation learning methods.\option{ such
as latent next state prediction, inverse curiosity module,
contrastive learning, diversity, various spectral decompositions...}
\item We expose new mathematical links between SFs, FB, and
other representations in RL.\option{
\item We discuss the implicit assumptions and limitations behind zero-shot
RL approaches.}
\end{itemize}


\begin{figure}
\begin{center}
\includegraphics[width=.59\textwidth]{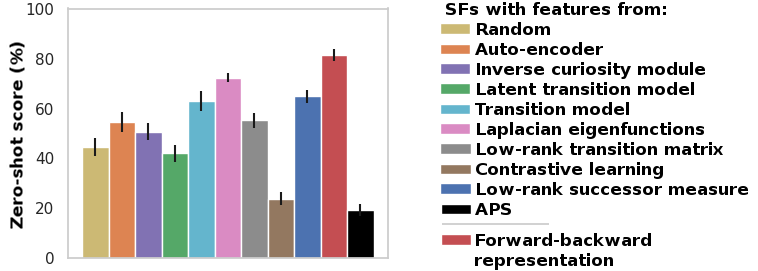}
\end{center}
\caption{Zero-shot scores of ten SF methods and FB,
as a percentage of the
supervised score of offline TD3
trained on the same replay buffer,
averaged on some tasks, environments and replay buffers from the Unsupervised RL
and ExORL benchmarks \cite{laskin2022cic,yarats2021reinforcement}.
FB and SFs with Laplacian eigenfunctions achieve zero-shot
scores 
approaching supervised RL.
\todo{?use error bars corresponding to stddev of score distribution}
\label{fig:global}
}
\end{figure}

\section{Problem and Notation; Defining Zero-Shot RL}
\label{sec:notation}

Let $\mathcal{M}=(S,A,P,\gamma)$ be
a reward-free Markov decision process (MDP)
with state space $S$, action space $A$, transition probabilities
$P(s'|s,a)$ from state $s$ to $s'$ given action $a$,
and discount factor $0 < \gamma < 1$ \cite{sutton2018reinforcement}.  If
$S$ and $A$ are finite, $P(s'|s,a)$ can be viewed as a stochastic matrix
$P_{sas'} \in \R^{(\abs{S}\times \abs{A})\times \abs{S}}$; in general,
for each $(s,a)\in S\times A$, $P(\d s'|s,a)$ is a probability measure on
$s'\in S$.  The notation $P(\d s'|s,a)$ covers all cases.  Given
$(s_0,a_0)\in S\times A$ and a policy $\pi\from S\to \mathrm{Prob}(A)$,
we denote $\Pr(\cdot|s_0,a_0,\pi)$ and $\E[\cdot|s_0,a_0,\pi]$ the
probabilities and expectations under state-action sequences $(s_t,a_t)_{t
\geq 0}$ starting at $(s_0,a_0)$ and following policy $\pi$ in the
environment, defined by sampling $s_t\sim P(\d s_t|s_{t-1},a_{t-1})$ and
$a_t\sim \pi(\d a_t | s_t)$. We define $P_\pi(\d s', \d a' | s, a)
\deq
P(\d s'| s, a) \pi(\d a' | s')$ and $P_\pi(\d s' | s) \deq \int P(\d s' | s, a)
\pi(\d a | s)$, the state-action transition probabilities and state
transition probabilities induced by $\pi$.  Given a reward
function $r\from S \to \R$, the $Q$-function of $\pi$ for $r$ is
$Q_r^\pi(s_0,a_0)\deq \sum_{t\geq 0} \gamma^t \E
[r(s_{t+1})|s_0,a_0,\pi]$. For simplicity, we assume the
reward $r$ depends only
on the next state $s_{t+1}$ instead  on the full triplet $(s_t, a_t,
s_{t+1})$, but this is
not essential.

We focus on offline unsupervised RL, where the
agent cannot interact with the environment. The agent only has
access to a static dataset of logged reward-free transitions
in the environment,
$\D = \{(s_i, a_i, s'_i)\}_{i\in \mathcal{I}}$ with $s'_i\sim P(\d
s'_i|s_i,a_i)$. These can come from any exploration
method or methods.

The offline setting disentangles the effects of the exploration method
and representation and policy learning: we test each zero-shot method
on several training datasets 
from several exploration methods.

We denote by $\rho(\d s)$ and $\rho(\d s, \d a)$ the (unknown) marginal
distribution of states and state-actions in the dataset $\D$. We
use both $\E_{s \sim \D}[\cdot ]$ and $\E_{s \sim \rho}[
\cdot]$ for expectations under the training distribution. 

\paragraph{Zero-shot RL: problem statement.}
The goal of zero-shot RL is to compute a compact representation
$\mathcal{E}$ of the
environment by observing samples of reward-free transitions $(s_t,a_t,s_{t+1})$ in
this environment. Once a reward function is
specified later, the agent must use $\mathcal{E}$ to immediately produce a good policy, via
only
elementary computations without
any further planning or learning.  Ideally, for any downstream task, the
performance of the returned policy should be close to the performance of
a supervised RL baseline trained on the same dataset labeled with the
rewards for that task.

Reward functions may be specified at test time either as 
a relatively small set of reward samples $(s_i,r_i)$,
or as
an explicit
function $s\mapsto r(s)$ (such as $1$ at a known goal state and $0$ elsewhere).
The method will be few-shot, zero-planning in the first case, and
truly zero-shot in the second case.

\section{Related work}

Zero-shot RL requires unsupervised learning and the 
absence of any planning or fine-tuning at test time.
The proposed strategies for zero-shot RL discussed in
Section~\ref{sec:intro} ultimately derive from successor representations
\cite{dayan1993improving} in finite spaces. In continuous spaces,
starting with a finite number of features $\phi$, successor features can
be used to produce policies within a family of tasks directly related to $\phi$
\cite{barreto2017successor,borsa2018universal,zhang2017deep,grimm2019disentangled},
often using hand-crafted $\phi$ or learning $\phi$ that best linearize training rewards. 
VISR \cite{hansen2019fast} and its
successor APS \cite{liu2021aps} use SFs with
$\phi$ automatically built online via diversity criteria
\cite{eysenbach2018diversity,gregor2016variational}.
We include APS among our baselines, as well as many new criteria to build
$\phi$ automatically.

Successor measures \cite{successorstates} avoid the need for $\phi$ by
directly learning models of the distribution of future states: doing this
for various policies yields a candidate zero-shot RL method,
forward-backward representations \cite{touati2021learning}, which has
been tested for goal-reaching in a few environments with discrete actions. FB uses a
low-rank model of long-term state-state relationships reminiscent of the
state-goal factorization from \cite{pmlr-v37-schaul15}.

Model-based RL (surveyed in \cite{moerland2020model}) misses the zero-planning requirement of
zero-shot RL. Still, learned models of the transitions between states can be
used jointly with SFs to provide zero-shot methods (\Transition, \Latent, and \LRAP methods below).

Goal-oriented and multitask RL has a long history
(e.g., \cite{foster2002structure, sutton2011horde,da2012learning,
pmlr-v37-schaul15,andrychowicz2017hindsight}). A parametric family of
tasks must be defined in advance (e.g., reaching arbitrary goal states).
New rewards cannot be set a posteriori: for example, a
goal-state-oriented method cannot handle dense rewards. Zero-shot task transfer
methods learn on tasks and can transfer to related
tasks only (e.g., \cite{oh2017zero,sohn2018hierarchical}); this can be
used, e.g., for sim-to-real transfer \cite{genc2020zero} or slight
environment changes, which is not covered here. Instead, we
aim at not having any predefined family of
tasks.

Unsupervised skill and option discovery methods, based for instance on diversity
\cite{eysenbach2018diversity,gregor2016variational} or eigenoptions
\cite{machado2017laplacian} can learn a variety of behaviors without
rewards. Downstream tasks require
learning a hierarchical controller to
combine the right skills or options for each task. Directly using  unmodified
skills has limited performance without heavy finetuning
\cite{eysenbach2018diversity}. Still, these methods 
can speed up downstream learning.

The unsupervised aspect of some of these methods (including DIAYN and APS) has been
disputed, because training still used end-of-trajectory signals, which are 
directly correlated to the downstream task in some common environments: without this signal, results drop
sharply \cite{laskin2022cic}.

\section{Successor Representations and Zero-Shot RL}
\label{sec:SR}


For a finite MDP, the
\emph{successor representation} (SR) \cite{dayan1993improving}
$M^\pi(s_0, a_0)$ of a state-action pair $(s_0,a_0)$ under a policy
$\pi$, is defined as the  discounted sum of future occurrences of each state: 
 \begin{equation}
\label{eq:defSR}
M^\pi(s_0,a_0, s)\deq \E \left[\tsum_{t\geq 0}\, \gamma^t \1_{\{ s_{t+1} = s\}} \mid
(s_0,\,a_0),\,\pi
\right] \quad \forall s \in S.
\end{equation}
In matrix form, SRs can be written as $M^\pi = P \sum_{t \geq 0}
\gamma^{t}P_\pi^t = P (\Id - \gamma P_\pi)^{-1}$, where $P_\pi$ is
the state transition probability. $M^\pi$ satisfies the matrix Bellman
equation $M^\pi=P+\gamma P_\pi M^\pi$.

Importantly, SRs disentangle the
dynamics of the MDP and the reward function: for any reward $r$ and
policy $\pi$, the $Q$-function can be expressed linearly as $Q^\pi_r =
M^\pi r$.

\paragraph{Successor features and successor measures.}
\emph{Successor
features} (SFs) \cite{barreto2017successor} extend SR to continous MDPs
by first assuming we are given a
\emph{basic feature} map $\phi\from S\to \R^d$ that embeds states into $d$-dimensional
space, and defining the expected discounted sum of future state
features:
\begin{equation}
\label{eq:defSF}
\psi^\pi(s_0,a_0)\deq \E \left[ \tsum_{t\geq 0}\, \gamma^t \phi(s_{t+1}) \mid
s_0,\,a_0,\,\pi \right].
\end{equation}
SFs have been introduced to make SRs compatible
 with function approximation. For a finite MDP, the original definition
 \eqref{eq:defSR} is recovered by letting $\phi$ be a one-hot state
 encoding into $\R^{\abs{S}}$.

Alternatively, 
\emph{successor measures} (SMs)
\cite{successorstates} extend SRs to continuous spaces by treating
the distribution of future visited states as a measure $M^\pi$ over the state space
$S$,
\begin{equation}
\label{eq:defM}
M^\pi(s_0,a_0,X)\deq \tsum_{t\geq 0}\, \gamma^t \Pr\left( s_{t+1} \in X \mid
s_0,\,a_0,\,\pi
\right) \quad \forall X\subset S.
\end{equation} 
SFs and SMs are related: by construction, $\psi^\pi(s_0,a_0)=\int_{s'}
M^\pi(s_0,a_0,\d
s')\,\phi(s')$.

\paragraph{Zero-shot RL from successor features and forward-backward
representations.} 
Successor representations provide a generic framework for zero-shot RL, by learning to represent the relationship
between
reward functions and $Q$-functions, as encoded in $M^\pi$.

Given a basic feature map $\phi\from  S \to \R^d$ to be learned via another
criterion, \emph{universal SFs}~\cite{borsa2018universal} learn the
successor features of a particular family of policies $\pi_z$ for $z\in
\R^d$,
\begin{equation}
\label{eq:defUSF}
\psi(s_0, a_0, z) = \E \left[ \tsum_{t\geq 0 }\, \gamma^t \phi(s_{t+1}) \mid
(s_0, a_0), \pi_z \right], \quad \pi_z(s) \deq \argmax\nolimits_{a} \transp{\psi(s, a,
z)} z.
\end{equation} 
Once a reward function $r$ is revealed, we use a few reward samples or
explicit knowledge of the function $r$ to perform a linear regression of
$r$ onto the features $\phi$. Namely, we
estimate $z_r \deq  \argmin_z \E_{s
\sim \rho}[ (r(s)-\phi(s)^\top z )^2 ] = \E_\rho [\phi \phi^\top]^{-1}
\E_\rho[\phi r]$.
Then we return the policy $\pi_{z_r}$.
This policy is guaranteed to be optimal for all rewards in the linear span of the
features $\phi$:

\begin{thm}[\cite{borsa2018universal}]
\label{thm:SF}
Assume that \eqref{eq:defUSF} holds.
Assume there exists a weight $w \in \R^d$ such that $r(s) =
\transp{\phi(s)}
w, \forall s \in S$. Then $z_r = w$, and $\pi_{z_r}$ is the optimal policy
for reward $r$.
\end{thm}

\emph{Forward-backward (FB) representations} \cite{touati2021learning} apply
a similar idea to a finite-rank model of successor measures.
They look for representations $F\from S\times A\times \R^d\to
\R^d$ and $B\from S\to \R^d$ such that the long-term transition
probabilities $M^{\pi_z}$ in \eqref{eq:defM} decompose as
\begin{equation}
\label{eq:defFB}
M^{\pi_z}(s_0,a_0,\d s')\approx \transp{F(s_0,a_0,z)}B(s') \,\rho(\d s'),
\quad \pi_z(s) \deq \argmax\nolimits_{a} \transp{F(s, a,
z)} z
\end{equation}
In a finite space, the first equation rewrites as the matrix
decomposition
$M^{\pi_z}=\transp{F_z}B\diag(\rho)$.

Once a reward function $r$ is revealed, we estimate $z_r \deq  \E_{s\sim
\rho} [r(s)B(s)]$
from a few reward samples or from
explicit knowledge of the function $r$ (e.g.\ $z_r=B(s)$ to reach $s$). Then we return the policy $\pi_{z_r}$.
If the approximation \eqref{eq:defFB} holds, 
this policy is guaranteed to be optimal for any reward function:

\begin{thm}[\cite{touati2021learning}]
\label{thm:FB}
Assume that \eqref{eq:defFB} holds. Then for any reward function $r$, the
policy $\pi_{z_r}$ is optimal for $r$, with optimal $Q$-function
$Q^\star_r=\transp{F(s,a,z_r)}z_r$.
\end{thm}

For completeness, we sketch the proofs of Theorems~\ref{thm:SF}--\ref{thm:FB} in
Appendix~\ref{app:proofs}.
Importantly, both theorems are
compatible with approximation: approximate solutions provide
approximately optimal policies.

\paragraph{Connections between SFs and FB.} A first difference between SFs
and FB is that SFs must be provided with basic
features $\phi$. The best $\phi$ is such
that the reward functions of the downstream tasks are linear in $\phi$.
But for unsupervised training without prior task knowledge, 
an external criterion is needed to learn $\phi$.
We test a series of such criteria
below. In contrast, FB uses a single criterion, avoiding the need for state featurization by
learning a model of state occupancy.


Second, SFs only cover rewards in the linear span of
$\phi$, while FB apparently covers any reward. But this difference
is not as stark as it looks: exactly solving the FB equation \eqref{eq:defFB} 
in continuous spaces requires $d=\infty$, and for finite $d$, the policies will
only be optimal for rewards in the linear span of $B$
\cite{touati2021learning}. Thus, in both cases, policies are exactly
optimal only for a $d$-dimensional family of rewards. Still, FB can use
an arbitrary large $d$ without any additional input or criterion.

FB representations are related to successor features: the FB
definition \eqref{eq:defFB} implies that $\psi(s,a,z)\deq F(s,a,z)$ are
the successor features of $\phi(s)\deq
(\E_\rho B\transp{B})^{-1} B(s)$. This follows from 
multiplying \eqref{eq:defM} and \eqref{eq:defFB} by
$\transp{B}(\E_\rho
B\transp{B})^{-1}$ on the right, and integrating over $s'\sim \rho$.
Thus, a posteriori, FB can be used to produce
both $\phi$ and $\psi$ in SF, although training is different.

This connection between FB and SF is one-directional: \eqref{eq:defFB} is a
stronger condition. In particular $F=B=0$ is not a solution: contrary to
$\psi=\phi=0$ in \eqref{eq:defUSF}, there
is no collapse. No additional criterion to train $\phi$ is required:
$F$ and $B$ are trained jointly to provide the best rank-$d$
approximation of the successor measures $M^\pi$. This summarizes an
environment by selecting the features that
best describe the relationship $Q^\pi_r=M^\pi r$ between rewards and $Q$-functions.

\section{Algorithms for Successor Features and FB Representations}

We now describe more precisely the algorithms used in our experiments.
The losses
used to train $\psi$ in SFs, and $F,B$ in FB, are described
in Sections~\ref{sec:psiloss} and \ref{sec:FBloss} respectively.

To obtain a full zero-shot RL algorithm, SFs must specify the basic
features $\phi$. Any representation learning method can be used for
$\phi$. We
use ten possible choices
(Section~\ref{sec:philoss}) based on existing or new representations for
RL:
random features as a baseline, autoencoders, next state and latent next
state transition models, inverse curiosity module, the diversity
criterion of APS, contrastive learning, and finally,
several spectral decompositions of the transition
matrix or its associated Laplacian.

Both SFs and FB define policies as an argmax of
$\transp{\psi(s,a,z)}z$ or $\transp{F(s,a,z)}z$ over actions $a$. With
continuous actions,
the argmax cannot be computed exactly. We train an
auxiliary policy network $\pi(s,z)$ to approximate this argmax, using the same standard method for SFs and FB (Appendix~\ref{sec:piloss}).


\subsection{Learning the Successor Features $\psi$}
\label{sec:psiloss}

The successor features $\psi$ satisfy the $\R^d$-valued Bellman
equation $\psi^\pi=P\phi+\gamma P_\pi \psi^\pi$, the collection of
ordinary Bellman equations for each component of $\phi$. The $P$ in
front of $\phi$ comes from using $\phi(s_{t+1})$ not $\phi(s_t)$ in
\eqref{eq:defSF}. Therefore, we can train $\psi(s,a,z)$ for each $z$ by
minimizing the Bellman residuals
$\norm{\psi(s_t,a_t,z)-\phi(s_{t+1})-\gamma \bar
\psi(s_{t+1},\pi_z(s_{t+1}),z)}^2$ where $\bar \psi$ is a non-trainable
target version of $\psi$ as in parametric $Q$-learning. This requires
sampling a transition $(s_t,a_t,s_{t+1})$ from the dataset and choosing
$z$. 
We sample random values of $z$ as described in
Appendix~\ref{sec:zsampling}.

This is the loss used in \cite{borsa2018universal}. But
this can be improved, since we do not use the full vector
$\psi(s,a,z)$: only $\transp{\psi(s,a,z)}z$ is needed for the policies.
Therefore, as in \cite{liu2021aps},
instead of the vector-valued Bellman residual above, we just
use
\begin{equation}
\loss(\psi)\deq \E_{(s_t,a_t,s_{t+1})\sim
\rho}\left(\transp{\psi(s_t,a_t,z)}z-\transp{\phi(s_{t+1})}z-\gamma 
\transp{\bar\psi(s_{t+1},\pi_z(s_{t+1}),z)}z\right)^2
\end{equation}
for each $z$. This trains $\transp{\psi(\cdot,z)}z$ as the $Q$-function
of reward
$\transp{\phi}z$, the only case needed, while
training the full vector $\psi(\cdot,z)$ amounts to
training the $Q$-functions of each policy $\pi_z$ for all rewards
$\transp{\phi}z'$ for all $z'\in \R^d$ including $z'\neq z$.
We have
found this improves performance.

\subsection{Learning FB Representations: the FB Training Loss}
\label{sec:FBloss}

The successor measure $M^\pi$ satisfies a Bellman-like equation $M^\pi =
P + \gamma P_\pi M^\pi$, as matrices in the finite case and as measures
in the general case \cite{successorstates}. We can learn FB by
iteratively minimizing the Bellman residual on the parametric model
$M=\transp{F}B\rho$. Using a suitable norm
$\norm{\cdot}_\rho$ for the Bellman residual (Appendix~\ref{app:FBloss})
leads to a loss expressed as expectations from the dataset:
\begin{align}
\label{eq:residualFB}
\mathcal{L}(F, B) &\deq \left \| \transp{F_z} B \rho -
\left(P + \gamma P_{\pi_z} \transp{\bar{F}_z} \bar{B}\rho \right) \right \|^2_\rho \\
\label{eq:FBloss}
 & =
 \E_{\substack{(s_t, a_t,s_{t+1}) \sim \rho \\ s' \sim \rho}} \left[ \Big( F(s_t, a_t, z)^\top B(s') - \gamma \bar{F}(s_{t+1}, \pi_z(s_{t+1}), z)^\top \bar{B}(s')  \Big)^2 \right]
\nonumber \\
 &\phantom{=}- 2 \E_{(s_t, a_t,s_{t+1}) \sim \rho } \left[ F(s_t, a_t, z)^\top B(s_{t+1}) \right]   
 + \texttt{Const}
\end{align}
where the constant term does not depend on $F$ and $B$, and 
where as usual $\bar{F_z}$ and $\bar{B}$ are non-trainable target
versions
of $F$ and $B$ whose parameters are updated with a slow-moving average of
those of $F$ and $B$. Appendix~\ref{app:FBloss} quickly derives this
loss, with pseudocode in Appendix~\ref{sec:pseudocode}. Contrary to \cite{touati2021learning}, 
the
last term involves $B(s_{t+1})$ instead of
$B(s_t)$, because
we use $s_{t+1}$ instead of $s_t$ for the successor measures \eqref{eq:defM}. 
We sample random values of $z$ as described in
Appendix~\ref{sec:zsampling}.

We include an auxiliary loss (Appendix~\ref{app:FBloss}) to normalize the covariance of $B$, $\E_\rho
B\transp{B}\approx \Id$,
as in \cite{touati2021learning} (otherwise one can, e.g., scale $F$ up and $B$
down since only $\transp{F}B$ is fixed).

As with SFs above, only $\transp{F(\cdot,z)}z$ is needed for the
policies, while the loss above on the vector $F$ amounts to training
$\transp{F(\cdot,z)}z'$ for all pairs $(z,z')$. The
full loss is needed for joint training of $F$ and $B$. But we include
an auxiliary loss $\loss'(F)$ to focus training on the diagonal $z'=z$.
This
is obtained by multiplying the Bellman gap in $\loss$ by
$\transp{B}(B\rho\transp{B})^{-1}z$ on the right, to make
$\transp{F(\cdot,z)}z$ appear:
\begin{equation}
\label{eq:auxFBloss}
\loss'(F)\deq \E_{(s_t,a_t,s_{t+1})\sim \rho} \left[
\left(
\transp{F(s_t,a_t,z)}z-\transp{B(s_{t+1})}(\E_\rho
B\transp{B})^{-1}z-\gamma \transp{\bar F(s_{t+1},\pi_z(s_{t+1}),z)}z
\right)^2
\right].
\end{equation}
This trains $\transp{F(\cdot,z)}z$ as the $Q$-function for reward
$\transp{B}(\E_\rho
B\transp{B})^{-1}z$. 
Though $\loss=0$ implies
$\loss'=0$, adding $\loss'$ reduces the error on the part used for policies.
This departs from \cite{touati2021learning}.

%

\subsection{Learning Basic Features $\phi$ for Successor Features}
\label{sec:philoss}

SFs must be provided with basic state features
$\phi$. Any representation learning method can be used to supply
$\phi$.  We focus on prominent RL representation learning baselines, and
on those used in previous zero-shot RL candidates such as APS.
We now describe the precise learning objective for each.

\textbf{Random Features (\Rand).} We use a non-trainable randomly initialized network as features.

\textbf{Autoencoder (\AEnc).} We learn a decoder $f\from \R^d \to S$ to
recover the state from its representation $\phi$:
\begin{equation}
\label{eq:Autoencoder}
\min_{f, \phi} \E_{s \sim \D} [(f(\phi(s)) - s)^2].
\end{equation}

\textbf{Inverse Curiosity Module (\ICM)} aims at extracting the
controllable aspects of the environment~\cite{pathak2017curiosity}. The
idea is to train an inverse dynamics model $g\from \R^d \times \R^d \to
A$ to predict the action used for a transition between two consecutive
states. We use the loss
\begin{equation}
\label{eq:ICM}
\min_{g, \phi} \E_{(s_t, a_t, s_{t+1}) \sim \D} [\norm{g(\phi(s_t), \phi(s_{t+1})) - a_t}^2].
\end{equation} 

\textbf{Transition model (\Transition).} This is a one-step forward dynamic model
$f\from \R^d \times A \to S$ that predicts the next state from the
current state
representation: 
\begin{equation}
\label{eq:Transition}
\min_{f, \phi} \E_{(s_t, a_t, s_{t+1}) \sim \mathcal{D}} [(f(\phi(s_t), a_t) - s_{t+1})^2].
\end{equation} 

\textbf{Latent transition model (\Latent).} This is similar to the transition model but instead of predicting the next state, it predicts its representation:
\begin{equation}
\label{eq:Latent}
\min_{f, \phi} \E_{(s_t, a_t, s_{t+1}) \sim \D} [(f(\phi(s_t), a_t) - \phi(s_{t+1}) )^2].
\end{equation} 
A clear failure case of this loss is when all states are mapped to the
same representation. To avoid this collapse, we compute $\phi(s_{t+1})$
using a non-trainable version of $\phi$, with parameters corresponding to
a slowly moving average of the parameters of $\phi$, similarly to BYOL
\cite{byol}.

\textbf{Diversity methods (\APS).} VISR \cite{hansen2019fast} and its
successor APS \cite{liu2021aps} tackle zero-shot RL using SFs with
features $\phi$ built online from a diversity criterion. This criterion maximizes
the mutual information between a policy parameter and the features of the
states visited by a policy using that parameter
\cite{eysenbach2018diversity, gregor2016variational}. VISR and APS use, respectively, a variational or
nearest-neighbor estimator for the mutual information. We directly use
the code provided for APS, and refer to \cite{liu2021aps} for the
details. Contrary to other methods, APS is not offline: it needs to be
trained on its own replay buffer.

\textbf{Laplacian Eigenfunctions (\lap).}
~\cite{wu2018laplacian} consider the symmetrized MDP graph Laplacian
induced by an exploratory policy $\pi$, defined as $\mathcal{L} = \Id -
\frac{1}{2} (P_\pi\diag(\rho)^{-1} + \diag(\rho)^{-1}(P_\pi)^\top)$. They
propose to learn the eigenfunctions of $\mathcal{L}$ via the spectral graph drawing objective~\cite{koren2003spectral}: 
\begin{equation}
\label{eq:Lap}
\min_{\phi} \E_{(s_t, s_{t+1}) \sim \D}\left[ \norm{\phi(s_t) -
\phi(s_{t+1})}^2\right] +
\lambda \E_{\substack{ s \sim \D \\ s' \sim \D}}\left[ (\transp{\phi(s)}
\phi(s'))^2 - \| \phi(s)\|^2_2 - \| \phi(s')\|^2_2\right]
\end{equation}
where the second term is an orthonormality regularization to
ensure that $\E_{s \sim \rho}[\phi(s) \transp{\phi(s)}] \approx \Id $, and
$\lambda > 0 $ is the regularization weight.
This is implicitly contrastive, pushing features of $s_t$
and $s_{t+1}$ closer while keeping features apart overall.
Such eigenfunctions have long been argued to play a key role in RL
\cite{mahadevan2007proto,machado2017laplacian}.

\textbf{Low-Rank Approximation of $P$ (\LRAP):} we 
learn features by estimating a low-rank model of the transition
probability densities: $P(\d s' | s, a) \approx \chi(s, a)^\top
\mu(s')\,\rho(\d s')$. Knowing $\rho$ is not needed: the corresponding loss on $\transp{\chi}\mu-P/\rho$
is readily expressed as
expectations over the dataset,
\begin{align}
\min_{\chi, \mu} &\E_{\substack{ (s_t, a_t) \sim \rho \\ s' \sim \rho }}
\left[\left (\chi(s_t, a_t)^\top \mu(s') - \frac{P(\d s' | s_t, a_t)}{\rho(\d s')} \right)^2 \right] \\
& =
\E_{\substack{ (s_t, a_t) \sim \rho \\ s' \sim \rho
}}[(\chi(s_t, a_t)^\top \mu(s'))^2]
-2 \E_{(s_t, a_t,s_{t+1}) \sim \rho} [ \chi(s_t, a_t)^\top
\mu(s_{t+1}) ]
 + \texttt{Const}
\label{eq:LRA_P}
\end{align}
We normalize $\E_{\rho} [\mu\transp{\mu}]\approx\Id$ with the same loss used
for $B$.
Then we use SFs with $\phi\deq \mu$. If the model $P=\transp{\chi}\mu\,\rho$ is
exact,
this provides exact optimal policies for any
reward (Appendix~\ref{sec:SFandP}, Thm.~\ref{thm:SFandP}).

This loss is implicitly contrastive: it compares samples $s_{t+1}$
to independent samples $s'$ from
$\rho$. It is an asymmetric extension of the Laplacian loss
\eqref{eq:Lap}
(Appendix~\ref{sec:lapandP}).
The loss~\eqref{eq:LRA_P} is also a special case of the FB loss~\eqref{eq:FBloss}
by setting $\gamma=0$, omitting $z$, and substituting
$(\chi,\mu)$ for $(F,B)$.
Indeed, FB
learns
a finite-rank model of $P(\Id - \gamma P_\pi)^{-1}$, which 
equals $P$ when $\gamma =0$. 

A related loss is introduced in~\cite{ren2022spectral}, but
involves a second unspecified, arbitrary probability distribution
$p$, which must cover the whole state space and whose analytic expression
must be known. It is unclear how to set a suitable $p$ in general.


\textbf{Contrastive Learning (\CL)} methods learn representations
by pushing positive pairs (similar states) closer together while keeping
negative pairs apart. Here, two states are considered similar if they
lie close on the same trajectory.  We use a SimCLR-like
objective~\cite{chen2020simple}:
\begin{equation}
\label{eq:CL}
\min_{\chi, \phi} - \E_{ \substack{ k \sim \texttt{Geom}(1-\gamma_{\CL}) \\ (s_t, s_{t+k}) \sim \D}} \left[ \log \frac{\exp(\texttt{cosine}(\chi(s_t), \phi(s_{t+k})))}{ \E_{s' \sim \D} \exp(\texttt{cosine}(\chi(s_t), \phi(s')))} \right]
\end{equation} 
where $s_{t+k}$ is the state encountered at step $t+k $
along the subtrajectory that starts at $s_t$, 
where $k$ is sampled from a geometric distribution of parameter
$(1-\gamma_{\CL})$, and $\texttt{cosine}(u, v) =
\frac{u^\top v}{\| u\|_2 \| v\|_2}, \forall u,v \in \R^d$ is the cosine
similarity function. Here $\gamma_{\CL}\in [0;1)$ is a parameter 
not necessarily set to the MDP's discount factor
$\gamma$.
\CL requires a dataset made of
full trajectories instead of isolated transitions.

\CL is tightly related to the spectral decomposition of the successor
measure $\sum_t \gamma_{\CL}^{t}
P_{\pi}^{t+1}$,
where $\pi$ is the behavior policy
generating the dataset trajectories.
Precisely, assuming that $\chi$ and $\phi$ are centered with unit norm,
and expanding the $\log$ and 
$\exp$ at second order,
the loss~\eqref{eq:CL} becomes
\begin{align}
\label{eq:CLapprox}
\eqref{eq:CL} \approx & 
\,\tfrac12
\E_{\substack{s\sim \D \\ s' \sim \D} }[ (\chi(s)^\top\phi(s'))^2]
- \E_{ \substack{ k \sim \texttt{Geom}(1-\gamma_{\CL}) \\ (s_t,
s_{t+k}) \sim \D}} \left[ \chi(s_t)^\top\phi(s_{t+k}) \right]
\end{align}
(compare \eqref{eq:LRA_P}).
Now, the law of $s_{t+k}$ given $s_t$ is
given by the stochastic matrix $(1-\gamma_{\CL})\sum_t \gamma_{\CL}^{t}
P_{\pi}^{t+1}$, the rescaled successor measure of 
$\pi$. Then one finds that \eqref{eq:CLapprox}
is minimized when $\chi$ and $\phi$ provide the singular
value decomposition of this matrix
in $L^2(\rho)$ norm (Appendix~\ref{sec:CLandSM}).
Formal links between contrastive learning and
spectral methods can be found in~\cite{tian2022deep,
balestriero2022contrastive}.

\textbf{Low-Rank Approximation of SR (\LRASR).} 
The \CL
method implicitly factorizes the successor measure of the exploration policy in Monte
Carlo fashion by sampling pairs $(s_t,s_{t+k})$ on the same
trajectory. This may suffer from high variance. 
To mitigate this, we propose
to factorize this successor measure
by temporal difference
learning instead of Monte Carlo. This is
achieved with an FB-like loss \eqref{eq:FBloss} except we drop the
 policies $\pi_z$ and learn successor measures for the exploration
policy only:
\begin{equation}
\label{eq:LRASR}
\min_{\chi, \phi}
\E_{\substack{(s_t, s_{t+1}) \sim
\D \\ s' \sim \D} }\left[ \left(\transp{\chi(s_t)}\phi(s') - \gamma
\transp{\bar{\chi}(s_{t+1})} \bar{\phi}(s') \right)^2 \right] 
- 2\E_{ \substack{ (s_t, s_{t+1}) \sim \D}} \left[
\transp{\chi(s_t)}\phi(s_{t+1}))) \right]
\end{equation}
with $\bar{\chi}$ and $\bar{\phi}$ target versions of $\chi$ and $\phi$.
We normalize $\phi$ to $\E_\rho[\phi\transp{\phi}]\approx \Id$ with the same
loss as for $B$.
Of all SF variants tested, this is the closest to FB.

\section{Experimental Results on Benchmarks}


%

%

Each of the 11 methods (FB and 10 SF-based models) has been
tested on 13 tasks in 4 environments from the Unsupervised RL and ExORL benchmarks
\cite{laskin2021urlb,yarats2021reinforcement}: Maze (reach 20 goals),
Walker (stand, walk, run, flip), Cheetah (walk, run, walk backwards, run
backwards), and Quadruped (stand, walk, run, jump); see Appendix~\ref{sec:envs}.
Each task and method was repeated for 3 choices
of replay buffer from ExORL:  RND, APS, and
Proto (except for the APS method, which can only train on the APS
buffer). Each of these 403 settings was repeated with 10 random seeds.  

The full setup
is described in Appendix~\ref{sec:expsetup}.
Representation dimension is $d=50$, except for Maze ($d=100$).
After model training,
tasks are revealed by 10,000 reward samples as in \cite{liu2021aps},
except for Maze, where a known goal is presented and used to set $z_r$
directly.
The code can be found at
\url{https://github.com/facebookresearch/controllable_agent} .

As toplines, we use online TD3 (with task rewards, and free environment
interactions not restricted to a
replay buffer), and offline TD3 (restricted to each replay buffer
labelled with task rewards). Offline TD3 gives an idea of the best achievable
performance given the training data in a buffer.

\begin{figure}
\begin{center}
\includegraphics[width=.98\textwidth]{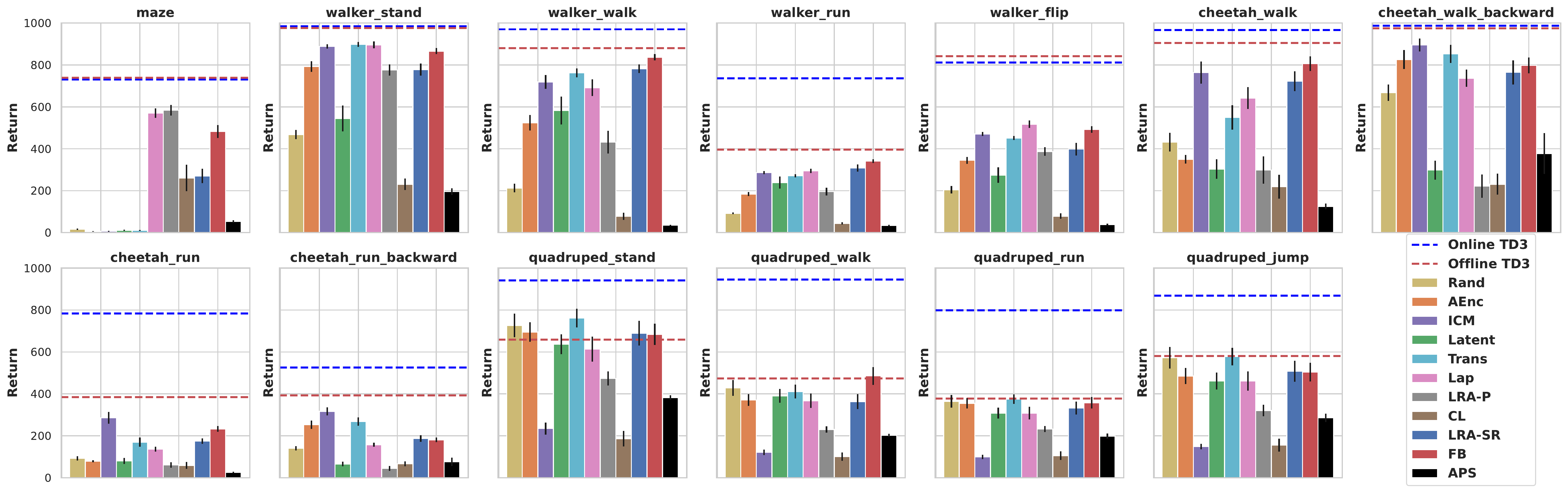}
\end{center}
\caption{Zero-shot scores for each task, with supervised online and
offline TD3 as toplines. Average over 3 replay buffers and 10 random
seeds.
\label{fig:taskscores}
}
\end{figure}

In Fig.~\ref{fig:taskscores} we plot the performance of each method
for each task in each environment, 
averaged over the three replay buffers and ten random seeds.
Appendix~\ref{sec:allresults} contains
the full results and more plots.

Compared to offline TD3 as a reference,
on the Maze tasks, \lap, \LRAP, and \FB perform well. On the Walker tasks,
\ICM, \Transition, \lap, \LRASR, and \FB perform well. On the Cheetah
tasks, \ICM, \Transition, \lap, \LRASR, and \FB perform well. On the
Quadruped tasks, many methods perform well, including,
surprisingly, random features. Appendix~\ref{sec:features} plots some
of the learned features on Maze.

On Maze, none of the encoder-based losses learn good policies, contrary to FB and
spectral SF methods. We believe this is because $(x,y)$ is already
a good representation of the 2D state $s$, so the encoders do
nothing.
Yet SFs on $(x,y)$ cannot solve the task (SFs on rewards of the form
$ax+by$ don't recover goal-oriented tasks): planning with SFs requires
specific representations.

Fig.~\ref{fig:global} reports aggregated scores over all tasks. To
average across tasks, we normalize scores:
for each task and replay buffer, performance is expressed as
a percentage of the performance of offline TD3 on the same replay buffer,
a natural supervised topline given the data. These normalized scores are
averaged over all tasks in each environment, then over environments to
yield the scores in Fig.~\ref{fig:global}.
The variations over environments, replay buffers and
random seeds are reported in Appendix~\ref{sec:allresults}.

These results are broadly consistent over replay buffers.
Buffer-specific results are
reported in
Appendices~\ref{sec:fullplots}--\ref{sec:buffer}.
Sometimes a replay buffer is restrictive, as attested by poor offline TD3
performance, starkly so for the Proto buffer on all Quadruped
tasks.  APS does not work well as a zero-shot RL method, but it does work
well as an exploration method: on average, the APS and RND replay buffers
have close results.

\FB and \lap are the only methods that perform consistently well, both
over tasks and over replay buffers.
Averaged on all tasks and buffers, \FB reaches 81\% of supervised offline TD3 performance,
and 85\% on the RND buffer. The second-best method is \lap with
74\% (78\% on the Proto buffer).

%

\section{Discussion and Limitations}

\paragraph{Can a few features solve many reward functions? Why are some
features better?}
The choice of features is critical.  Consider goal-reaching tasks:
the obvious way to learn to reach arbitrary goal states via SFs is to
define one feature per possible goal state (one-hot encoding). This
requires $\abs{S}$ features, and does not scale to continuous spaces. But
much better features exist.  For instance, with an size-$n$ cycle
$S=\{0,\ldots,n-1\}$ with actions that move left and right modulo $n$,
then \emph{two} features suffice instead of $n$: SFs with
$\phi(s)=(\cos(2\pi s/n),\sin(2\pi s/n))$ provides exact optimal policies
to reach any arbitrary state. On a $d$-dimensional grid
$S=\{0,\ldots,n-1\}^d$, just $2d$ features (a sine and cosine in each
direction) are sufficient to reach any of the $n^d$ goal states via SFs.

Goal-reaching is only a subset of possible RL tasks, but this clearly
shows that some features are better than others.
The sine and cosine are
the main eigenfunctions of the graph Laplacian on the grid:
such features have long been argued to play a special role in RL
\cite{mahadevan2007proto,machado2017laplacian}. FB-based methods are theoretically known to
learn such eigenfunctions \cite{successorstates}. Yet a precise
theoretical link
to downstream performance is still lacking.

\paragraph{Are these finite-rank models reasonable?} FB crucially relies on a
finite-rank model of $\sum \gamma^t P_\pi^t$, while some SF
variants above rely on finite-rank models of $P$ or the corresponding
Laplacian. It turns out such approximations are very different for $P$ or
for $\sum \gamma^t P_\pi^t$.

Unfortunately, despite the popularity of low-rank $P$ assumptions in the
theoretical literature, $P_\pi$ is \emph{always} close to $\Id$ in situations
where $s_{t+1}$ is close to $s_t$, such as any continuous-time physical system
(Appendix~\ref{sec:LRapprox}).
Any low-rank model of $P_\pi$ will be poor; actually $P_\pi$ is better
modeled as $\Id-\,\text{low-rank}$, thus approximating the Laplacian.
On the other hand, $P_\pi^t$ with large $t$ gets close to rank one under
weak assumptions (ergodicity), as $P_\pi^t$ converges to an equilibrium
distribution when $t\to \infty$. Thus the spectrum of the successor measures
$\sum\gamma^t P_\pi^t$ is usually more spread out (details and examples
in Appendix~\ref{sec:LRapprox}), and a low-rank model makes sense. The eigenvalues of $P_\pi$ are close to
$1$ and there is little signal to differentiate between eigenvectors, but
differences become clearer over time on $P_\pi^t$.
This may explain the better performance of \FB and \LRASR compared to \LRAP.

\paragraph{Limitations.} First, these experiments are still
small-scale and performance is not perfect, so there is space for
improvement. Also, all the environments tested here were deterministic: all
algorithms still make sense in stochastic environments, but experimental
conclusions may differ.

Second, even though these methods can be coupled with any exploration
technique, this will obviously not cover tasks too different
from the actions in the replay buffer, as with any offline RL method.

These zero-shot RL algorithms learn to summarize the long-term future for
a wide range of policies (though without synthesizing trajectories). This
is a lot: in contrast, world models only learn
policy-independent one-step transitions.  So the question remains
of how far this can scale.  A priori, there is no
restriction on the inputs (e.g., images, state history...). Still, for
large problems, some form of prior seems unavoidable. For SFs, priors can
be integrated in $\phi$, but rewards must be linear in $\phi$. For FB,
priors can be integrated in $B$'s input.  \cite{touati2021learning} use FB with
pixel-based inputs for $F$, but only the agent's $(x,y)$ position for
$B$'s input: this recovers all rewards that are functions of $(x,y)$
(linear or not).  Breaking the symmetry of $F$ and $B$ reduces the strain
on the model by restricting predictions to fewer variables, such as an
agent's future state instead of the full environment.
%


\section{Conclusions}

Zero-shot RL methods avoid test-time planning by summarizing long-term
state-state relationships for particular policies. We systematically
tested forward-backward representations and many new models of successor
features on zero-shot tasks. 
\option{We also uncovered algebraic links between SFs, FB, contrastive
learning, and spectral methods. Overall,}
SFs suffer from their dependency to basic feature
construction, with only Laplacian eigenfunctions working reliably.
Notably, planning with SFs requires specific features: SFs
can fail with
generic encoder-type feature learning, even if
the learned representation of states is reasonable (such as $(x,y)$).
Forward-backward
representations were best across the board, and provide reasonable
zero-shot RL performance.

%
%
%
%
%

\newpage

\section*{Acknowledgements}

The authors would like to thank Olivier Delalleau, Armand Joulin,
Alessandro Lazaric, Sergey Levine, Matteo Pirotta, Andrea Tirinzoni, and
the anonymous reviewers for helpful comments and questions on the
research and manuscript.

\bibliography{biblio.bib}
\bibliographystyle{alpha}

\newpage
\appendix

\section*{List of Appendices}

Appendix~\ref{app:proofs} sketches a proof of the basic properties
of SFs and the FB representation, Theorems~\ref{thm:SF}
and \ref{thm:FB}, respectively from \cite{borsa2018universal} and
\cite{touati2021learning}.

Appendix~\ref{app:FBloss} derives the loss \eqref{eq:FBloss} we use
to learn $F$ and $B$, and the auxiliary orthonormalization loss.

Appendix~\ref{sec:CLandSM} describes the precise mathematical relationship
between the contrastive loss
\eqref{eq:CL} and successor measures of the exploration
policy.

Appendix~\ref{sec:LRapprox} further discusses why low-rank models
make more sense on successor measures than on the transition matrix $P$.

Appendix~\ref{sec:SFandP} proves that if $P$ is low-rank given by
$\transp{\chi}\mu$, then successor features using $\phi=\mu$ (but not $\chi$)
provide optimal policies.

Appendix~\ref{sec:lapandP} proves that the loss \eqref{eq:LRA_P}
used to learn low-rank $P$ is
an asymmetric version of the Laplacian eigenfunction loss \eqref{eq:Lap}.

Appendix~\ref{sec:expsetup} describes the detailed experimental setup:
environments, architectures, policy training, methods for sampling of $z$, and
hyperparameters.
The full code can be found at \\
\url{https://github.com/facebookresearch/controllable_agent}

Appendix~\ref{sec:allresults} contains the full table of experimental results, as
well as aggregate plots over several variables (per task, per replay
buffer, etc.).

Appendix~\ref{sec:features} analyzes the learned features. We use feature
rank analysis to study the degree of feature collapse for some methods.
We also provide t-SNE visualizations of the learned embeddings for all
methods (for the Maze environment).

Appendix~\ref{sec:sensitivity} provides hyperparameter sensitivity plots
(latent dimension $d$, learning rate, batch size, mixing ratio for $z$
sampling).

Appendix~\ref{sec:gobaseline} describes a further baseline where we take
a goal-oriented method inspired from \cite{ma2020universal}, and extend it to
dense rewards by linearity. Since results were very poor except for Maze (which
is goal-oriented), we did not discuss it in the main text.

Appendix~\ref{sec:pseudocode} provides
PyTorch snippets for the key losses, notably the \FB
loss, the \SF loss as well as the various feature learning methods for \SF.

\newpage
\section{Sketch of Proof of Theorems~\ref{thm:SF} and \ref{thm:FB}}
\label{app:proofs}

To provide an intuition behind Theorems~\ref{thm:SF} and \ref{thm:FB} on
SFs and FB, we include here a sketch of proof in the finite state case.
Full proofs in the general case can be found in the Appendix of
\cite{touati2021learning}.

For Theorem~\ref{thm:SF} (SFs), let us assume that the reward is linear
in the features $\phi$, namely, $r(s)=\transp{\phi(s)}w$ for some $w\in
\R^d$. Then by definition, $z_r=w$ since $z_r$ is the linear regression
of the reward on the features (assuming features are linearly
independent). Using the definition \eqref{eq:defUSF} of the successor
features $\psi$, and taking the dot product with $z_r$, we obtain
\begin{align}
\transp{\psi(s_0,a_0,z_r)}z_r=\E
\left[\sum_t \gamma^t
\transp{\phi(s_{t+1})}z_r \mid s_0,a_0,\pi_{z_r} \right]
=\E
\left[\sum_t \gamma^t
r(s_{t+1}) \mid s_0,a_0,\pi_{z_r} \right]
\end{align}
since $r=\transp{\phi}w$. This means that $\transp{\psi(s_0,a_0,z_r)}z_r$
is the $Q$-function of reward $r$ for policy $\pi_{z_r}$. At the same
time, by the definition \eqref{eq:defUSF}, $\pi_{z_r}$ is defined as the argmax of
$\transp{\psi(s_0,a_0,z_r)}z_r$. Therefore, the policy $\pi_{z_r}$ is the
argmax of its own $Q$-function, meaning it is the optimal policy for
reward $r$.

For Theorem~\ref{thm:FB} (FB), let us assume that FB perfectly satisfies
the training criterion \eqref{eq:defFB}, namely,
$M^{\pi_z}=\transp{F_z}B\diag(\rho)$ in matrix form.
Thanks to the definition \eqref{eq:defSR} of successor representations $M^\pi$, 
for any
policy $\pi_z$, the $Q$-function for the reward $r$ can be written
as $Q^{\pi_z}_r = M^{\pi_z} r$ in matrix form. This is equal to $F_z^\top B
\diag(\rho) r $. Thus, if we define $z_r \deq B\diag(\rho) r= \E_{s \sim \rho}[ B(s) r(s)]$, we
obtain $Q^{\pi_z}_r = F_z^\top z_r$ for any $z \in \R^d$. In particular,
the latter holds for $z=z_r$ as well: $F_{z_r}^\top z_r$
is the $Q$-function of $\pi_{z_r}$. Again, the policies $\pi_z$ are defined in
\eqref{eq:defFB} as the greedy
policies of $F_z^\top z$, for any $z$. Therefore,
$\pi_{z_r}$ is the argmax of its own $Q$-function. Hence,
$\pi_{z_r}$ is the optimal policy for the reward $r$.

\newpage
\section{Derivation of the Forward-Backward Loss}
\label{app:FBloss}

Here we quickly derive the loss \eqref{eq:FBloss} used to train $F$ and
$B$ such that $M^\pi(s,a,\d s')\approx \transp{F(s,a)}B(s')\,\rho(\d s')$.
Training is based on the Bellman equation satisfied by $M^\pi$.  This
holds separately for each policy parameter $z$, so in this section we
omit $z$ for simplicity.

Here $\rho(\d s')$ is the distribution of states in the dataset.
Importantly, the resulting loss does not require to know this measure
$\rho(\d s')$, only to be able to sample states
$s\sim \rho$ from the dataset.

The successor measure $M^\pi$ satisfies a Bellman-like equation $M^\pi =
P + \gamma P_\pi M^\pi$, as matrices in the finite case and as measures
in the general case \cite{successorstates}. We can learn FB by
iteratively minimizing the Bellman residual $M^\pi-(P+\gamma P_\pi
M^\pi)$ on the parametric model
$M=\transp{F}B\rho$.

$M^\pi(s,a,\d s')$ is a measure on $s'$ for each $(s,a)$, so it is not
obvious how to measure the size of the Bellman residual. In general, we can define a
norm on such objects $M$ by taking the density with respect to the reference
measure $\rho$,
\begin{equation}
\norm{ M}_\rho^2 \deq
\E_{\substack{(s,a) \sim \rho \\ s' \sim \rho} } \left [\left(\frac{M(s
,a,\d
s')}{\rho(\d s')} \right)^2 \right]
\end{equation}
where $\frac{M(s, a,\d s')}{\rho(\d
s')}$ is the density of $M$ with respect to $\rho$. \footnote{This is the
dual norm on measures of the $L^2(\rho)$ norm on functions. It amounts to
learning a model of $M(s,a,\d s')$ by learning relative densities to
reach $s'$ knowing we start at
$(s,a)$, relative to the average density $\rho(\d s')$ in the
dataset.}
For finite states,
$M$ is a matrix $M_{sas'}$ and this is just a $\rho$-weighted Frobenius matrix
norm, $\norm{M}^2_\rho=\sum_{sas'} M_{sas'}^2\, \rho(s,a)/\rho(s')$. (This
is also how we proceed to learn a low-rank approximation of $P$ in
\eqref{eq:LRA_P}.)

We define the loss on $F$ and $B$ as the norm of the Bellman residual on
$M$ for the model $M=\transp{F}B\rho$. As usual in temporal difference
learning, we use fixed, non-trainable target networks $\bar F$ and $\bar
B$ for the right-hand-side of the Bellman equation. Thus, the Bellman
residual is $\transp{F}B\rho-\left(P + \gamma P_\pi \transp{\bar{F}}
\bar{B}\rho \right)$, and the loss is
\begin{align}
\label{eq:mainFBloss}
\mathcal{L}(F, B) &\deq \left \| \transp{F} B \rho -
\left(P + \gamma P_\pi \transp{\bar{F}} \bar{B}\rho \right) \right \|^2_\rho
\end{align}

When computing the norm $\norm{\cdot}_\rho$, the denominator $\rho$
cancels out with the $\transp{F}B\rho$ terms, but we are left with a
$P/\rho$ term. This term can still be integrated, because integrating
$P(\d s'|s,a)/\rho(\d s')$ under $s'\sim \rho$ is equivalent to
directly integrating under $s'\sim P(\d s'|s,a)$, namely, integrating
under transitions $(s_t,a_t,s_{t+1})$ in the environment. This plays out as follows:
\begin{align}
\mathcal{L}(F, B)
& = \E_{\substack{(s_t. a_t) \sim \rho \\ s' \sim \rho}} \left[ \left( F(s_t, a_t)^\top B(s') - \frac{P(\d s' | s_t, a_t)}{\rho(\d s')} - \gamma \E_{s_{t+1} \sim P(\d s_{t+1} | s_t, a_t)}[\bar{F}(s_{t+1}, \pi(s_{t+1}))^\top \bar{B}(s')]\right)^2\right]
 \\& =
 \E_{\substack{(s_t, a_t,s_{t+1}) \sim \rho \\ s' \sim \rho}} \left[ \Big( F(s_t, a_t)^\top B(s') - \gamma \bar{F}(s_{t+1}, \pi(s_{t+1}))^\top \bar{B}(s')  \Big)^2 \right]
\nonumber \\
 &\phantom{=}- 2 \E_{(s_t, a_t,s_{t+1}) \sim \rho } \left[ F(s_t, a_t)^\top B(s_{t+1}) \right]   
 + \texttt{Const}
\end{align}
where $\texttt{Const}$
is a constant term that we can discard since it does not depend on $F$
and $B$. \NDY{I removed the expression for the constant term, because it
is actually even more complicated than what was written, there were
two terms missing due to taking $\E_{s_1}$ out of the square for $\bar
F\bar B$}

Apart from the discarded constant term, all terms in this final
expression can be sampled from the dataset. Note that we have a $-2
\transp{F(s_t,a_t)}B(s_{t+1})$ term where \cite{touati2021learning} have a
$-2\transp{F(s_t,a_t)}B(s_t)$ term: this is because we define successor
representations \eqref{eq:defSR} using $s_{t+1}$ while
\cite{touati2021learning} use $s_t$.

See also Appendix~\ref{sec:pseudocode} for pseudocode (including the
orthonormalization loss, and double networks for $Q$-learning as
described in Appendix~\ref{sec:expsetup}).

\paragraph{The orthonormalization loss.} An auxiliary loss is used to normalize $B$ so that $\E_{s\sim \rho}
[B(s)\transp{B(s)}]\approx \Id$. This loss is
\begin{align}
\mathcal{L}_{\mathrm{norm}}(B)
&\deq
\norm{\E_\rho [B\transp{B}]-\Id}^2_{\mathrm{Frobenius}}
\\&=
\E_{s\sim \rho,\, s'\sim \rho}
\left[ (\transp{B(s)}
B(s'))^2
- \norm{B(s)}^2_2
- \norm{B(s')}^2_2\right]+\texttt{Const}.
\end{align}
(The more complex expression in \cite{touati2021learning} has the same gradients up to a
factor $4$.)

\paragraph{The auxiliary loss $\loss'$ \eqref{eq:auxFBloss}.} Learning $F(s,a,z)$ is
equivalent to learning $\transp{F(s,a,z)}z'$ for all vectors $z'$. Yet
the definition of the policies $\pi_z$ in FB only uses
$\transp{F(s,a,z)}z$. Thus, as was done for SFs in
Section~\ref{sec:psiloss}, one may wonder if there is a scalar rather
than vector loss to train $F$, that would reduce the error in the
directions of $F$ used to define $\pi_z$.

In the case of FB, the full vector loss is needed to train $F$ and $B$.
However, one can add the following auxiliary loss on $F$ to reduce the
errors in the specific direction $\transp{F(s,a,z)}z$. This is obtained as follows.

Take the Bellman gap in the main FB loss \eqref{eq:mainFBloss}: this
Bellman gap is
$\transp{F} B \rho -
\left(P + \gamma P_\pi \transp{\bar{F}} \bar{B}\rho \right)$. To
specialize this Bellman gap in the direction of $\transp{F(s,a,z)}z$, multiply by
$\transp{B} (B\rho\transp{B})^{-1}z$ on the right: this yields
$\transp{F}z -\left(P \transp{B} (B\rho\transp{B})^{-1}z+\gamma P_\pi
\transp{\bar{F}}z\right)$ (using that we compute the loss at $\bar B=B$).

This new loss is the Bellman gap on $\transp{F}z$, with reward $P
\transp{B} (B\rho\transp{B})^{-1}z$.

This is the loss $\loss'$ described in \eqref{eq:auxFBloss}. It is a
particular case of the main FB loss: $\loss=0$ implies $\loss'=0$, since
we obtained $\loss'$ by multiplying the Bellman gap of $\loss$.

We use $\loss'$
on top of the main FB loss to reduce errors in the direction
$\transp{F(s,a,z)}z$. However, in the end, the differences are modest.

\newpage
\section{Relationship Between Contrastive Loss and SVD of Successor
Measures}
\label{sec:CLandSM}

Here we prove the precise relationship between the contrastive loss
\eqref{eq:CL} and SVDs of the successor measure of the exploration
policy.

Intuitively, both methods push states together if they lie on the same
trajectory, by increasing the dot product between the representations of
$s_t$ and $s_{t+k}$. This is formalized as follows.

Let $\pi$ be the policy used to produce the trajectories in the dataset.
Define 
\begin{equation}
\tilde M\deq (1-\gamma_\CL)\sum_{t\geq 0} \gamma_\CL^t
P_\pi^{t+1}
\end{equation}
where $\gamma_\CL$ is the parameter of the geometric distribution used to
choose $k$ when sampling $s_t$ and $s_{t+k}$.

$\tilde M$ is a stochastic matrix in the discrete case, and a
probability measure over $S$ in the general case: it is the normalized
version of the successor measure \eqref{eq:defM} with $\pi$ the
exploration policy.

By construction, the distribution of $s_{t+k}$ knowing
$s_t$ with $k \sim \texttt{Geom}(1-\gamma_{\CL})$ is described by $\tilde
M$. Therefore, we can rewrite the loss as
\begin{align}
& \phantom{=} - \E_{ \substack{ k \sim \texttt{Geom}(1-\gamma_{\CL})
\\ (s_t, s_{t+k}) \sim \D}} \left[ \log
\frac{\exp(\texttt{cosine}(\phi(s_t), \mu(s_{t+k})))}{ \E_{s' \sim \D}
\exp(\texttt{cosine}(\phi(s_t), \mu(s')))} \right]
\\
&= -\E_{s\sim \rho,\,s'\sim \rho} \left[ 
\frac{\tilde M(s,\d s')}{\rho(\d s')}
\log
{\exp(\texttt{cosine}(\phi(s), \mu(s')))} \right]
\nonumber
\\
&\phantom{=} +\E_{s\sim \rho} \left[ 
\log
\E_{s' \sim \D}
\exp(\texttt{cosine}(\phi(s), \mu(s'))) \right]
\end{align}

Assume that $\phi$ and $\mu$ are centered with unit norm, namely,
$\norm{\phi(s)}_2=1$ and $\E_{s\sim \rho} \phi(s)=0$ and likewise for
$\mu$. With unit norm, the cosine becomes just a dot product, and the
loss is
\begin{equation}
\cdots = -\E_{s\sim \rho,\,s'\sim \rho} \left[ 
\frac{\tilde M(s,\d s')}{\rho(\d s')}
\transp{\phi(s)} \mu(s')\right]
+\E_{s\sim \rho} \left[ 
\log
\E_{s' \sim \D}
\exp(\transp{\phi(s)}\mu(s')) \right].
\end{equation}
A second-order Taylor expansion provides
\begin{equation}
\log \E \exp X=\E X + \tfrac12 \E [X^2] -\tfrac12 (\E X)^2 +O (\abs{X}^3)
\end{equation}
and therefore, with $\E \phi=\E \mu=0$, the loss is approximately
\begin{align}
\cdots &\approx -\E_{s\sim \rho,\,s'\sim \rho} \left[ 
\frac{\tilde M(s,\d s')}{\rho(\d s')}
\transp{\phi(s)} \mu(s')\right]
+\tfrac12 \E_{s\sim \rho,\,s'\sim \rho} \left[
\left(
\transp{\phi(s)}\mu(s')
\right)^2
\right]
\\&=\tfrac12 \E_{s\sim \rho,\, s'\sim \rho} \left[
\left(
\transp{\phi(s)} \mu(s')-\frac{\tilde M(s,\d s')}{\rho(\d s')}
\right)^2
\right]+\texttt{Const}
\end{align}
where the constant term does not depend on $\phi$ and $\mu$.

This is minimized when $\transp{\phi}\mu$ is the SVD of $\tilde M/\rho$
in the $L^2(\rho)$ norm.

\newpage
\section{Do Finite-Rank Models on the Transition Matrix and on
Successor Measures Make Sense?}
\label{sec:LRapprox}

It turns out finite-rank models are very different for $P$ or
for $\sum \gamma^t P_\pi^t$.
In typical situations, the spectrum
of $P$ is concentrated around $1$ while that of $\sum \gamma^t P_\pi^t$
is much more spread-out.

Despite the popularity of low-rank $P$ in theoretical RL works, $P$ is
\emph{never} close to low-rank in continuous-time systems: then $P$ is actually
always close to the identity.
Generally speaking, $P$ cannot be low-rank if most actions have a small
effect. By definition, for any feature function $\phi$, $(P_\pi \phi)(s)=\E
[\phi(s_{t+1})|s_t=s]$. Intuitively, if actions have a small effect, then $s_{t+1}$ is
close to $s_t$, and $\phi(s_{t+1})\approx \phi(s_t)$ for continuous
$\phi$. This means that $P_\pi \phi$ is close to $\phi$, so that $P_\pi$ is
close to the identity on a large subspace of feature functions $\phi$. 
In the theory of continuous-time Markov processes, the time-$t$ transition kernel
is given by $P_t=e^{t A}$ with $A$ the infinitesimal generator of the
process \cite[\S20.1]{wilmer2009markov}
\cite[\S8.1]{oksendal1998stochastic}, hence $P_t$ is $\Id+O(t)$ for small timesteps $t$.
In
general, the transition matrix $P$ is better modeled as $\Id+\,$low-rank,
which corresponds to a low-rank model of the
Markov chain Laplacian $\Id-P$.

On the other hand, though $\sum \gamma^t P^t_\pi$ is never exactly
low-rank (it is invertible),  it has
 meaningful low-rank approximations
under weak assumptions.
For large $t$, $P_\pi^t$ becomes rank-one under weak
assumptions (ergodicity), as it converges to the equilibrium distribution
of the transition kernel.
For large $\gamma$, the sum $\sum \gamma^t P^t_\pi$ is dominated by large
$t$. 
Most eigenvalues of $P_\pi$ are close to $1$,
but taking powers $P^t_\pi$ sharpens the differences between eigenvalues:
with $\gamma$ close to $1$, going from $P_\pi$ to $\sum \gamma^t P^t_\pi=(\Id-\gamma
P_\pi)^{-1}$ changes an eigenvalue $1-\eps$ into $1/\eps$.

In short, on $P$ itself, there is little learning signal to differentiate
between eigenvectors, but differences become visible over time. This may
explain why FB works better than low-rank decompositions directly based
on $P$ or the Laplacian.

For instance, consider the nearest-neighbor random walk on a length-$n$
cycle $\{0,1,\ldots, n-1 \mod n\}$, namely, moving in dimension $1$.
(This extends to any-dimensional grids.)
The
associated $P_\pi$ is not
low-rank in any reasonable sense: the corresponding stochastic matrix
is concentrated around the diagonal, and many eigenvalues are close to
$1$. Precisely, the eigenvalues are $\cos(2k\pi/n)$ with integer
$k=\{0,\ldots,n/2\}$.
This
is $\approx 1-2\pi^2(k/n)^2$ when $k\ll n$. 
Half of the eigenvalues are between $\sqrt{2}/2$ and $1$.

However, when
$\gamma \to 1$, $\sum \gamma^t P_\pi^t$ has one eigenvalue $1/(1-\gamma)$ and the
other eigenvalues are $\frac{1}{1-\cos(2k\pi/n)}\approx n^2/2\pi^2k^2$ with positive integer $k$:
there is one large eigenvalue, then the others decrease like
$\mathrm{cst}/k^2$.
With
such a spread-out spectrum, a finite-rank model makes sense.

\newpage
\section{Which Features are Analogous Between Low-Rank $P$ and SFs?}
\label{sec:SFandP}

Here we explain the relationship between a low-rank model of $P$ and
successor features. More precisely, if transition probabilities from
$(s,a)$ to $s'$ can be written exactly as $\transp{\chi(s,a)}\mu(s')$,
then SFs with basic features $\phi\deq \mu$ will provide optimal policies
for \emph{any} reward function (Theorem~\ref{thm:SFandP}).

Indeed, under the finite-rank model $P(\d s'|s,a)=\transp{\chi(s,a)}\mu(s')\rho(\d s')$, 
rewards only matter via
the reward features $\E_{s'\sim \rho} \mu(s')r(s')$: namely, two rewards with
the same reward features have the same $Q$-function, as the dynamics
produces the same expected rewards. Then $Q$-functions are linear in
these reward features, 
and using successor features with $\phi\deq \mu$ provides
the correct $Q$-functions, as follows.

\begin{thm}
\label{thm:SFandP}
Assume that $P(\d s'|s,a)=\transp{\chi(s,a)}\mu(s')\rho(\d s')$. Then
successor features using the basic features $\phi\deq \mu$ provide optimal
policies for any reward function.
\end{thm}

This is why we use $\mu$ rather than
$\chi$ for the SF basic features.
This is also why we avoided the traditional notation
$P(s'|s,a)=\transp{\phi(s,a)}\mu(s')$ often used for low-rank $P$, which
induces a conflict of notation with the $\phi$ in SFs, and suggests the
wrong analogy.

Meanwhile, $\chi$ plays a role more analogous to SFs' $\psi$, although
for one-step transitions instead of multistep transitions as in SFs:
$Q$-functions are linear combinations of the features $\chi$. In
particular, the optimal $Q$-function for reward $r$ is
$Q^\star_r=\transp{\chi}w_r$ for some $w_r$. But contrary to successor
features, there is no simple correspondence to compute $w_r$ from $r$.

\begin{proof}
Let $\pi$ be any policy.
On a finite space in
matrix notation, and omitting $\rho$ for simplicity, the assumption $P=\transp{\chi}\mu$
implies $P_\pi=\transp{\chi_\pi}\mu$
where $\chi_\pi(s)\deq \E_{a\sim \pi(s)} \chi(s,a)$ are the
$\pi$-averaged features. Then,
\begin{align}
Q^\pi_r&=
P\sum_{t\geq 0} \gamma^t P_\pi^t r
\\&=\transp{\chi} \mu \sum_{t\geq 0} \gamma^t
\left(\transp{\chi_\pi}\mu\right)^t r
\\&=\transp{\chi} \left(
\tsum_{t \geq 0}\, \gamma^t \left(\mu \transp{\chi_\pi}\right)^t
\right) \mu r.
\end{align}

Thus, $Q$-functions are expressed as
$Q^\pi_r(s,a)=\transp{\chi(s,a)}w(\pi,r)$ with $w(\pi,r)=\left(
\tsum_{t \geq 0}\, \gamma^t (\mu \transp{\chi_\pi})^t
\right) \mu r$.

Moreover, rewards only matter via $\mu r$. Namely, two rewards
with the same $\mu r$ have the same $Q$-function for every policy.

In full generality on continuous spaces with $\rho$ again, the same holds
with $\mu(s)\rho(\d s)$ instead of $\mu(s)$, and $\E_{s\sim \rho}
\mu(s)r(s)$ instead of $\mu r$.

Now, let $r$ be any reward function, and let $r'$ be its
$L^2(\rho)$-orthogonal projection onto the space generated by the
features $\mu$. By construction, $r-r'$ is $L^2(\rho)$-orthogonal to
$\mu$, namely, $\E_{\rho} \mu(r-r')=0$. So
$\E_{\rho} \mu r=\E_{\rho} \mu r'$.
Therefore, by the above, $r$ and $r'$ have the same $Q$-function for
every policy.

By definition, $r'$ lies in the linear span of the features $\mu$. By
Theorem~\ref{thm:SF}, SFs with features $\phi=\mu$ will provide optimal
policies for $r'$. Since $r$ and $r'$ have the same $Q$-function for
every
policy, an optimal policy for $r'$ is also optimal for $r$.
\end{proof}


\newpage
\section{Relationship between Laplacian Eigenfunctions and Low-Rank $P$
Learning}
\label{sec:lapandP}

The loss \eqref{eq:LRA_P} used to learn a low-rank model of the transition
probabilities $P$ is an asymmetric version of the Laplacian
eigenfunction loss \eqref{eq:Lap} with $\lambda=1$.

Said equivalently, if we use the low-rank $P$ loss \eqref{eq:LRA_P} constrained with
$\chi=\mu$ to learn a low-rank model of $P_\pi$ instead of $P$ (with
$\pi$ the exploration policy), then we get the Laplacian eigenfunction
loss \eqref{eq:Lap} with $\lambda=1$.

Indeed, set $\lambda=1$ in \eqref{eq:Lap}. Assume that the distributions
of $s_t$ and $s_{t+1}$ in the dataset are identical on average (this
happens, e.g., if the dataset is made of long trajectories or if $\rho$ is
close enough to the invariant distribution of the exploration policy). Then, in the
Laplacian loss \eqref{eq:Lap}, the norms from the first term cancel those
from the second, and the Laplacian loss simplifies to
\begin{align}
\eqref{eq:Lap}&=
\E_{(s_t, s_{t+1}) \sim \D}\left[ \norm{\phi(s_t) -
\phi(s_{t+1})}^2\right] +
\E_{\substack{ s \sim \D \\ s' \sim \D}}\left[ (\transp{\phi(s)}
\phi(s'))^2 - \| \phi(s)\|^2_2 - \| \phi(s')\|^2_2\right]
\\&=\E_{s_t\sim \D}\norm{\phi(s_t)}^2+\E_{s_{t+1}\sim
\D}\norm{\phi(s_t)}^2-2\E_{(s_t, s_{t+1}) \sim \D}
\left[\transp{\phi(s_t)}\phi(s_{t+1})\right]
\\&\phantom{=}+\E_{\substack{ s \sim \D \\ s' \sim \D}}\left[ (\transp{\phi(s)}
\phi(s'))^2 - \| \phi(s)\|^2_2 - \| \phi(s')\|^2_2\right]
\\&= -2\E_{(s_t, s_{t+1}) \sim
\D}\left[\transp{\phi(s_t)}\phi(s_{t+1})\right]
+\E_{\substack{ s \sim \D \\ s' \sim \D}} \left[ (\transp{\phi(s)}
\phi(s'))^2\right].
\end{align}

This is the same as the low-rank loss \eqref{eq:LRA_P} if we omit actions
$a$ and constrain $\chi=\mu$.

\newpage
\section{Experimental Setup}
In this section we provide additional information about our experiments.

Code snippets for the main losses are given in
Appendix~\ref{sec:pseudocode}.
The full code can be found at \\
\url{https://github.com/facebookresearch/controllable_agent}

\label{sec:expsetup}

\subsection{Environments}
\label{sec:envs}
\begin{figure}
	\begin{center}
	\includegraphics[width=.2\textwidth]{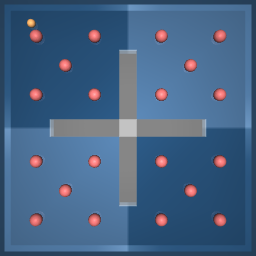}
	\includegraphics[width=.2\textwidth]{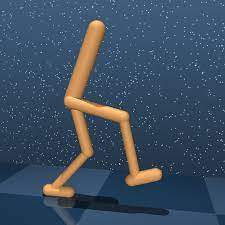}
	\includegraphics[width=.2\textwidth]{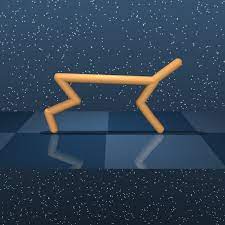}
	\includegraphics[width=.2\textwidth]{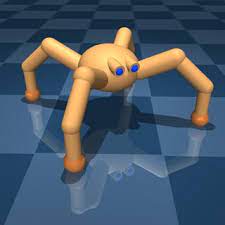}
	\caption{Maze, Walker, Cheetah and Quadruped environments used in our experiments.
	 In the Mmaze domain (left), we show an example of an initial state (yellow point) and the 20 test goals (red circles). 
	\label{fig:envs}
	}
	\end{center}
	\end{figure}

All the environments considered in this paper are based on the
\textit{DeepMind Control Suite}~\cite{tassa2018deepmind}.

\begin{itemize}
	\item \textbf{Point-mass Maze}: a 2-dimensional continuous maze with four rooms.
	 The states are 4-dimensional vectors consisting of positions and velocities
	  of the point mass $(x, y, v_x, x_y)$, and the actions are 2-dimensional vectors.
	  At test, we assess the performance of the agents on 20 goal-reaching tasks 
	  (5 goals in each room described by their $(x,y)$ coordinates).
	\item \textbf{Walker}:  a planar walker. States are 24-dimensional vectors consisting of positions and velocities of robot joints, and actions are 6-dimensional vectors.
	We consider 4 different tasks at test time: $\texttt{walker\_stand}$ reward is a combination of terms encouraging an upright torso and some minimal
	torso height, while $\texttt{walker\_walk}$ and $\texttt{walker\_run}$ rewards include a component encouraging some minimal forward velocity.
	 $\texttt{walker\_flip}$ reward includes a component encouraging some mininal angular momentum.
	\item \textbf{Cheetah}:  a running planar biped. States are 17-dimensional vectors consisting of positions and velocities of robot joints, and actions are 6-dimensional vectors.
	We consider 4 different tasks at test time: $\texttt{cheetah\_walk}$ and $\texttt{cheetah\_run}$ rewards are linearly proportional
	to the forward velecity up to some desired values: 2 m/s for $\texttt{walk}$ and 10 m/s for $\texttt{run}$.
	Similarly, $\texttt{walker\_walk\_backward}$ and $\texttt{walker\_run\_backward}$ rewards encourage reaching some minimal backward velocities.
	\item \textbf{Quadruped}: a four-leg ant navigating in 3D space. States and actions are 78-dimensional and 12-dimensional vectors, respectively. We consider 4 tasks at test time:
	$\texttt{quadruped\_stand}$ reward encourages an upright torso. $\texttt{quadruped\_walk}$ and $\texttt{quadruped\_run}$ include a term 
	encouraging some minimal torso velecities.
	$\texttt{quadruped\_walk}$ includes a term encouraging some
	minimal height of the center of mass. \todo{that's only 3 tasks}
\end{itemize}

\subsection{Architectures}
\label{sec:archi}

We use the same architectures for all methods.
\begin{itemize}
\item The backward representation network $B(s)$ and the feature
network $\phi(s)$ are represented by a feedforward neural network with three hidden layers,
each with 256 units, that takes as input a state and outputs a
$L2$-normalized embedding of radius $\sqrt{d}$. \todo{Why??? this loses a
lot of generality!!!}
\item For both successor features $\psi(s,a,z)$ and forward network
$F(s,a,z)$, we first preprocess separately $(s, a)$ and $(s, z)$ by two feedforward networks with two hidden layers (each with 1024 units) to 512-dimentional space.
Then we concatenate their two outputs and pass it into another 2-layer
feedforward network (each with 1024 units) to output a $d$-dimensional vector.

\item For the policy network $\pi(s,z)$, we first preprocess separately $s$ and $(s, z)$ by two feedforward networks with two hidden layers (each with 1024 units) to 512-dimentional space.
Then we concatenate their two outputs and pass it into another 2-layer feedforward network (each with 1024 units) to output
to output a $d_{A}$-dimensional vector, then we apply a \texttt{Tanh}
activation as the action space is $[-1,1]^{d_{A}}$. \todo{correct?}
\end{itemize}

For all the architectures, we apply a layer normalization~\cite{ba2016layer} and
 \texttt{Tanh} activation in the first layer in order to standardize the states and actions.
 We use $\texttt{Relu}$ for the rest of layers. We also pre-normalized $z$: $z \leftarrow \sqrt{d} \frac{z}{\| z\|_2}$ 
 in the input of $F$, $\pi$ and $\psi$.
Empirically, we observed that removing preprocessing and pass directly a concatenation of $(s, a, z)$ directly to the network leads to unstable training.
 The same holds when we preprocess $(s, a)$ and $z$ instead of $(s, a)$ and $(s, z)$,
  which means that the preprocessing of $z$ should be influenced by the current state. 

For maze environments, we added an additional hidden layer after the preprocessing
(for both policy and forward / successor features) as it helped to improve the results.

\subsection{Sampling of $z$}
\label{sec:zsampling}

We mix two methods for sampling $z$:
\begin{enumerate}
\item
We sample $z$ uniformly in the sphere of radius $\sqrt{d}$ in $\R^d$
(so each component of $z$ is of size $\approx 1$). 
\item
We sample $z$ using the formula for $z_r$ corresponding to the reward for
reaching a random goal state $s$
in Theorems~\ref{thm:SF}--\ref{thm:FB}. Namely, we set $z=B(s)$ for FB
and $z= \left( \sum_{i=1}^m \phi(s_i) \phi(s_i)^\top \right)^{+} \phi(s)$ 
for SFs, where
$s\sim \rho$ is a random state sampled from the replay buffer and $s_i$
are states in a minibatch.
\end{enumerate}

For the main series of results, we used a $50\%$ mix ratio for those two
methods. Different algorithms can benefit from different ratios: this is
explored in Appendix~\TODO.

\todo{}

\subsection{Learning the Policies $\pi_z$: Policy Network}
\label{sec:piloss}

As the action space is continuous, we could not compute the $\argmax$
over action in closed form. Instead,
we consider a latent-conditioned policy network  $\pi_\eta: S \times Z \to A$, 
and we learn the policy parameters $\eta$ by performing stochastic
gradient ascent 
on the objective $\E_{s, z} [F(s, \pi_\eta(s), z)^\top z]$ for FB or
$\E_{s, z} [\psi(s, \pi_\eta(s), z)^\top z]$ for SFs.

We also incorporate
techniques introduced in the TD3 paper~\cite{fujimoto2018addressing} to
address function approximation error in actor-critic methods: double
networks and target policy smoothing, adding noise $\eps$ to the actions.

Let $\theta_1$ and $\theta_2$ the parameters of two forward networks and
let $\omega$ the parameters of the backward network. Let  $\theta_1^{-}$,
$\theta_2^{-}$ and $\omega^{-}$ be the parameters of their corresponding target networks. 

Let $\{(s_i, a_i, s_i^\mathrm{next})\}_{i \in I} \subset \mathcal{D}$ a mini-batch of size $|I| = b$ of transitions
and let $\{z_i \}_{i \in I}$ a mini-batch of size $|I| = b$ of latent
variables sampled according to~\ref{sec:zsampling}.
 The empirical version of the main FB loss in~\eqref{eq:FBloss} is (with
 an additional $1/2$ factor):
\begin{align}
 \mathscr{L} (\theta_k, \omega) & = \frac{1}{2 b(b-1)} \sum_{ \substack{i, j \in I^2 \\ i \neq j}} \left( F_{\theta_k}(s_i, a_i,z_i)^\top B_\omega(s_j^\mathrm{next})  - \gamma \min_{l=1,2} F_{\theta_l^{-}}(s_i^\mathrm{next}, \pi_\eta(s_i^\mathrm{next}) + \epsilon_i ,z_i)^\top B_{\omega^{-}}(s_j^\mathrm{next})  \right)^2 \nonumber \\
 & - \frac{1}{b} \sum_{i \in I} F_{\theta_k}(s_i, a_i,z_i)^\top  B_\omega(s_i^\mathrm{next})  \quad \forall k =1,2
\end{align}
where $\epsilon_i$ is sampled from a truncated centered Gaussian with variance $\sigma^2$ (for policy smoothing).
The empirical version of the auxiliary $F$ loss in~\eqref{eq:auxFBloss} is: for $k = 1, 2$,

\begin{align}
	\mathscr{L'} (\theta_k) & = \frac{1}{b} \sum_{ \substack{i\in I}} \left( F_{\theta_k}(s_i, a_i,z_i)^\top z_i  
	- B_{\omega}(s_i^\mathrm{next})^\top \Cov^{+} z_i 
	- \gamma \min_{l=1,2} F_{\theta_l^{-}}(s_i^\mathrm{next}, \pi_\eta(s_i^\mathrm{next}, z) + \epsilon_i ,z_i)^\top z_i  \right)^2
\end{align}
where $\Cov^{+}$ is the pseudo-inverse of the empirical covariane matrix $\Cov = \frac{1}{b}\sum_{i \in I} B_{\omega}(s_{i}) B_{\omega}(s_{i})^\top$.
We use $1/d$ as regularization coefficient in front of $\mathscr{L'} (\theta_k)$.

For policy training, the empirical loss is:

\begin{align}
	\label{eq: policy_loss}
 \mathscr{L} (\eta) = - \frac{1}{b} \sum_{i \in I} \min_{l=1,2}F_{\theta_l}(s_i, \pi_\eta(s_i, z) + \epsilon_i,z_i)^\top z_i 
\end{align} 

The same techniques are also used for SFs.
\TODO

\subsection{Hyperparameters}

Table~\ref{table:hyperparams} summarizes the hyperparameters used in our
experiments.

A hyperparameter sensitivity analysis for two domains (Walker and
Cheetah) is included in Appendix~\ref{sec:sensitivity}.

\begin{table}[hb!]
	\caption{\label{table:hyperparams} Hyperparameters used in our experiments.}
	\centering
	\begin{tabular}{lc}
	\hline
	Hyperparameter       & Value \\
	\hline
	\: Replay buffer size &  $5 \times 10^6$ ($ 10 \times 10^6$ for maze) \\
	\: Representation dimension &  $50$ ($100$ for maze) \\
	\: Batch size & $1024$ \\
	\: Discount factor $\gamma$ & $0.98$ ($0.99$ for maze) \\
	\: Optimizer & Adam \\
	\: Learning rate & $10^{-4}$ \\
	\: Mixing ratio for $z$ sampling & $0.5$ \\
	\: Momentum coefficient for target networks & $0.99$ \\
	\: Stddev $\sigma$ for policy smoothing & $0.2$ \\
	\: Truncation level for policy smoothing & $0.3$ \\
	\: Number of gradient steps  &  $10^6$ \\
	\: Number of reward labels for task inference & $10^4$ \\
	\: Discount factor $\gamma_{\CL}$ for $\CL$ & $0.6$ ($0.2$ for maze) \\
	\: Regularization weight for orthonormality loss (spectral methods) & 1 \\
	\hline
	\end{tabular}
	
	\end{table}

\paragraph{Hyperparameter tuning.}
Since all methods share the same core, we chose to use the same
hyperparameters rather than tune per method, which could risk leading to
less robust conclusions.

We did not do hyperparameter sweeps for each baseline and task, first
because this would have been too intensive given the number of setups,
and, second, this would be too close to going back to a supervised method for
each task.

Instead, to avoid any overfitting, we tuned architectures and
hyperparameters by hand on the Walker environment only, with the RND
replay buffer, and reused these parameters across all methods and tasks
(except for Maze, on which all methods behave differently). We identified
some trends by monitoring learning curves and downstream performance on
Walker, and we fixed a configuration that led to overall good performance
for all the methods. We avoided a full sweep to avoid overfitting based
on Walker, to focus on robustness.

For instance, the learning rate $10^{-4}$ seemed to work well with
\emph{all}
methods, as seen in Appendix~\ref{sec:sensitivity}.

Some trends were common between all methods: indeed, all the methods share a
common core (training of the successor features $\psi$ or $F$, and
training of the policies $\pi$) and differ by the training of the basic
features $\phi$ or $B$.

For the basic features $\phi$, the various representation learning losses
were easy to fit and had low hyperparameter sensitivity: we always
observed smooth decreasing of losses until convergence. The challenging
part was learning $\psi$ and $F$ and their corresponding policies, which is
common across methods.

\newpage
\section{Detailed Experimental Results}
\label{sec:allresults}

We report the full experimental results, first as a table
(Section~\ref{sec:fulltable}).

In Section~\ref{sec:aggplots} we plot aggregated results for easier
interpretation: first, aggregated across all tasks with a plot of the
variability for each method; second, aggregated over environments (since
each environment corresponds to a trained zero-shot model); third, for
each individual task but still aggregated over replay buffers.

In Section~\ref{sec:fullplots} we plot all individual results per task
and replay buffer.

In Section~\ref{sec:buffer} we plot aggregate results split by replay buffer.

\todo{should we also plot unnormalized scores aggregated over all tasks?
I don't want people saying the results are too dependent upon offline TD3
normalization.}

\subsection{Full Table of Results}
\label{sec:fulltable}

\begin{table}[h]
	\centering
    \footnotesize
	\resizebox{1.2\columnwidth}{!}{%
	\begin{tabular}{llllllllllllll}
	\toprule
	        \textbf{Buffer} & \textbf{Domain} & \textbf{Task} & & & &
		& & \textbf{Method} \\
		&  &  & \Rand & \AEnc & \ICM & \Latent & \Transition & \lap & \LRAP & \CL & \LRASR & \FB & \APS \\
	   \toprule
	   \multirow{13}{*}{APS} & cheetah & run & 145$\pm$7 & 97$\pm$3 & \hl{432$\pm$8} & 49$\pm$14 & 133$\pm$15 & 198$\pm$4 & 8$\pm$1 & 2$\pm$0 & \hl{247$\pm$10} & \hl{267$\pm$33} & 25$\pm$3 \\
 &  & run-backward & 189$\pm$20 & \hl{365$\pm$6} & \hl{404$\pm$3} & 32$\pm$3 & \hl{382$\pm$3} & 221$\pm$4 & 1$\pm$0 & 14$\pm$6 & 261$\pm$5 & 238$\pm$7 & 98$\pm$21 \\
 &  & walk & 665$\pm$60 & 404$\pm$19 & \hl{928$\pm$54} & 302$\pm$67 & 287$\pm$53 & \hl{900$\pm$49} & 75$\pm$31 & 2$\pm$1 & \hl{918$\pm$22} & 844$\pm$51 & 144$\pm$11 \\
 &  & walk-backward & 653$\pm$75 & 982$\pm$0 & \hl{986$\pm$0} & 453$\pm$105 & \hl{985$\pm$0} & 937$\pm$17 & 8$\pm$1 & 150$\pm$51 & \hl{983$\pm$0} & 981$\pm$1 & 452$\pm$77 \\
 & maze & reach & 11$\pm$5 & 5$\pm$1 & 8$\pm$3 & 10$\pm$4 & 15$\pm$5 & \hl{432$\pm$18} & \hl{436$\pm$16} & 12$\pm$3 & 145$\pm$8 & \hl{410$\pm$16} & 59$\pm$7 \\
 & quadruped & jump & \hl{784$\pm$5} & \hl{727$\pm$12} & 164$\pm$20 & 554$\pm$24 & 624$\pm$14 & \hl{718$\pm$18} & 309$\pm$50 & 102$\pm$29 & 632$\pm$20 & 649$\pm$23 & 311$\pm$24 \\
 &  & run & \hl{487$\pm$1} & 459$\pm$5 & 91$\pm$19 & 382$\pm$14 & 411$\pm$16 & \hl{491$\pm$3} & 238$\pm$21 & 53$\pm$15 & 448$\pm$6 & \hl{476$\pm$8} & 196$\pm$10 \\
 &  & stand & \hl{966$\pm$3} & \hl{925$\pm$16} & 248$\pm$46 & 752$\pm$36 & 890$\pm$17 & \hl{963$\pm$1} & 497$\pm$53 & 56$\pm$9 & 872$\pm$20 & \hl{924$\pm$13} & 417$\pm$10 \\
 &  & walk & \hl{543$\pm$19} & 444$\pm$8 & 108$\pm$25 & 489$\pm$20 & 490$\pm$20 & \hl{524$\pm$13} & 228$\pm$26 & 45$\pm$14 & 463$\pm$18 & \hl{712$\pm$29} & 205$\pm$6 \\
 & walker & flip & 158$\pm$10 & 317$\pm$31 & \hl{452$\pm$3} & 299$\pm$53 & \hl{471$\pm$15} & \hl{454$\pm$12} & 340$\pm$18 & 69$\pm$26 & 186$\pm$21 & 413$\pm$16 & 39$\pm$2 \\
 &  & run & 96$\pm$4 & 127$\pm$8 & \hl{290$\pm$9} & \hl{359$\pm$11} & 263$\pm$12 & \hl{289$\pm$10} & 115$\pm$11 & 63$\pm$11 & 204$\pm$26 & \hl{346$\pm$14} & 35$\pm$3 \\
 &  & stand & 486$\pm$27 & 617$\pm$37 & \hl{925$\pm$19} & \hl{868$\pm$57} & 864$\pm$24 & \hl{895$\pm$9} & 643$\pm$35 & 205$\pm$51 & 591$\pm$35 & 822$\pm$26 & 176$\pm$17 \\
 &  & walk & 177$\pm$30 & 462$\pm$58 & 724$\pm$41 & \hl{857$\pm$8} & \hl{816$\pm$30} & 386$\pm$40 & 159$\pm$20 & 76$\pm$44 & 671$\pm$19 & \hl{817$\pm$15} & 34$\pm$2 \\
\hline
 \multirow{13}{*}{Proto} & cheetah & run & 58$\pm$8 & 84$\pm$10 & \hl{333$\pm$14} & 27$\pm$5 & \hl{322$\pm$5} & 142$\pm$2 & 149$\pm$3 & 0$\pm$0 & \hl{209$\pm$10} & \hl{210$\pm$13} & - \\
 &  & run-backward & 149$\pm$6 & 262$\pm$7 & \hl{329$\pm$3} & 30$\pm$5 & \hl{274$\pm$7} & 146$\pm$2 & 133$\pm$6 & 16$\pm$13 & \hl{230$\pm$6} & 157$\pm$7 & - \\
 &  & walk & 287$\pm$38 & 327$\pm$51 & \hl{961$\pm$24} & 112$\pm$11 & \hl{929$\pm$13} & 722$\pm$19 & 770$\pm$31 & 1$\pm$0 & 860$\pm$31 & \hl{908$\pm$18} & - \\
 &  & walk-backward & 664$\pm$39 & 973$\pm$3 & \hl{987$\pm$0} & 101$\pm$15 & \hl{982$\pm$0} & 798$\pm$16 & 629$\pm$31 & 151$\pm$86 & \hl{979$\pm$1} & 742$\pm$46 & - \\
 & maze & reach & 27$\pm$5 & 7$\pm$1 & 7$\pm$2 & 15$\pm$4 & 8$\pm$1 & \hl{571$\pm$16} & \hl{556$\pm$15} & 19$\pm$5 & 134$\pm$7 & \hl{326$\pm$16} & - \\
 & quadruped & jump & \hl{196$\pm$29} & 185$\pm$37 & 137$\pm$27 & \hl{209$\pm$32} & \hl{282$\pm$27} & 177$\pm$26 & 184$\pm$25 & 57$\pm$14 & 113$\pm$12 & 183$\pm$24 & - \\
 &  & run & 134$\pm$17 & \hl{234$\pm$8} & 88$\pm$13 & 123$\pm$17 & \hl{191$\pm$14} & 125$\pm$14 & \hl{166$\pm$19} & 65$\pm$16 & 99$\pm$9 & 137$\pm$14 & - \\
 &  & stand & \hl{413$\pm$56} & \hl{321$\pm$42} & 220$\pm$29 & 270$\pm$38 & \hl{436$\pm$34} & 231$\pm$52 & 264$\pm$34 & 86$\pm$12 & 215$\pm$47 & 287$\pm$53 & - \\
 &  & walk & 148$\pm$21 & \hl{171$\pm$21} & 122$\pm$10 & 156$\pm$24 & \hl{212$\pm$12} & 135$\pm$16 & \hl{172$\pm$21} & 40$\pm$13 & 100$\pm$26 & \hl{280$\pm$52} & - \\
 & walker & flip & 133$\pm$12 & 346$\pm$6 & \hl{510$\pm$13} & 443$\pm$30 & 456$\pm$10 & \hl{548$\pm$33} & 281$\pm$22 & 78$\pm$21 & \hl{551$\pm$13} & \hl{507$\pm$18} & - \\
 &  & run & 84$\pm$2 & 234$\pm$9 & 259$\pm$18 & \hl{347$\pm$20} & 303$\pm$9 & 280$\pm$22 & 183$\pm$28 & 45$\pm$5 & \hl{391$\pm$15} & \hl{336$\pm$9} & - \\
 &  & stand & 415$\pm$26 & \hl{905$\pm$9} & \hl{910$\pm$13} & 582$\pm$62 & \hl{951$\pm$5} & \hl{937$\pm$5} & 687$\pm$41 & 253$\pm$49 & 874$\pm$23 & 902$\pm$25 & - \\
 &  & walk & 125$\pm$21 & 632$\pm$33 & 839$\pm$17 & 791$\pm$17 & 832$\pm$16 & \hl{883$\pm$30} & 300$\pm$31 & 92$\pm$16 & \hl{867$\pm$12} & \hl{917$\pm$7} & - \\
\hline
 \multirow{13}{*}{RND} & cheetah & run & 64$\pm$3 & 68$\pm$5 & 96$\pm$8 & \hl{183$\pm$26} & 66$\pm$4 & 50$\pm$5 & 6$\pm$1 & \hl{163$\pm$14} & 138$\pm$17 & \hl{247$\pm$9} & - \\
 &  & run-backward & 96$\pm$6 & \hl{162$\pm$18} & \hl{160$\pm$22} & 60$\pm$4 & 143$\pm$17 & 90$\pm$7 & 2$\pm$0 & 124$\pm$13 & 82$\pm$8 & \hl{185$\pm$17} & - \\
 &  & walk & 289$\pm$22 & 337$\pm$28 & 401$\pm$40 & \hl{567$\pm$70} & 286$\pm$30 & 330$\pm$65 & 29$\pm$11 & \hl{622$\pm$28} & 446$\pm$36 & \hl{827$\pm$41} & - \\
 &  & walk-backward & 469$\pm$38 & \hl{542$\pm$70} & \hl{743$\pm$60} & 345$\pm$59 & \hl{572$\pm$100} & 499$\pm$40 & 14$\pm$1 & 517$\pm$60 & 352$\pm$45 & \hl{793$\pm$66} & - \\
 & maze & reach & 9$\pm$2 & 4$\pm$1 & 4$\pm$1 & 8$\pm$4 & 8$\pm$4 & \hl{707$\pm$12} & \hl{759$\pm$7} & \hl{736$\pm$3} & 532$\pm$20 & \hl{710$\pm$8} & - \\
 & quadruped & jump & \hl{770$\pm$7} & 474$\pm$40 & 176$\pm$13 & 663$\pm$20 & \hl{806$\pm$14} & 490$\pm$48 & 447$\pm$31 & 326$\pm$72 & \hl{731$\pm$9} & 651$\pm$8 & - \\
 &  & run & \hl{465$\pm$4} & 415$\pm$10 & 98$\pm$13 & 418$\pm$10 & 478$\pm$8 & 399$\pm$26 & 301$\pm$12 & 263$\pm$40 & \hl{461$\pm$6} & \hl{429$\pm$3} & - \\
 &  & stand & \hl{919$\pm$19} & 770$\pm$28 & 426$\pm$45 & 830$\pm$31 & \hl{973$\pm$2} & 720$\pm$27 & 552$\pm$32 & 529$\pm$70 & \hl{944$\pm$11} & 815$\pm$2 & - \\
 &  & walk & \hl{586$\pm$28} & 486$\pm$37 & 90$\pm$13 & \hl{527$\pm$21} & \hl{552$\pm$33} & 410$\pm$30 & 310$\pm$14 & 210$\pm$42 & 516$\pm$30 & \hl{528$\pm$10} & - \\
 & walker & flip & 267$\pm$28 & 332$\pm$18 & 461$\pm$9 & 34$\pm$3 & 399$\pm$19 & \hl{569$\pm$30} & \hl{512$\pm$27} & 50$\pm$9 & 454$\pm$23 & \hl{578$\pm$10} & - \\
 &  & run & 96$\pm$8 & 167$\pm$12 & 251$\pm$11 & 47$\pm$22 & 250$\pm$6 & 299$\pm$20 & \hl{325$\pm$10} & 38$\pm$5 & \hl{350$\pm$14} & \hl{388$\pm$8} & - \\
 &  & stand & 516$\pm$36 & 733$\pm$22 & 813$\pm$10 & 191$\pm$62 & \hl{853$\pm$18} & 836$\pm$38 & \hl{904$\pm$21} & 326$\pm$41 & 828$\pm$20 & \hl{890$\pm$15} & - \\
 &  & walk & 152$\pm$35 & 457$\pm$33 & 518$\pm$74 & 30$\pm$2 & 607$\pm$31 & \hl{748$\pm$75} & \hl{818$\pm$36} & 53$\pm$10 & \hl{853$\pm$15} & \hl{760$\pm$19} & - \\
	   \bottomrule
	   \end{tabular}
	}
\caption{Score of each method, split by task and replay buffer. Average
over ten random seeds, with $\pm 1\sigma$ estimated standard deviation on this
average estimator. For Maze, we report the average over the 20 goals
defined in the environment. We highlight the three leading methods (four when confidence intervals overlap) for each task.}
\end{table}

\newpage
\subsection{Aggregate Plots of Results}
\label{sec:aggplots}

Here we plot the results, first averaged over everything, then by
environment averaged over the tasks of that environment, and finally by task.

\begin{figure}[h]
\begin{center}
\includegraphics[width=.7\textwidth]{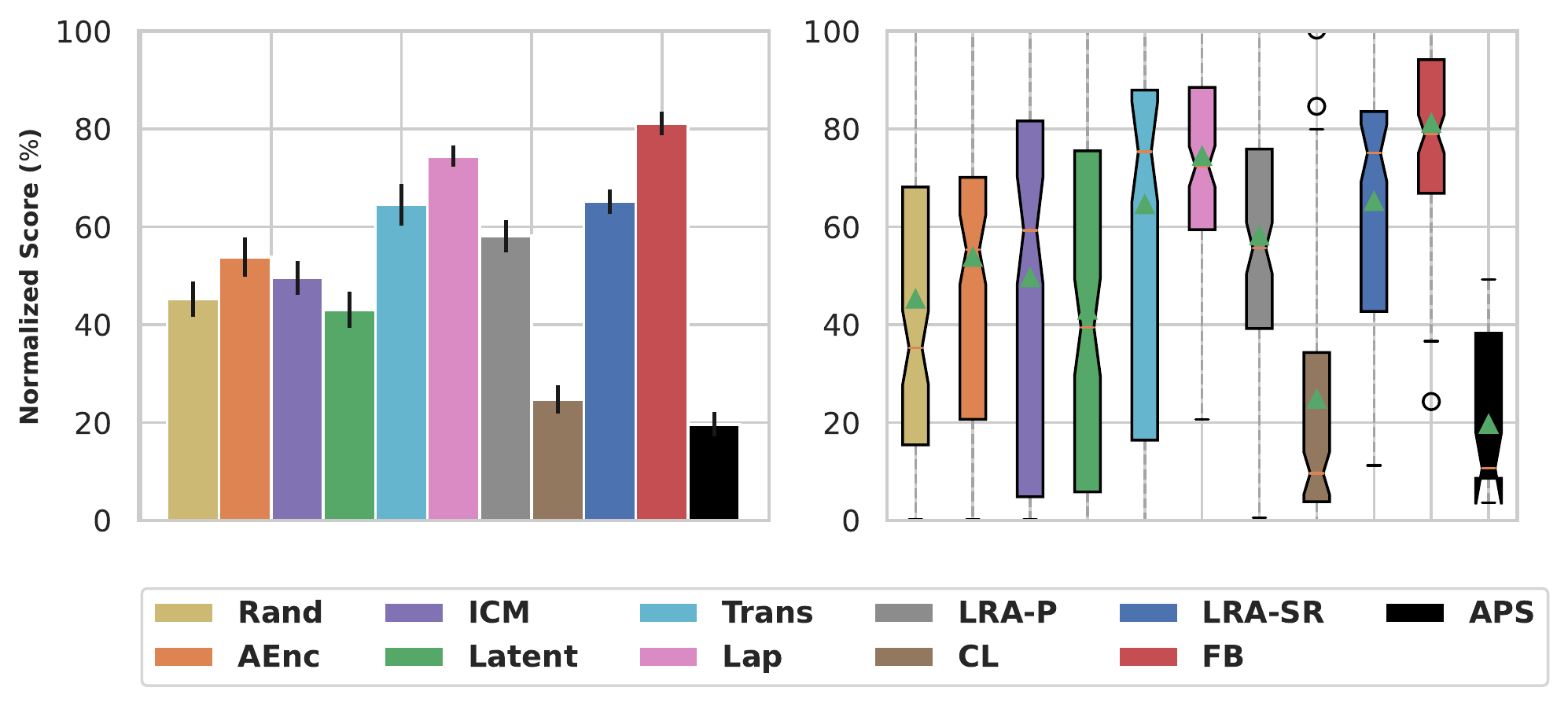}
\end{center}
\caption{Zero-shot scores of ten SF methods and FB, aggregated over tasks
using normalized scores as described in the text. To assess variability,
the box plot on the
right shows the variations of the distribution of normalized scores over
random seeds, environments, and replay buffers.
\label{fig:global_full}
}
\end{figure}

\begin{figure}[h]
\begin{center}
\includegraphics[width=.9\textwidth]{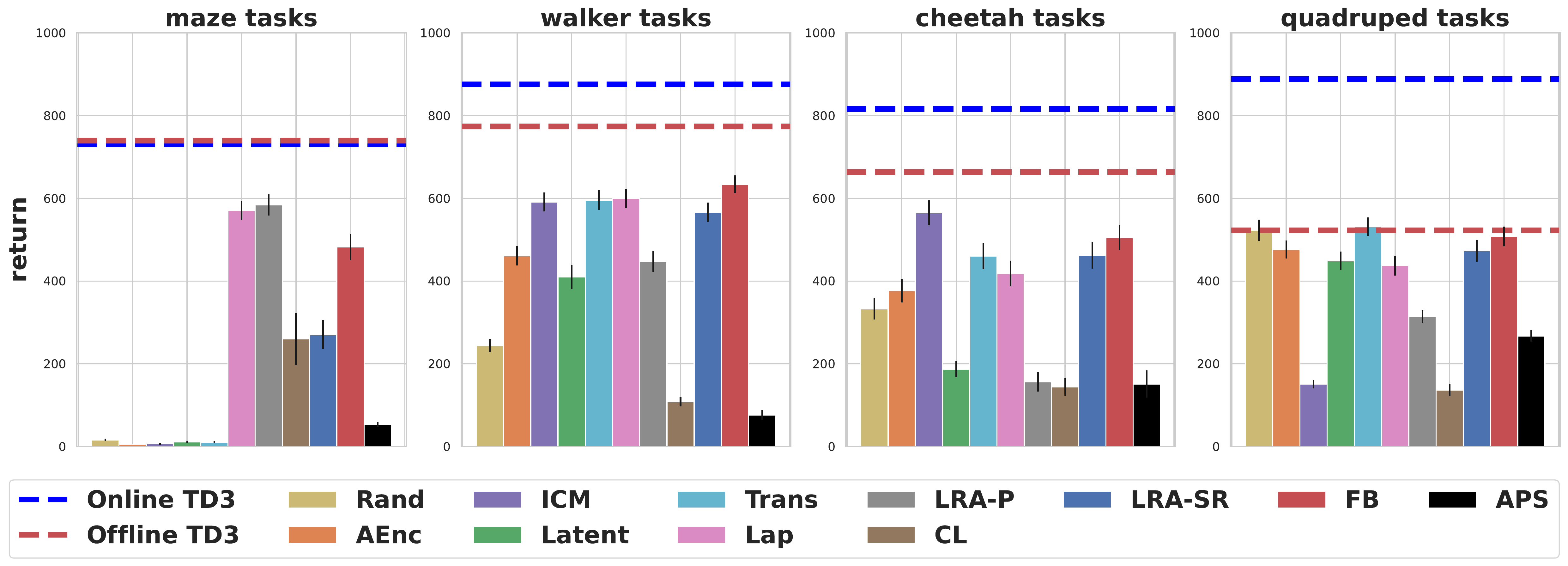}
\end{center}
\caption{Zero-shot scores averaged over tasks for each environment, with supervised online and
offline TD3 as toplines. Average over 3 replay buffers and 10 random
seeds.
\label{fig:envscores}
}
\end{figure}

\begin{figure}[h!]
\begin{center}
\includegraphics[width=.95\textwidth]{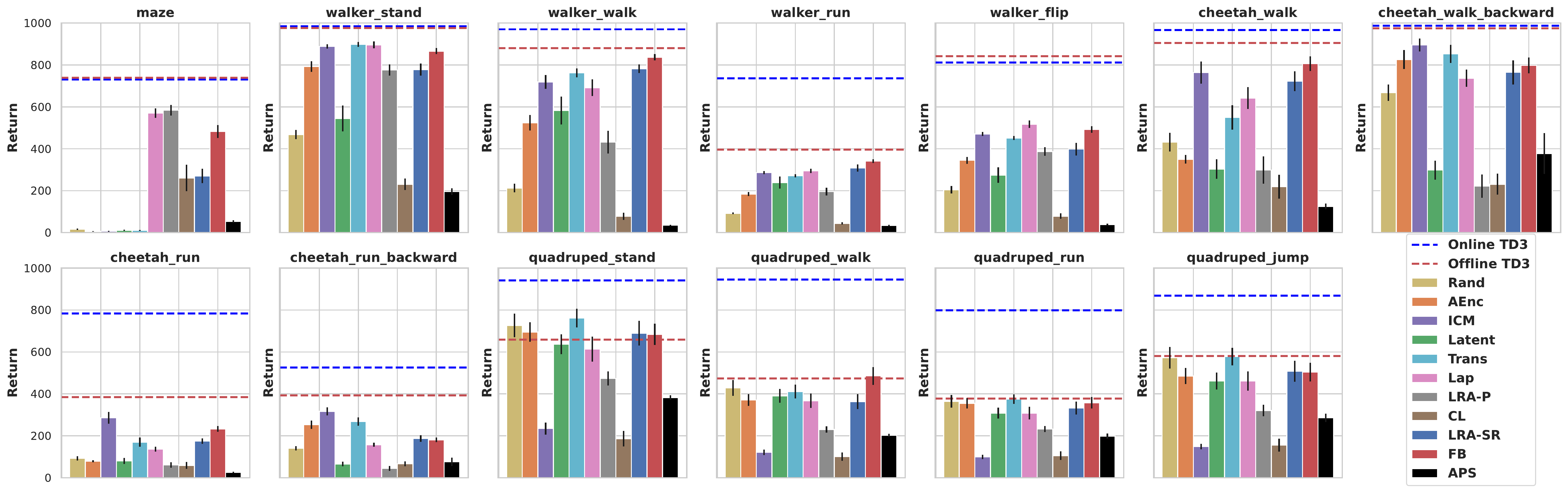}
\end{center}
\caption{Zero-shot scores for each task, with supervised online and
offline TD3 as toplines. Average over 3 replay buffers and 10 random
seeds.
}
\end{figure}

\newpage

\newpage
\subsection{Full Plots of Results per Task and Replay Buffer}
\label{sec:fullplots}

\begin{figure}[h]
\begin{center}
\includegraphics[width=.95\textwidth]{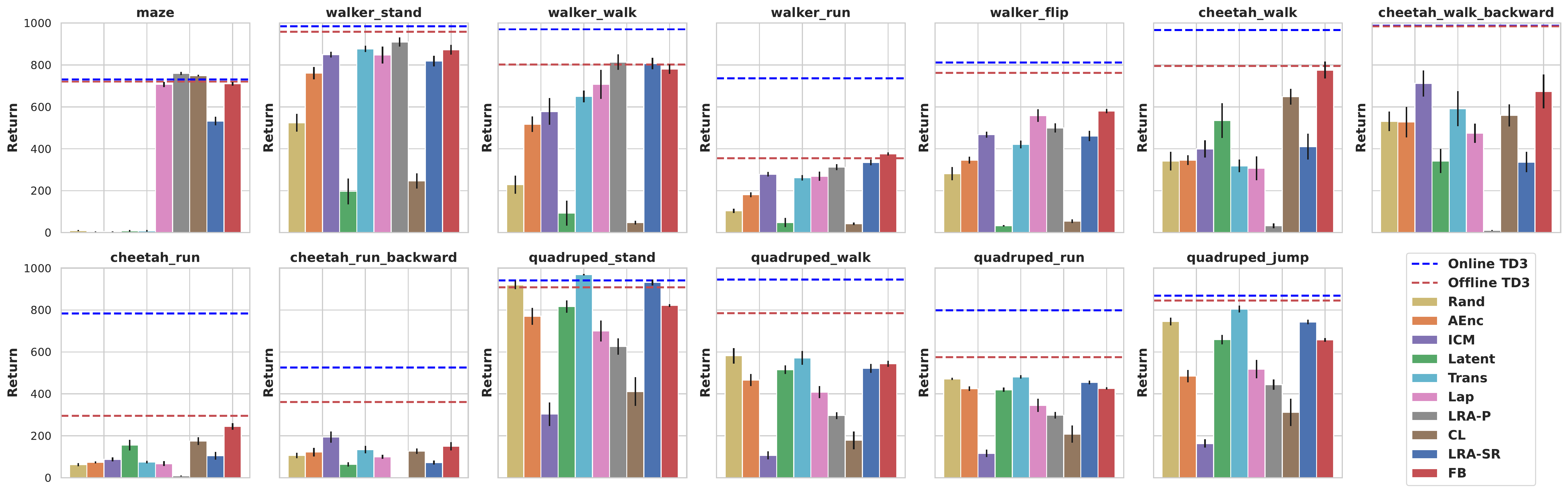}
\caption{Per-task results on the RND replay buffer, average over 10
random seeds.
\label{fig:rnd_per_task}
}
\end{center}
\end{figure}

\begin{figure}[h]
\begin{center}
\includegraphics[width=.95\textwidth]{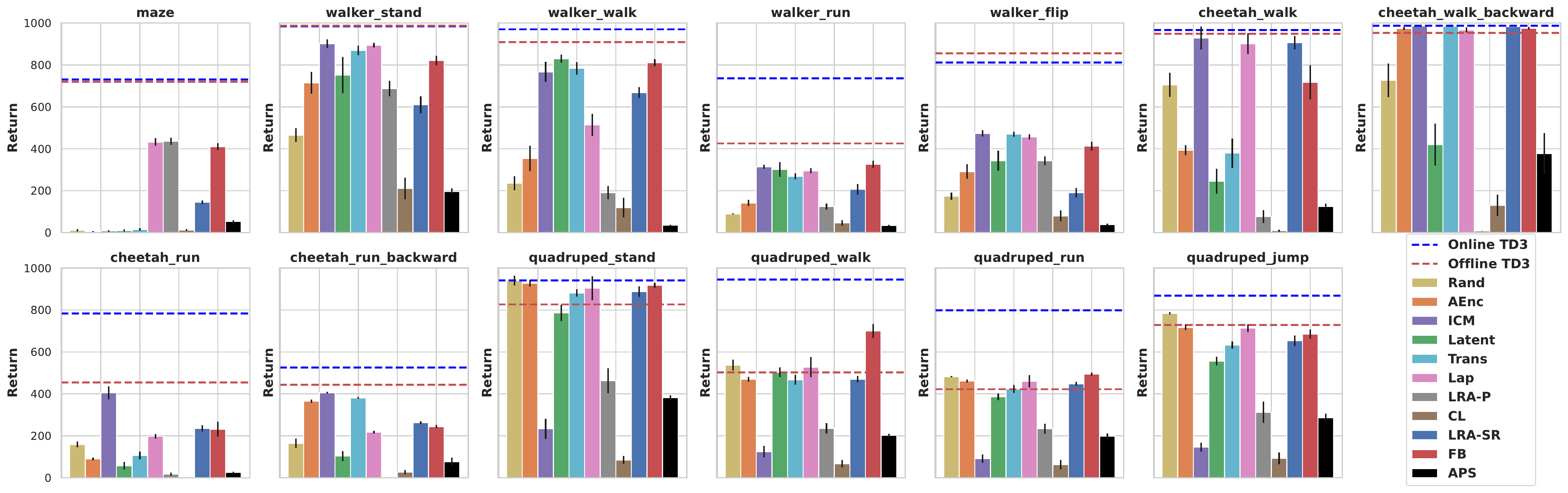}
\caption{Per-task results on the APS replay buffer, average over 10 random seeds.
\label{fig:aps_per_task}
}
\end{center}
\end{figure}

\begin{figure}[h]
\begin{center}
\includegraphics[width=.95\textwidth]{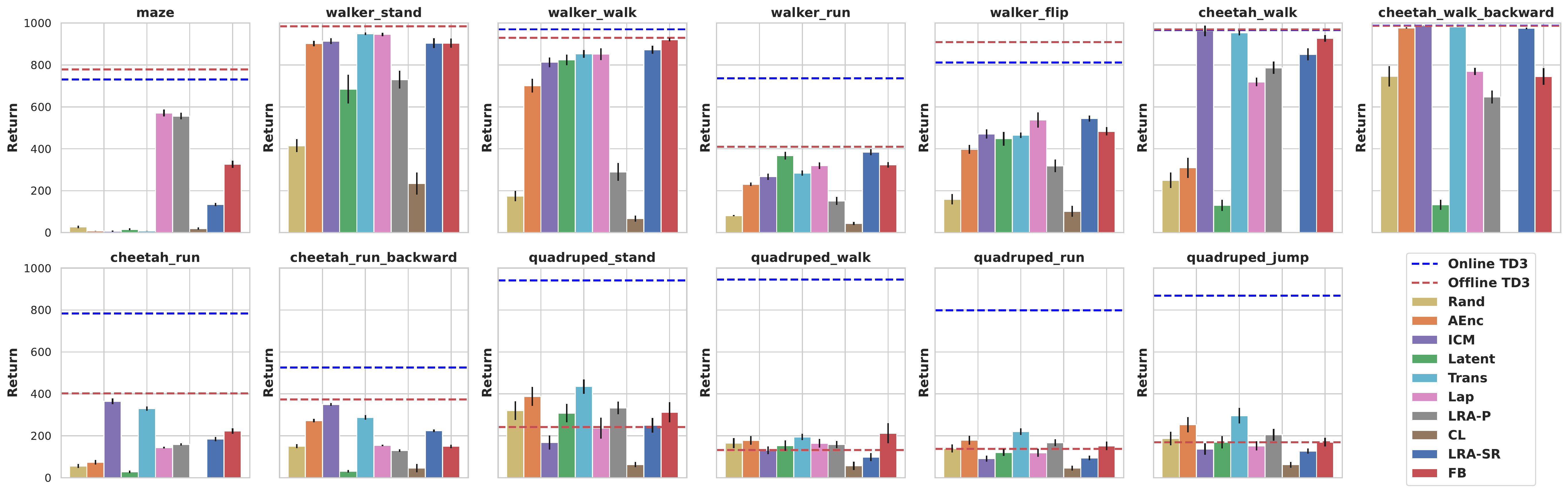}
\caption{Per-task results on the Proto replay buffer, average over 10 random seeds.
\label{fig:proto_per_task}
}
\end{center}
\end{figure}

\newpage
\subsection{Influence of the Replay Buffer}
\label{sec:buffer}

\begin{figure}[h]
\begin{center}
\includegraphics[width=.85\textwidth]{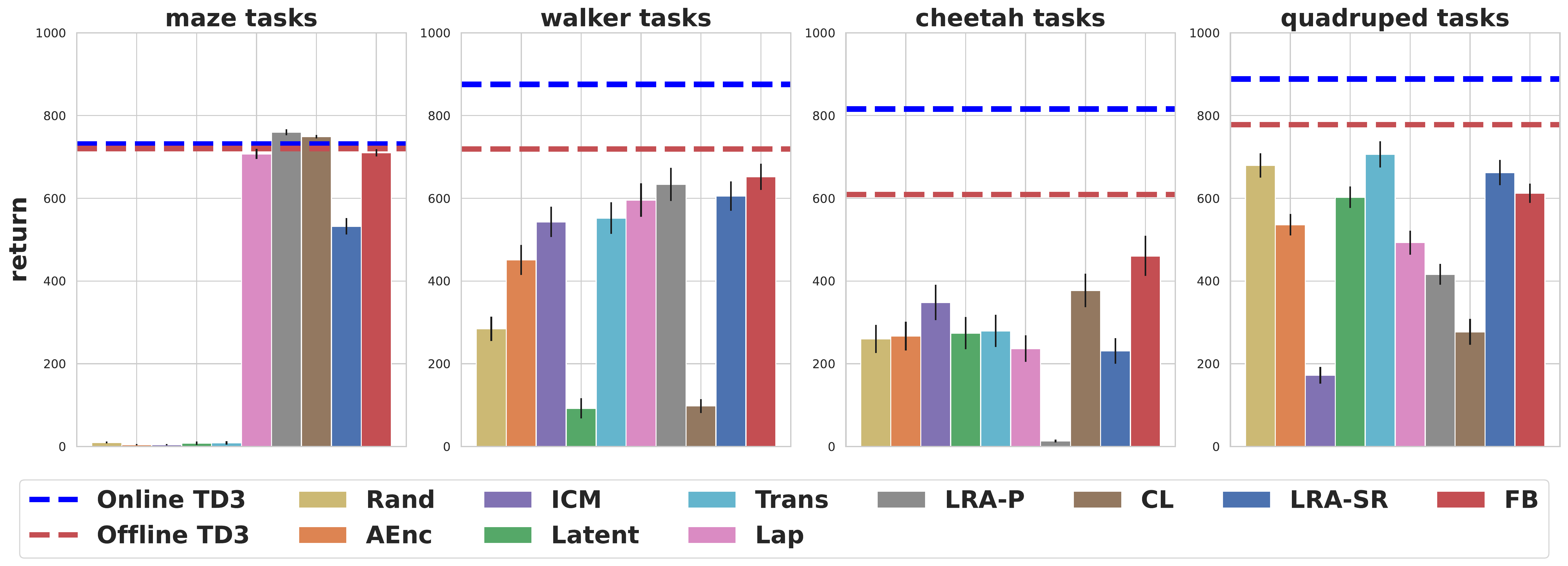}

\bigskip

\includegraphics[width=.85\textwidth]{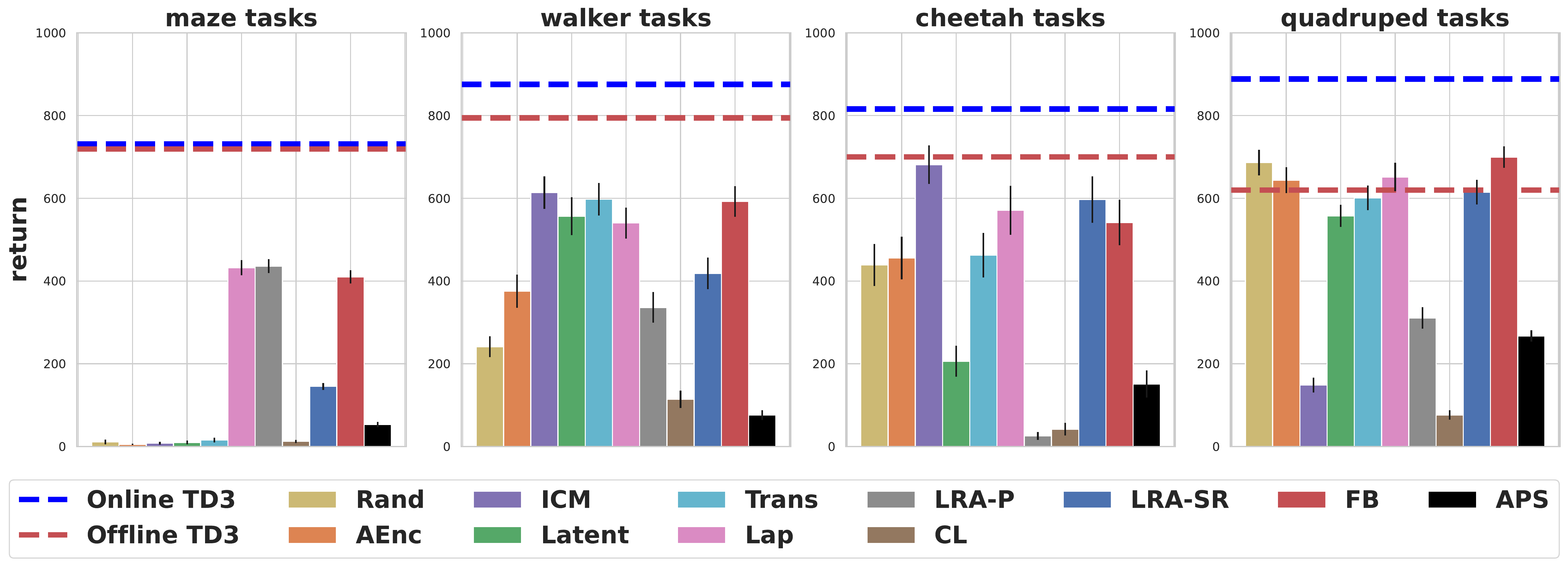}

\bigskip

\includegraphics[width=.85\textwidth]{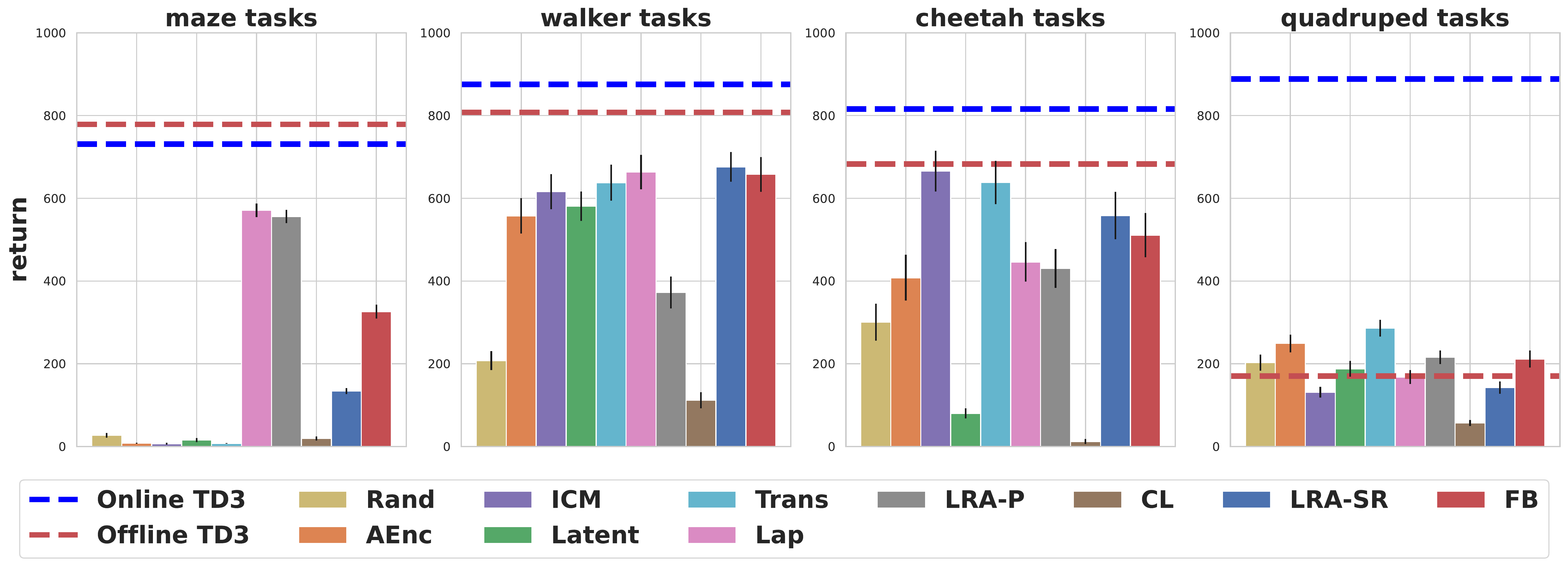}
\caption{Results on each replay buffer: RND (top), APS (middle), Proto
(bottom). Average over 4 tasks for the
Walker, Cheetah and Quadruped environments, average over 20 goals for Maze.
\label{fig:RNDbuffer}
\label{fig:APSbuffer}
\label{fig:protobuffer}
}
\end{center}
\end{figure}

Here we plot the influence of the replay buffer, by reporting results
separated by replay buffer, but averaged over the tasks corresponding to
each environment
(Fig~\ref{fig:RNDbuffer}).

Overall, there is a clear failure case of the Proto buffer on the
Quadruped environment: the TD3 supervised baseline performs poorly for all
tasks in that environment.

Otherwise, results are broadly consistent on the different replay
buffers: with a few exceptions, the same methods succeed or fail on the same
environments.

%
%

\newpage

\begin{figure}[h]
	\begin{center}
	\includegraphics[width=.55\textwidth]{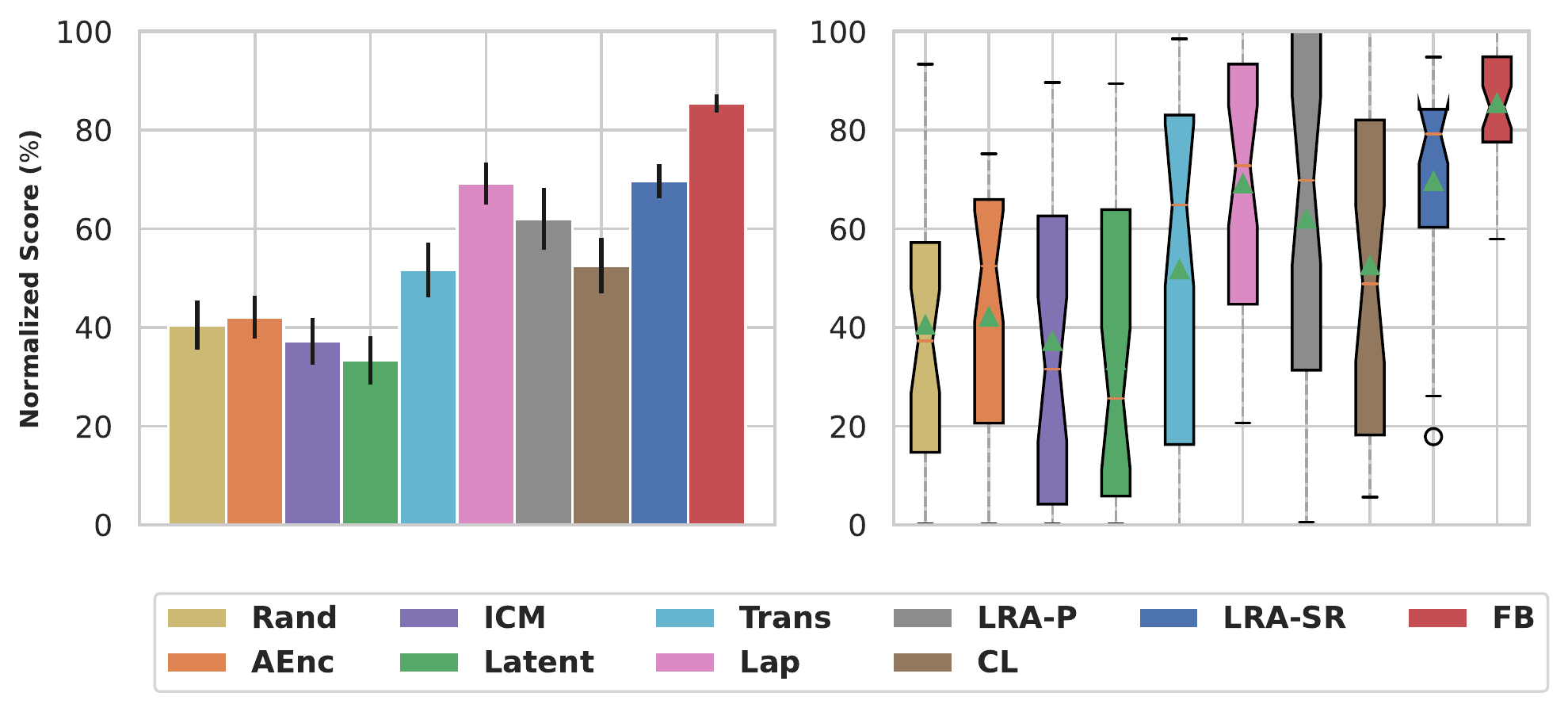}

	\bigskip
	\includegraphics[width=.55\textwidth]{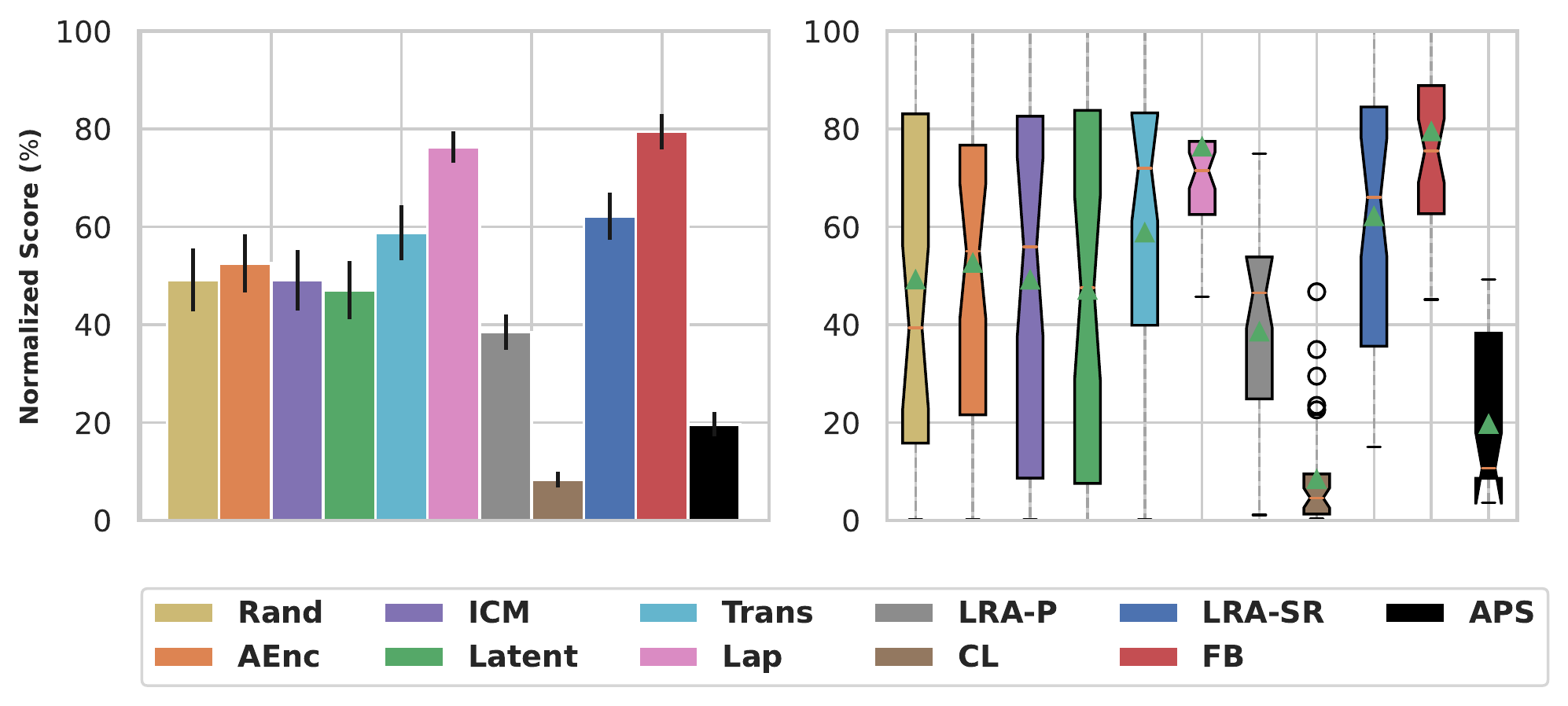}

	\bigskip
	\includegraphics[width=.55\textwidth]{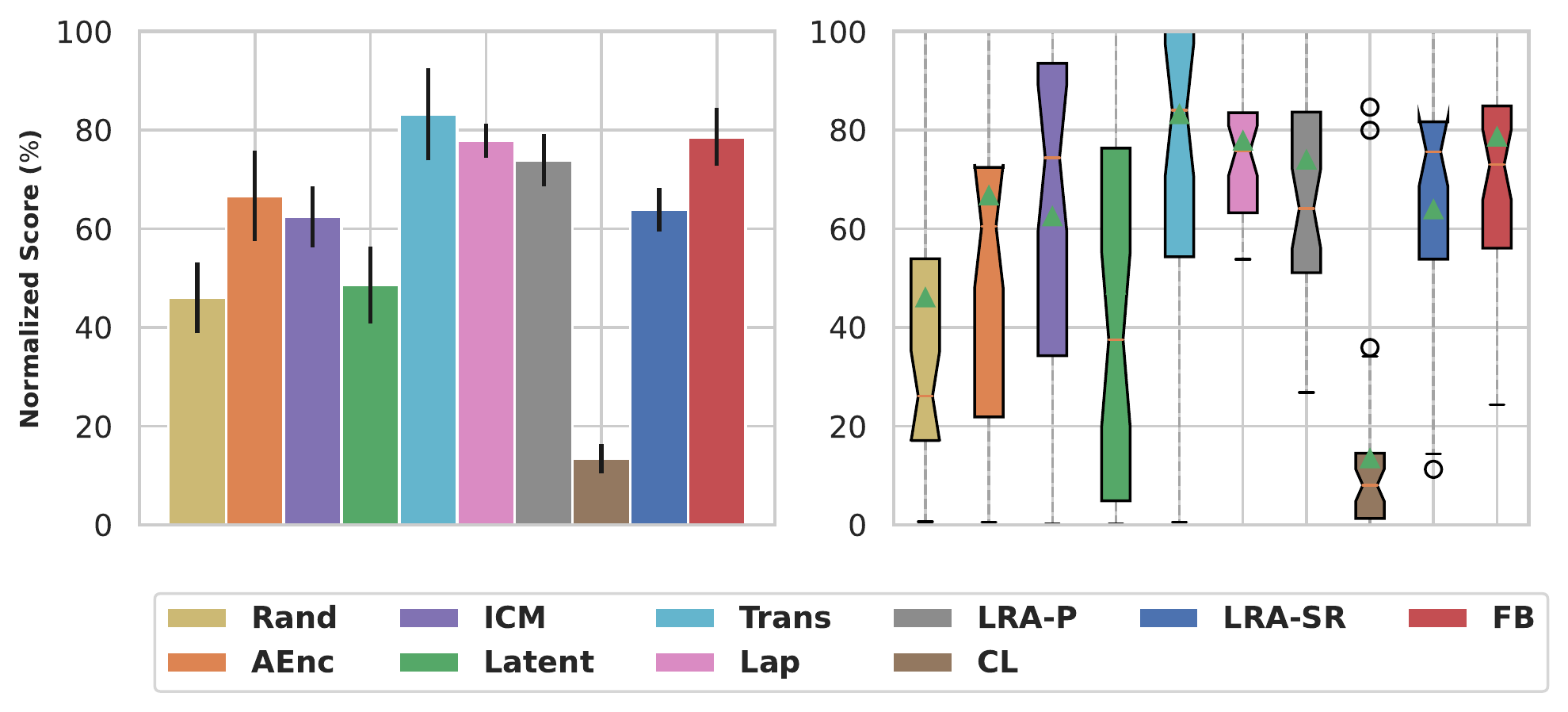}
	\end{center}
	\caption{Zero-shot scores by replay buffer,
	as a percentage of
	the supervised score of offline TD3 trained on the same buffer, 
	averaged over tasks and environments and random seeds. Top: RND
	buffer;
	middle: APS buffer; bottom: Proto buffer.}
	\label{fig:rnd_global_score}
	\label{fig:aps_global_score}
	\label{fig:proto_global_score}
\end{figure}

In
Fig.~\ref{fig:rnd_global_score},
we plot results aggregated over all tasks but split by replay buffer. Box plots
further show variability within random seeds and environments for a given
replay buffer.

Overall
method rankings are broadly consistent between RND and APS, except for
\CL. Note that the aggregated
normalized score for Proto is largely influenced by the failure on
Quadruped: normalization by a very low baseline (Fig.~\ref{fig:protobuffer},
Quadruped plot for Proto),
somewhat artificially
 pushes Trans high
up (this shows the limit of using scores normalized by offline TD3 score),
while the other methods' rankings are more similar to RND and APS.

\FB and \lap
work very well in all replay buffers, with \LRASR and \Transition a bit
behind due to their feailures on some tasks.

%
%
%

\newpage
\section{Analysis of The Learned Features}
\label{sec:features}

\subsection{Feature Rank}

Here we test the hypothesis that feature collapse for some methods is
responsible for some cases of bad performance. This is especially
relevant for Maze, where some methods may have little incentive to learn
more features beyond the original two features $(x,y)$.

We report in Table~\ref{tab:feature_rank} the effective rank of learned
features $B$ or $\phi$: this is computed as in \cite{lyle2022understanding}, as the
fraction of eigenvalues of
 $\E_{s\sim \rho} B(s)\transp{B(s)}$ or $\E_{s\sim \rho}
 \phi(s)\transp{\phi(s)}$ above a certain threshold. 

 \begin{table}[h]
	\centering
    \footnotesize	
	\begin{tabular}{lllllllllll}
	\toprule
	          \textbf{Domain}  & & & & & & \textbf{Method} \\
		  &  \Rand & \AEnc & \ICM & \Latent & \Transition & \lap & \LRAP & \CL & \LRASR & \FB  \\
	   \toprule
	   Maze & 0.38 &  0.31 & 0.33 & 0.32 & 0.15 & 1.0 & 1.0 & 1.0 & 1.0 & 1.0 \\
	   Walker & 1,0 &  0.91 & 1.0 & 1.0 & 0.90 & 1.0 & 1.0 & 0.65 & 1.0 & 1.0 \\
	   Cheetah & 1.0 &  0.96 & 1.0 & 1.0 & 1.0 & 1.0 & 1.0 & 1.0 & 1.0 & 1.0 \\
	   Quadruped &  1.0 &  0.98 & 0.30 & 1.0 & 0.15 & 1.0 & 0.84 & 1.0 & 1.0 & 1.0 \\

	   \bottomrule
	   \end{tabular}

\caption{Feature rank~\cite{lyle2022understanding} of $\phi$ or $B$ for
each method, computed as
 $\frac1d \,\#\left\{  \sigma \in \texttt{eig}\left(\frac{1}{n} \sum_{i=1}^n
 \phi(s_i) \phi(s_i)^\top\right)  \mid \sigma > \epsilon \right\}$, trained on RND replay buffer and averaged over 10
 random seeds. We use $n=100,000$ samples to estimate the covariance, and
 $\epsilon=10^{-4}$.
	\label{tab:feature_rank}}
\end{table}

For most methods and all environments except Maze, more than $90\%$
(often $100\%$) of eigenvalues are above $10^{-4}$,
 so the effective rank is close to full.

The Maze environment is a clear exception: on Maze, for the \Rand, \AEnc, \Transition, \Latent and \ICM methods,
 only about one third of the eigenvalues are above $10^{-4}$.
The other methods keep $100 \%$ of the eigenvalues above $10^{-4}$.
This is perfectly aligned with the performance of each method on Maze.

So rank reduction does happen for some methods.  This reflects
the fact that two features $(x,y)$ already convey the necessary
information about states and dynamics, but are not sufficient to solve
the problem via successor features.
In the Maze environment, auto-encoder or transition models can perfectly
optimize their loss just by keeping the original two features $(x,y)$,
and they have no incentive to learn other features, so the effective rank
could have been 2.

These methods may benefit from auxiliary losses to prevent eigenvalue
collapse, similar to the orthonormalization loss used for $B$. We did not
include such losses, because we wanted to keep the same methods
(autoencoders, ICM, transition model...) used in the literature.  But
even with an auxiliary loss, keeping a full rank could be achieved just
by keeping $(x,y)$ and then blowing up some additional irrelevant
features.

\newpage
\subsection{Embedding Visualization via t-SNE}

\begin{figure}[h]
	\label{fig: tsne}
	\begin{center}
	\includegraphics[width=.27\textwidth]{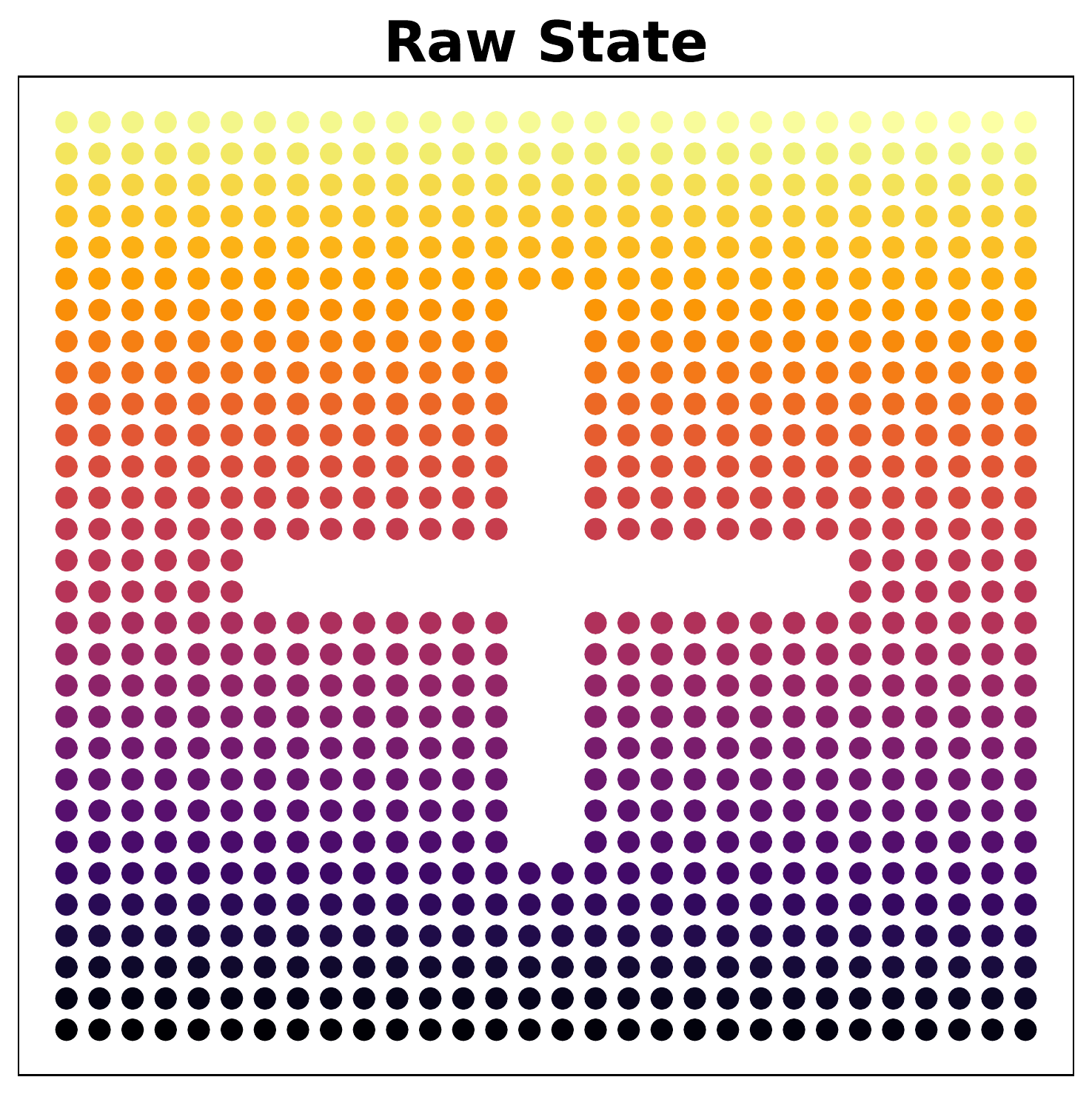}
	\includegraphics[width=.27\textwidth]{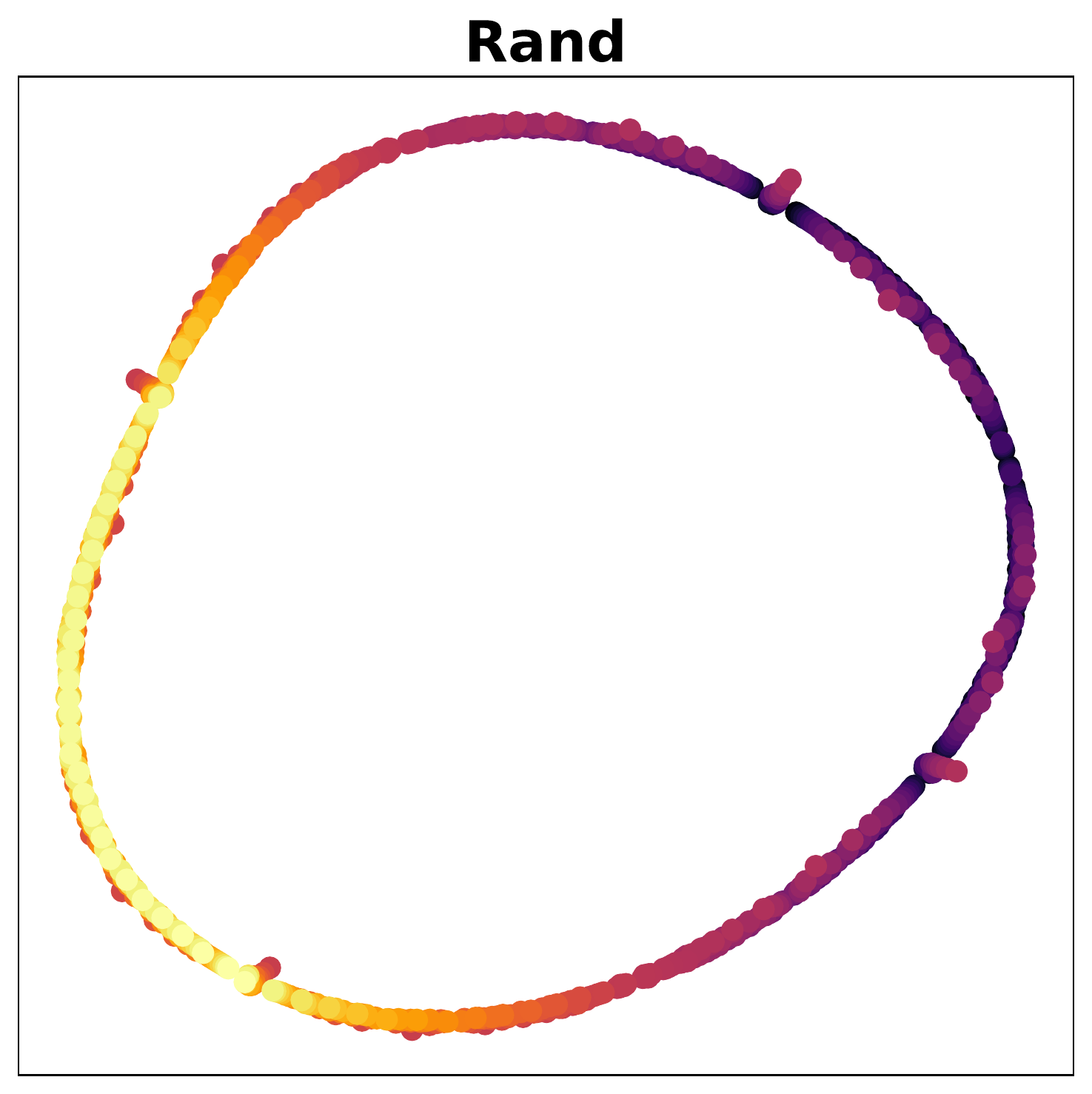}
	\includegraphics[width=.27\textwidth]{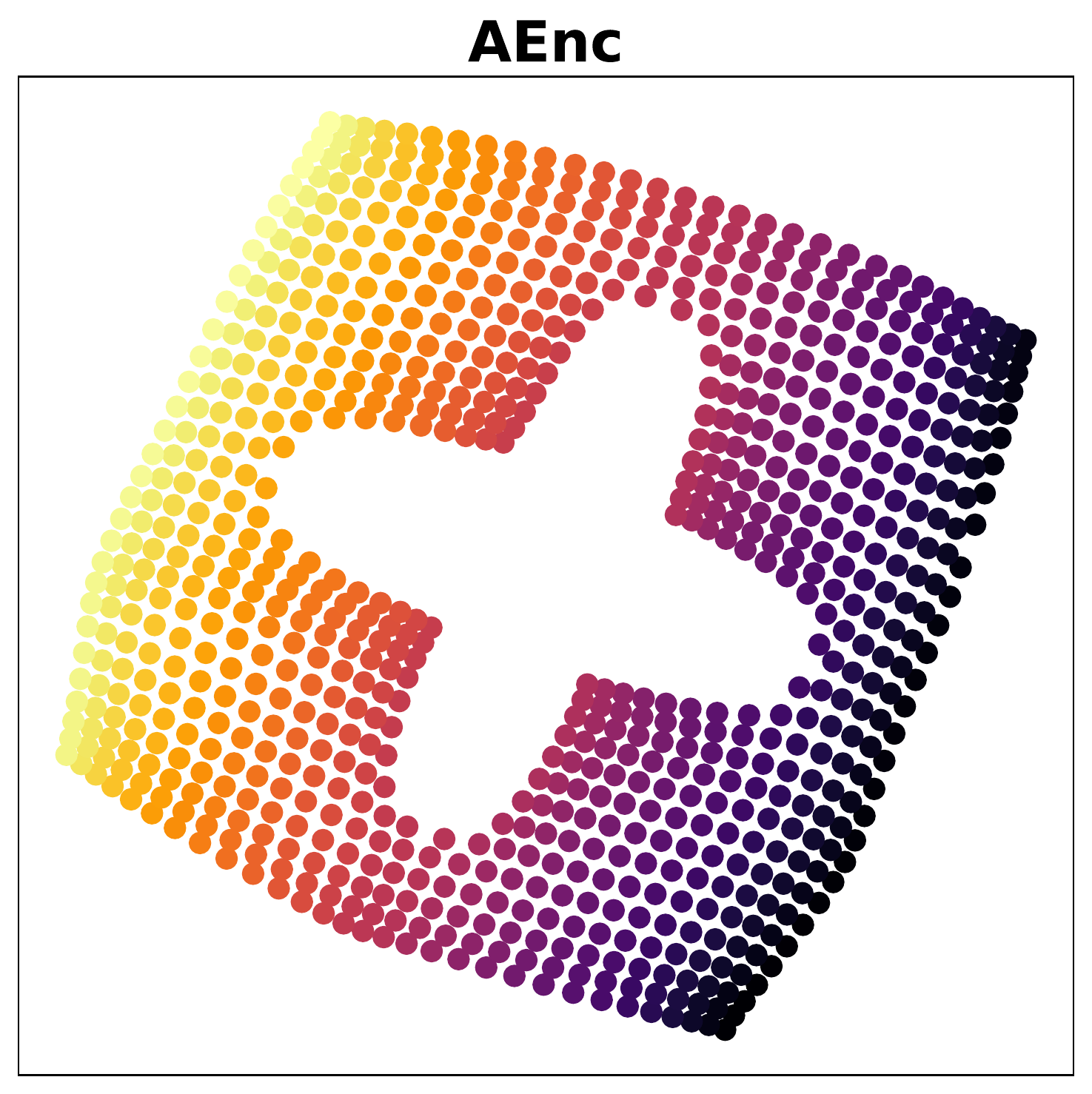}
	\includegraphics[width=.27\textwidth]{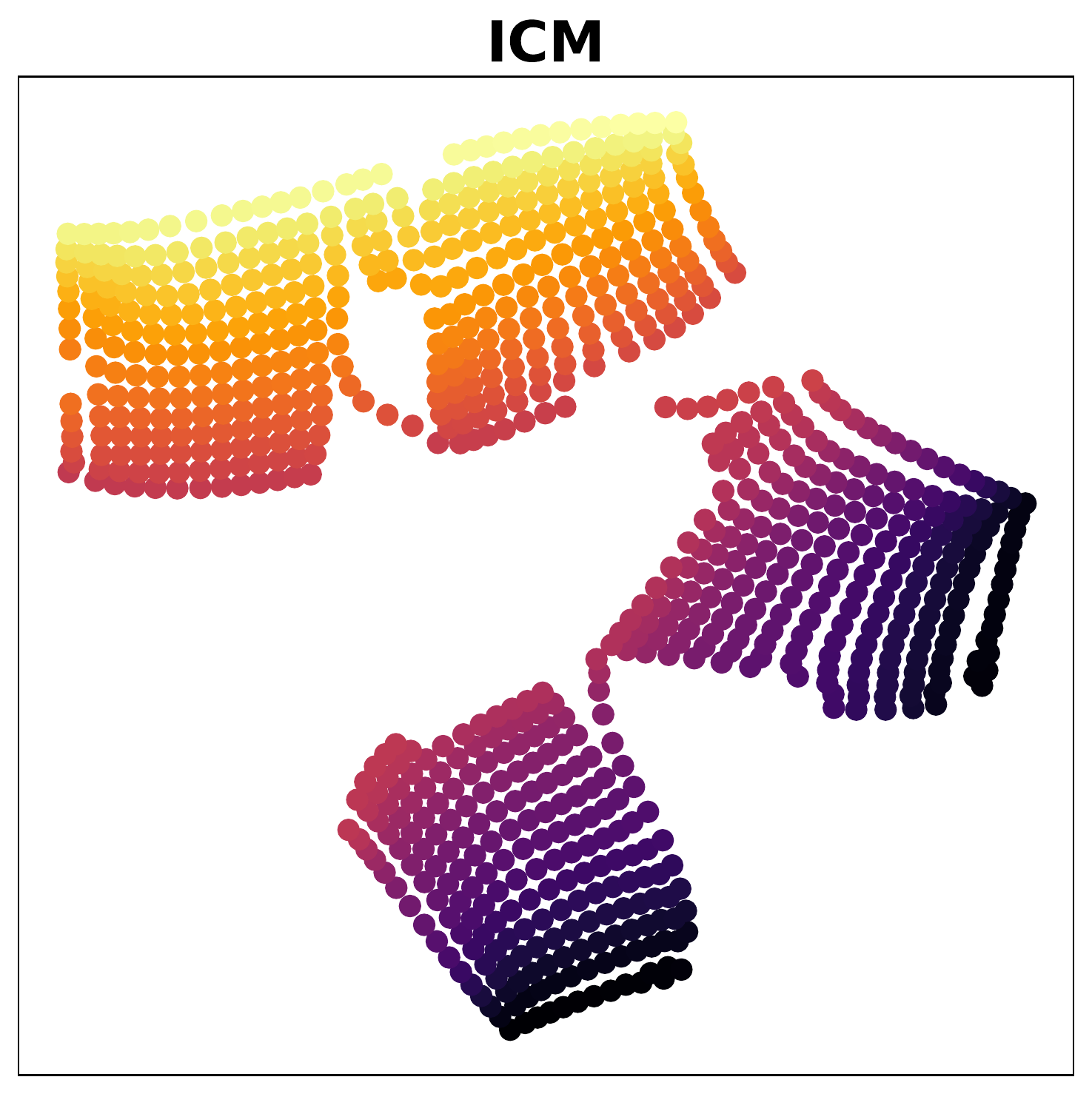}
	\includegraphics[width=.27\textwidth]{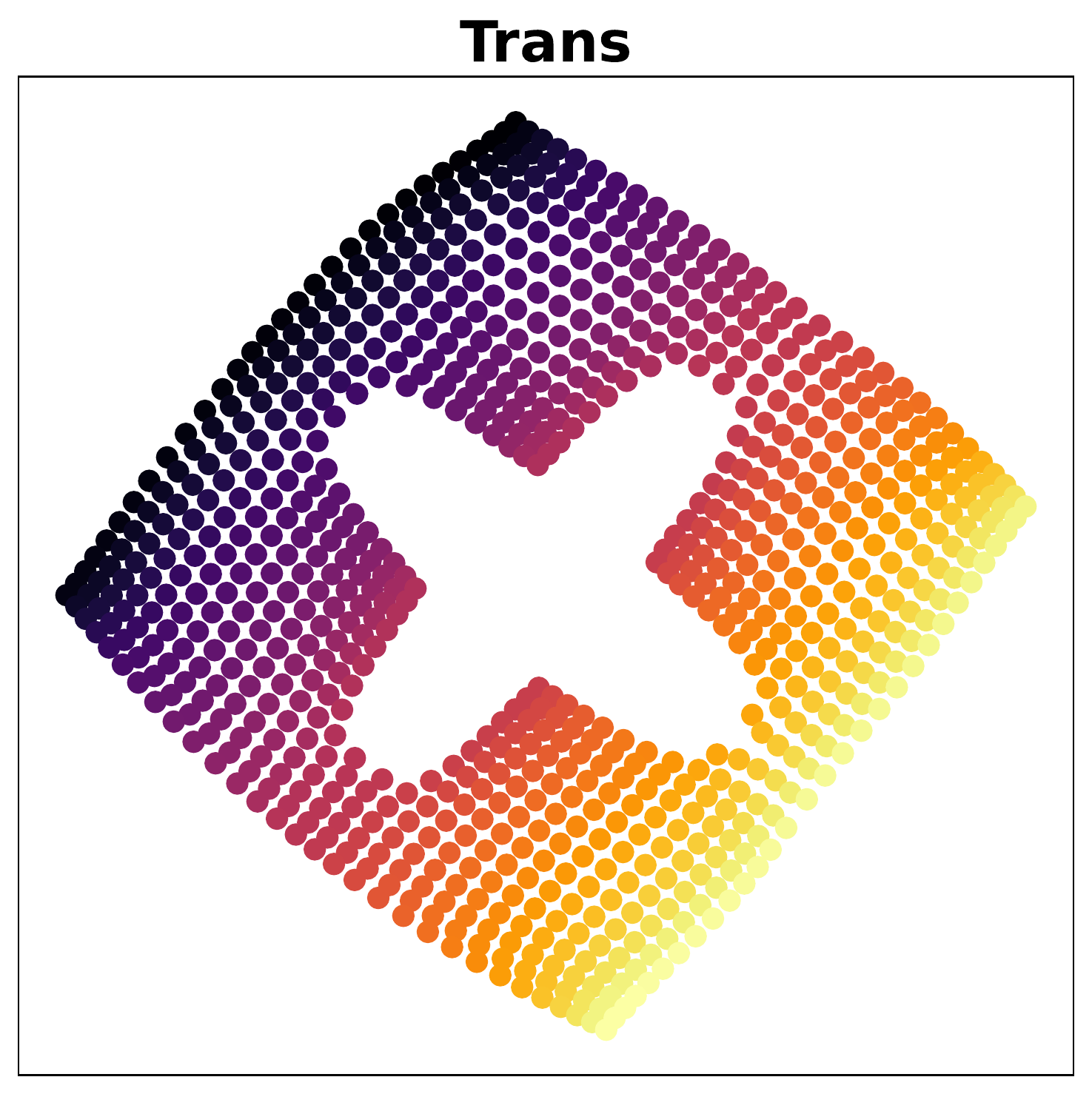}
	\includegraphics[width=.27\textwidth]{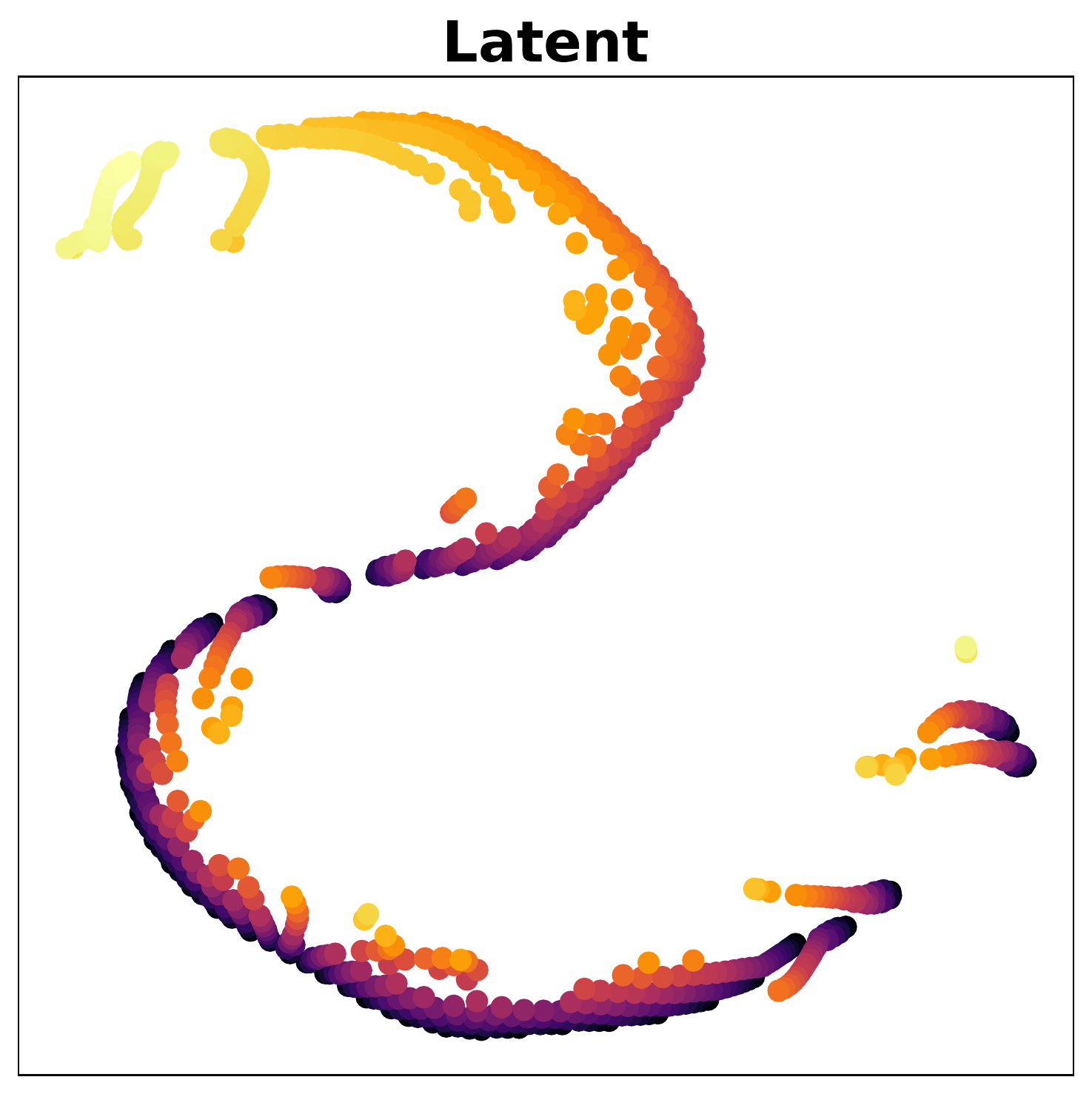}
	\includegraphics[width=.27\textwidth]{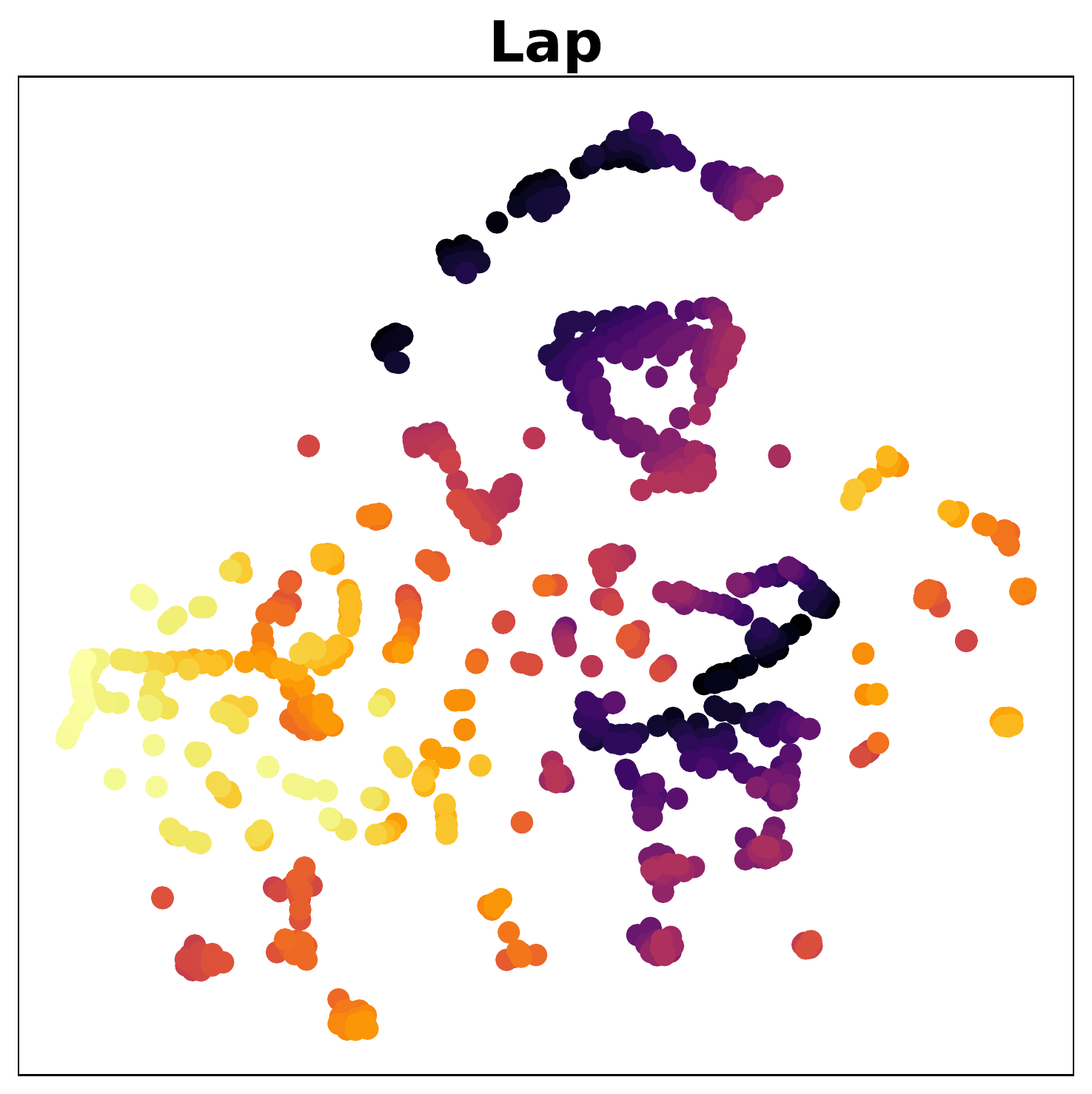}
	\includegraphics[width=.27\textwidth]{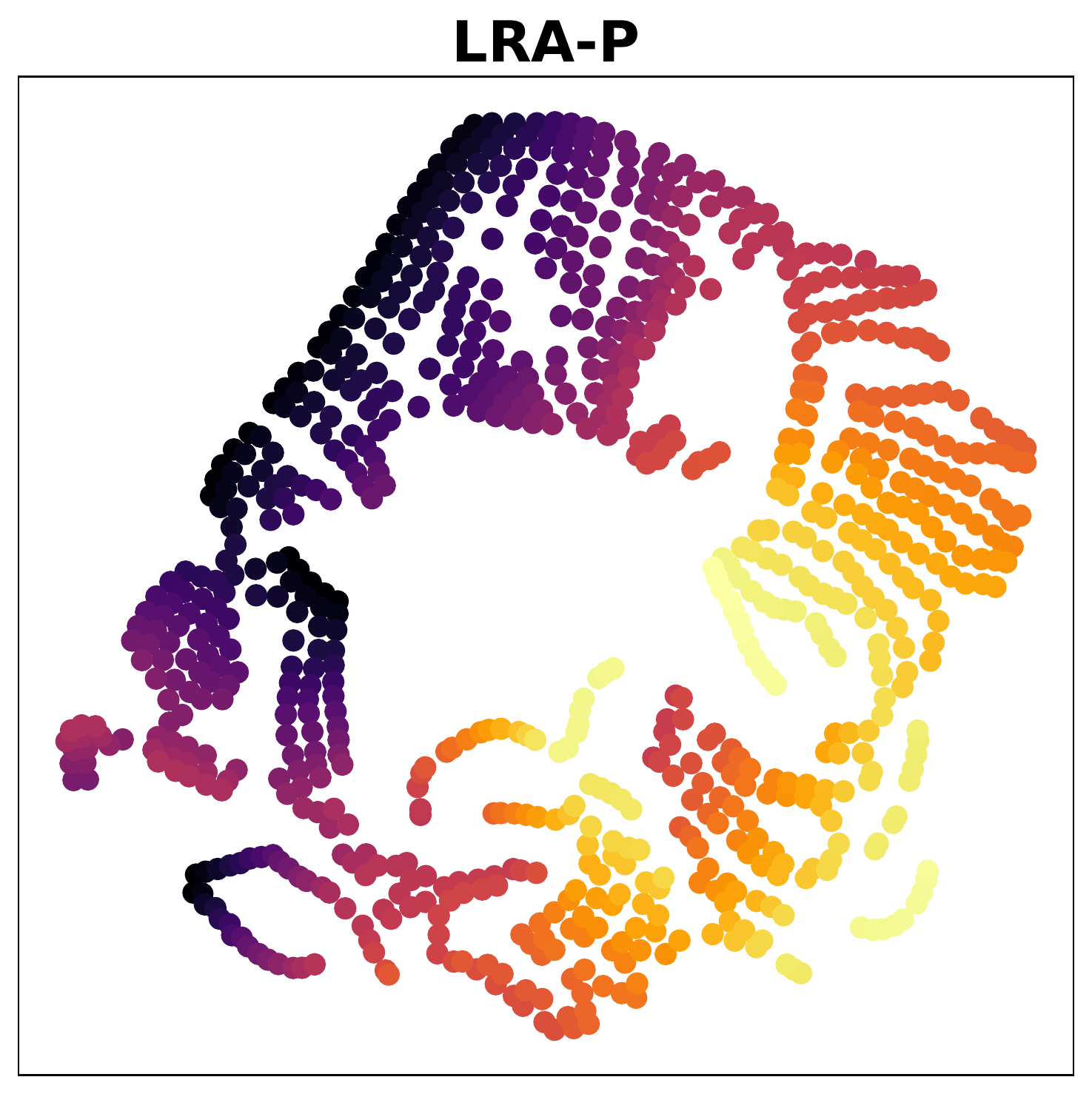}
	\includegraphics[width=.27\textwidth]{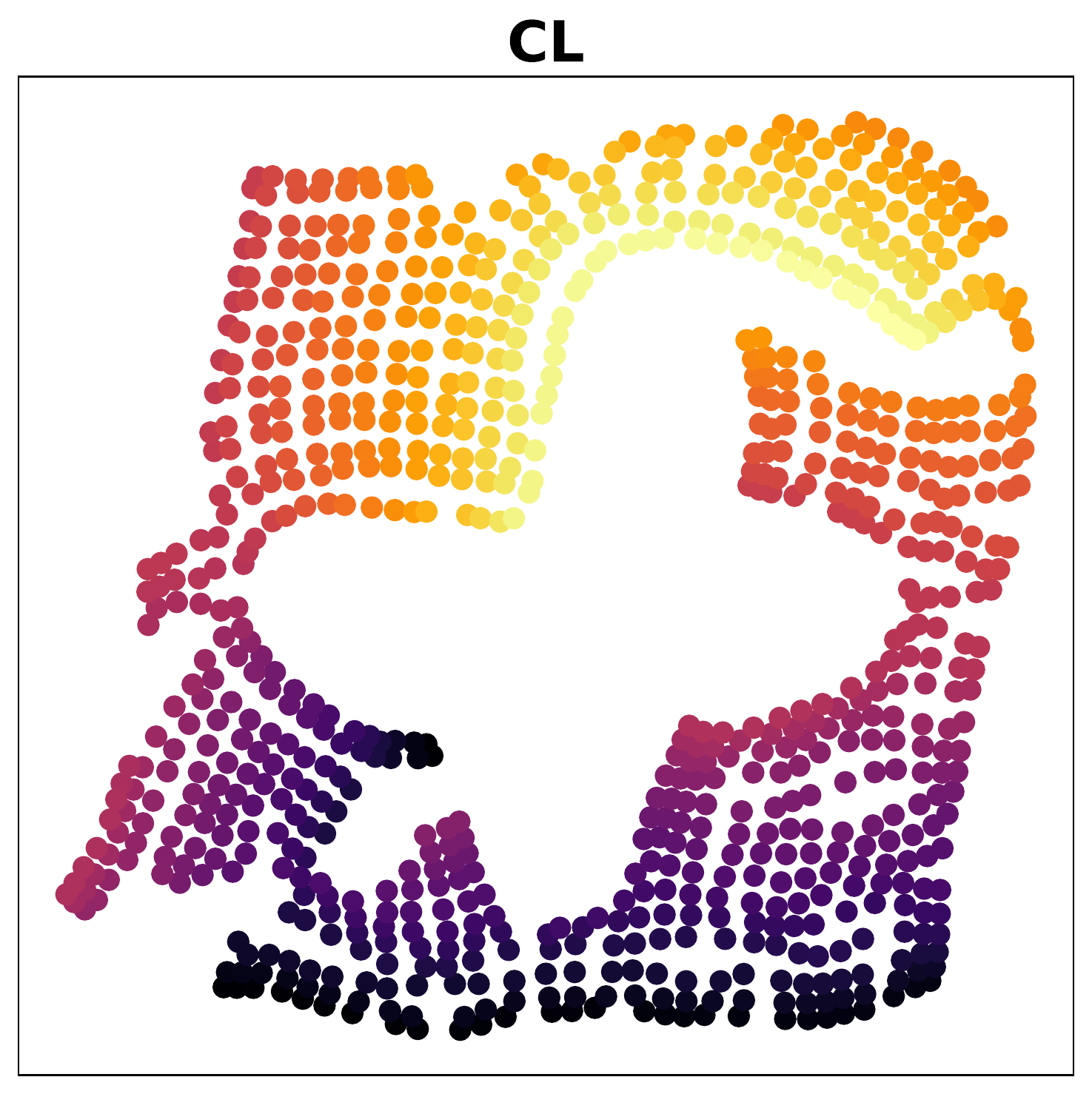}
	\includegraphics[width=.27\textwidth]{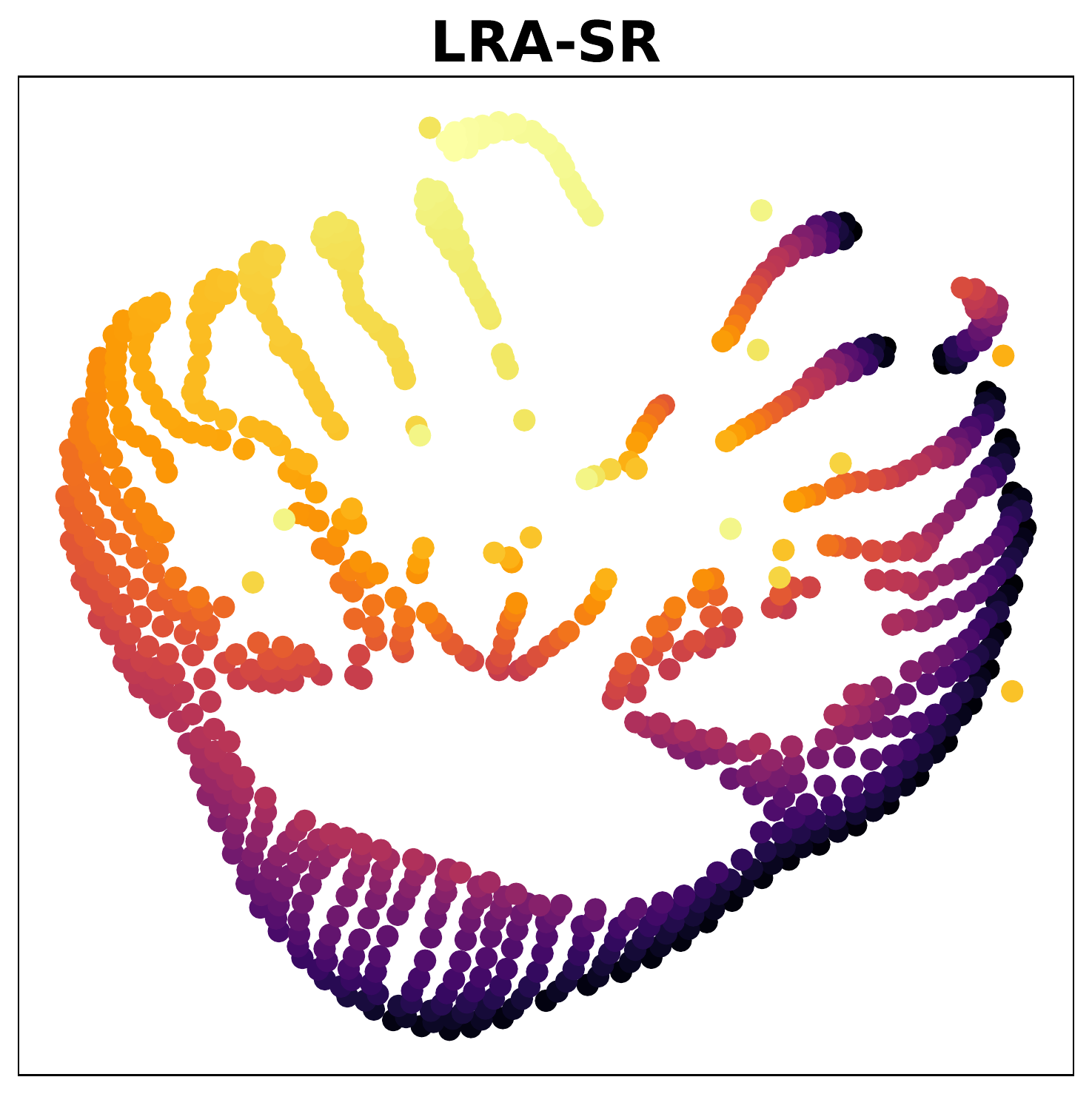}
	\includegraphics[width=.27\textwidth]{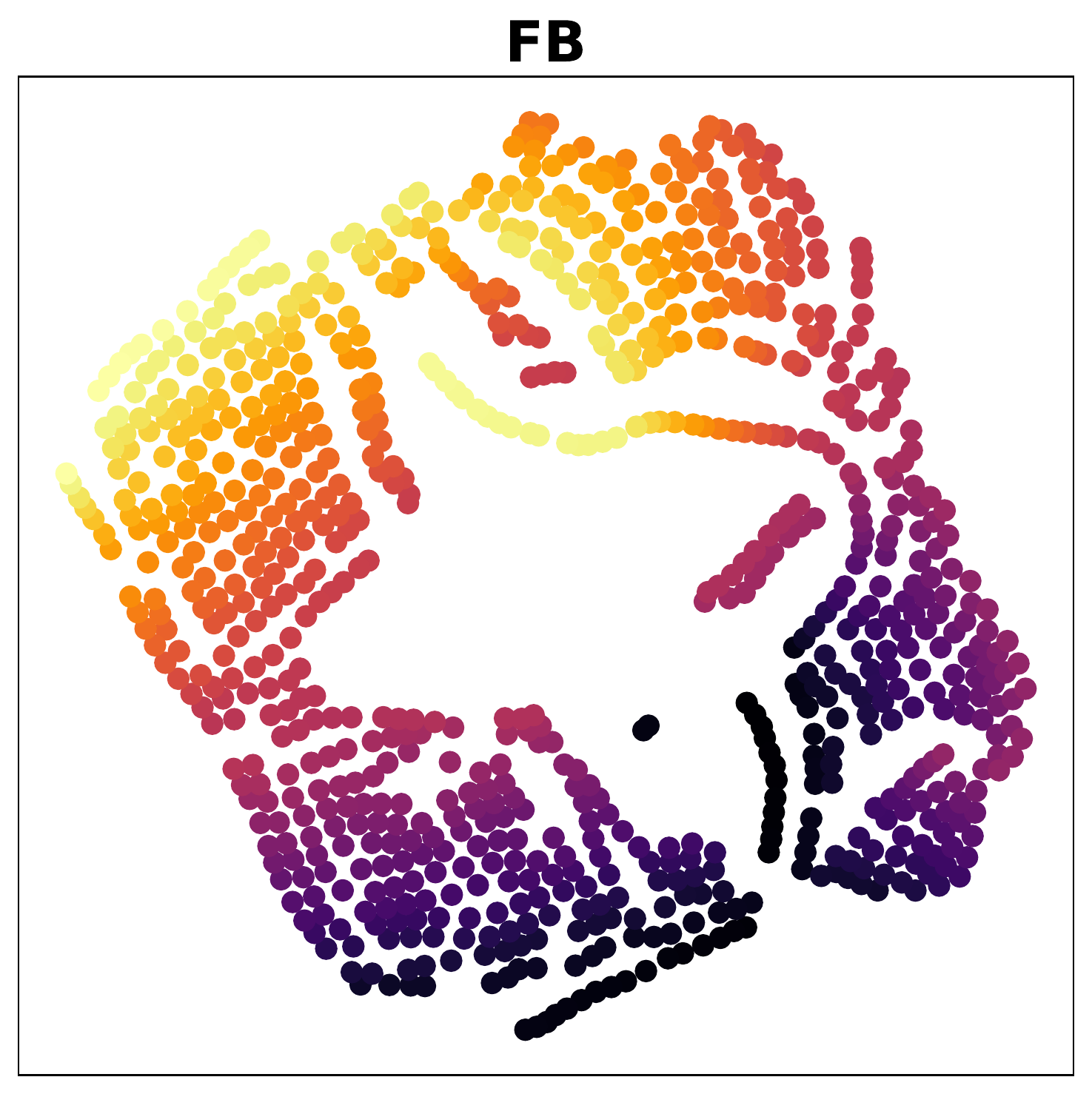}
	\end{center}
	\caption{Visualization of embedding vectors obtained by each method ($\phi$ for \SF and $B$ for \FB) on the maze domain after projecting them in
	two-dimensional space with t-SNE .}
\end{figure}

We visualize the learned state feature embeddings for Maze by projecting them into 2-dimensional space using t-SNE~\cite{van2008visualizing} in Fig.~\ref{fig: tsne}.

\AEnc, \Transition, \FB, \LRAP and \CL all recover a picture of the maze.
 For the other methods the picture is less clear 
 (notably, \Rand gets a near-circle, possibly as an instance of concentration of measure theorems).

Whether t-SNE preserves the shape of the maze does not appear to be correlated to performance:
 \AEnc and \Transition learn nice features but perform poorly,
  while the t-SNE of \lap is not visually interpretable but performance is good.

\newpage
\section{Hyperparameters Sensitivity}
\label{sec:sensitivity}

\begin{figure}[h]
	\label{fig:sensitivity_walker}
	\begin{center}
	\includegraphics[width=0.65\textwidth]{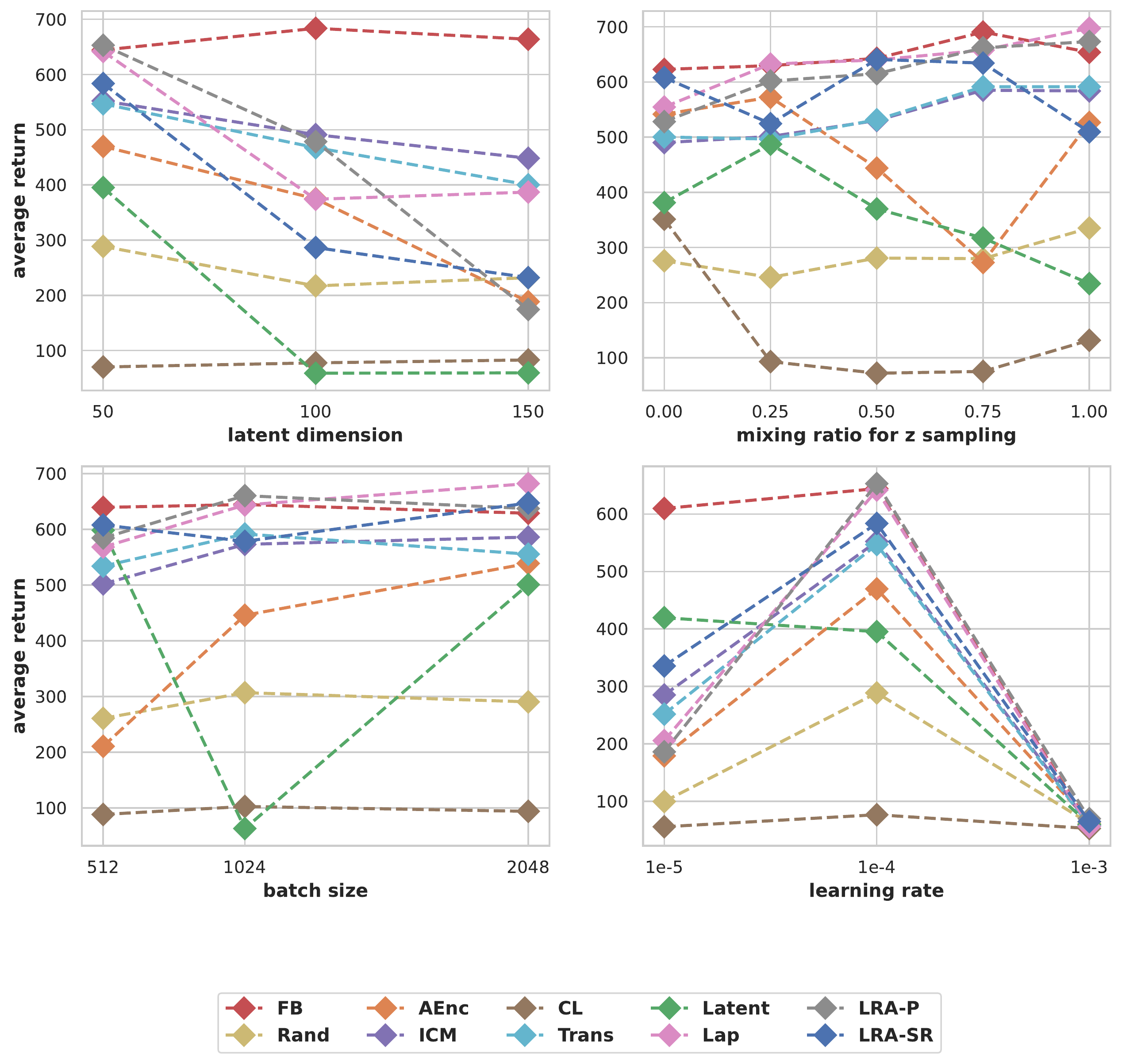}
	\end{center}
	\caption{Return for each method trained in Walker domain on the
	RND replay buffer, for different choices of hyperparameters.
	 Average over Walker tasks and 5 random seeds. 
	}
\end{figure}

\begin{figure}[h!]
	\label{fig:sensitivity_cheetah}
	\begin{center}
	\includegraphics[width=0.65\textwidth]{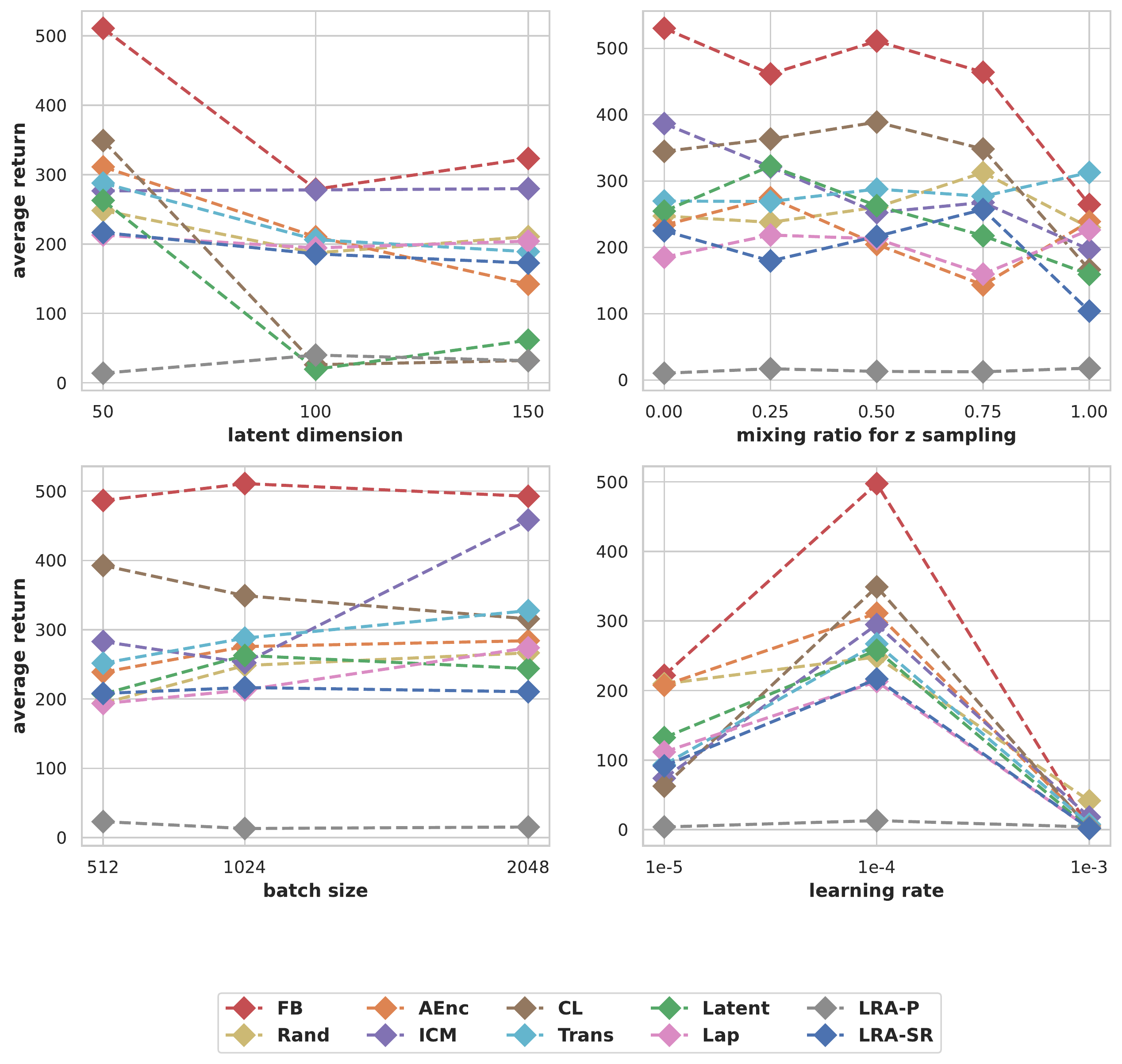}
	
	\end{center}
	\caption{Return for each method trained in Cheetah domain, on the
	RND replay buffer, for different choices of hyperparameters.
	Average over Cheetah tasks and 5 random seeds.}
\end{figure}

\newpage
\section{Goal-Oriented Baselines}
\label{sec:gobaseline}

Goal-oriented methods such as universal value functions (UVFs)
\cite{pmlr-v37-schaul15} learn a $Q$-function $Q(s,a,g)$ indexed by a
goal description $g$. When taking for $g$ the set of all possible target
states $s'\in S$, these methods can learn to reach arbitrary target
states.
For instance, \cite{ma2020universal} use a mixture of universal SFs and
UVFs to learn goal-conditioned agents and policies.

In principle, such goal-oriented methods are not designed to deal with dense
rewards, which are linear combinations of goals $g$. Indeed, the optimal
$Q$-function for a linear combination of goals is not the linear
combination of the optimal $Q$-functions.

Nevertheless, we may still try to see how such linear combinations
perform. The linear combination may be applied at the level of the
$Q$-functions, or at the level of some goal descriptors, as follows.

Here, as an additional baseline for our experiments, we test a slight
modification of the scheme from \cite{ma2020universal}, as suggested by
one reviewer.
We learn two embeddings $(\psi, w)$ using state-reaching tasks. The
$Q$-function for reaching state $s'$ (goal $g=s'$) is modeled as
\begin{equation}
Q(s_t,a_t,s')=\transp{\psi(s_t,a_t,w(s'))}w(s').
\end{equation}
trained for reward $\1_{s_{t+1}=s'}$. A policy network $\pi(s,w(s'))$
outputs the action at $s$ for reaching goal $s'$.
Thus, the training loss is 
\begin{equation}
	\loss(\psi, w)\deq \E_{\substack{ (s_t,a_t,s_{t+1})\sim
	\rho \\ s'\sim \rho}}\left[\transp{\psi(s_t,a_t, w(s'))} w(s')- \1_{\{s_{t+1} = s'\}}-\gamma 
	\transp{\bar\psi(s_{t+1},\pi(s_{t+1}, \bar w(s')),\bar w(s'))}
	\bar w(s')\right]^2
\end{equation}

Similarly to the other baselines, the policy network $\pi(s,w(s'))$ is trained by gradient ascent on the
policy parameters to maximize
\begin{equation}
	\E_{\substack{ s \sim \rho \\ s' \sim \rho}} [Q(s,\pi(s,w(s')),s')]=\E_{\substack{ s \sim \rho \\ s' \sim \rho}} \left[\transp{\psi(s,\pi(s, w(s')),w(s'))} w(s') \right]
\end{equation}

At test time, given a reward $r$, we proceed as in FB and estimate
\begin{equation}
z=\E_{s\sim \rho}[
r(s)w(s)
]
\end{equation}
and use policy $\pi(s,z)$. This amounts to extending from goal-reaching
tasks to dense tasks by linearity on $w$, even though in principle, this
method should only optimize for single-goal rewards.

We use the same architectures for $(\psi, w)$ as for $(F, B)$, with double networks $\psi_1, \psi_2$ and policy smoothing as in~\eqref{eq: policy_loss}.

The sparse reward $\1_{\{s_{t+1} = s'\}}$ could suffer from large variance in continuous state spaces. We mitigate this either by:
 \begin{itemize}
	\item biasing the sampling of $s'$ by setting $s'$ to $s_{t+1}$ half of the time.
	\item replacing the reward by a less sparse one, $\1_{ \{ \|
	s_{t+1}- s'\|_2 \leq \epsilon \}}$.
 \end{itemize}

Table~\ref{tab: uvf} reports normalized scores for each domain, for the
two variants just described,
 trained on the RND replay buffer, averaged over tasks and over 10 random seeds.

The first variant scores $84 \%$ on Maze, $62\%$ on Quadruped tasks, $17\%$ on Walker tasks,
  and $2\%$ on Cheetah tasks. The second variant is worse overall.

So overall, this works well on Maze (as expected, since this is a
goal-oriented problem), moderately well on Quadruped tasks (where most other methods work well), and poorly on the other environments.
 This is expected, as this method is not designed to handle 
 dense combinations of goals.

\begin{table}[h]
	\centering
    \footnotesize	
	\begin{tabular}{lllll}
	\toprule
	\textbf{Reward} & & & \textbf{Domain} \\
	          & Maze & Walker & Cheetah & Quadruped \\
	   \toprule
	   $\1_{ \{ s = s' \}}$ & 83.82 & 17.06 & 1.60 &  61.92 \\
	$ \1_{ \{ \| s- s'\|_2 \leq \epsilon \}}$ & 87.44 & 12.42 &  0.74 & 20.25 \\
	   \bottomrule
	   \end{tabular}

\caption{
	\label{tab: uvf}
Normalized score for each domain of two variants of USF from~\cite{ma2020universal} trained on RND replay buffer, averaged over tasks and 10 random seeds}
\end{table}

\paragraph{Relationship with FB.} The use of universal SFs in \cite{ma2020universal} is quite different
from the use of universal SFs in \cite{barreto2017successor} and
\cite{borsa2018universal}. The latter is mathematically related to FB, as
described at the end of Section~\ref{sec:SR}. The former uses SFs as an
intermediate tool for a goal-oriented model, and is more distantly
related to
FB (notably, it is designed to deal with single goals, not linear
combinations of goals such as dense rewards).

A key difference between FB and goal-oriented methods is the following.
Above, we use $\transp{\psi(s,a,w(g))}w(g)$ as a model of the
optimal $Q$-function for the policy with goal $g$. This is reminiscent of
$\transp{F(s,a,z)}B(g)$ with $z=B(g)$, the value of $z$ used to reach
$g$ in FB.

However, even if we restrict FB to goal-reaching by using $z=B(g)$ only (a
significant restriction), then FB learns $\transp{F(s,a,B(g))} B(g')$ to
model the successor measure, i.e., the number of visits to each goal $g'$
for the policy with goal $g$, starting at $(s,a)$. Thus, FB learns an
object indexed by $(s,a,g,g')$ for all pairs $(g,g')$.

Thus, FB learns more information (it models successor measures instead of
$Q$-functions), and allows for recovering linear combinations of goals in
a principled way.  Meanwhile, even assuming perfect neural network
optimization in goal-reaching methods, there is no reason the goal-oriented policies would
be optimal for arbitrary linear combinations of goals, only for single
goals.

\newpage

\section{Pseudocode of Training Losses}
\label{sec:pseudocode}

Here we provide PyTorch snippets for the key losses, notably the \FB
loss, \SF loss as well as the various feature learning methods for \SF.

\begin{lstlisting}[language=Python, caption=Pytorch code for \FB training loss]
	
def compute_fb_loss(agent, obs, action, next_obs, z, discount): 

	# compute target successor measure

	with torch.no_grad():
		mu = agent.policy_net(next_obs, z)
		next_action = TruncatedNormal(mu=mu, stddev=agent.cfg.stddev, clip=agent.cfg.stddev_clip)
		target_F1, target_F2 = agent.forward_target_net(next_obs, z, next_action)  # batch x z_dim
		target_B = agent.backward_target_net(next_obs)  # batch x z_dim
		target_M1, target_M2 = [torch.einsum('sd, td -> st', target_Fi, target_B) for target_Fi in [F1, F2]] # batch x batch
		target_M = torch.min(target_M1, target_M2)

	# compute the main FB loss

	F1, F2 = agent.forward_net(obs, z, action)
	B = agent.backward_net(next_obs)
	M1, M2 = [torch.einsum('sd, td -> st', Fi, B)  for Fi in [F1, F2]] # batch x batch
	I = torch.eye(*M1.size(), device=M1.device)
	off_diag = ~I.bool()
	fb_offdiag: tp.Any = 0.5 * sum((M - discount * target_M)[off_diag].pow(2).mean() for M in [M1, M2])
	fb_diag: tp.Any = -sum(M.diag().mean() for M in [M1, M2])
	fb_loss = fb_offdiag + fb_diag

	# compute the auxiliary loss

	next_Q1, nextQ2 = [torch.einsum('sd, sd -> s', target_Fi, z) for target_Fi in [target_F1, target_F2]]
	next_Q = torch.min(next_Q1, nextQ2)
	cov = torch.matmul(B.T, B) / B.shape[0]
	inv_cov = torch.linalg.pinv(cov)
	implicit_reward = (torch.matmul(B, inv_cov) * z).sum(dim=1) # batch_size
	target_Q = implicit_reward.detach() + discount * next_Q # batch_size

	Q1, Q2 = [torch.einsum('sd, sd -> s', Fi, z) for Fi in [F1, F2]]
	q_loss = F.mse_loss(Q1, target_Q) + F.mse_loss(Q2, target_Q)
	q_loss /= agent.cfg.z_dim
	fb_loss += q_loss

	# compute Orthonormality losss

	Cov = torch.matmul(B, B.T)
	orth_loss_diag = - 2 * Cov.diag().mean()
	orth_loss_offdiag = Cov[off_diag].pow(2).mean()
	orth_loss = orth_loss_offdiag + orth_loss_diag
	fb_loss += agent.cfg.ortho_coef * orth_loss

	return fb_loss
\end{lstlisting}

\begin{lstlisting}[language=Python, caption=Pytorch code for \SF training loss]
def compute_sf_loss(agent, obs, action, next_obs, z, discount): 

	# compute target q-value
	with torch.no_grad():
		mu = agent.policy_net(next_obs, z)
		next_action = TruncatedNormal(mu=mu, stddev=agent.cfg.stddev, clip=agent.cfg.stddev_clip)
		next_F1, next_F2 = agent.successor_target_net(next_obs, z, next_action)  # batch x z_dim
		target_phi = agent.feature_net(next_goal).detach()  # batch x z_dim
		next_Q1, next_Q2 = [torch.einsum('sd, sd -> s', next_Fi, z) for next_Fi in [next_F1, next_F2]]
		next_Q = torch.min(next_Q1, next_Q2)
		target_Q = torch.einsum('sd, sd -> s', target_phi, z) + discount * next_Q

	F1, F2 = agent.successor_net(obs, z, action)
	Q1, Q2 = [torch.einsum('sd, sd -> s', Fi, z) for Fi in [F1, F2]]
	sf_loss = F.mse_loss(Q1, target_Q) + F.mse_loss(Q2, target_Q)

	return sf_loss
\end{lstlisting}

\begin{lstlisting}[language=Python, caption=Pytorch code for Laplacian Eigenfunctions \lap loss]

def compute_phi_loss(agent, obs, next_obs):

	phi = agent.feature_net(obs)
	next_phi = agent.feature_net(next_obs)
	loss = (phi - next_phi).pow(2).mean()

	# compute Orthonormality losss 

	Cov = torch.matmul(phi, phi.T)
	I = torch.eye(*Cov.size(), device=Cov.device)
	off_diag = ~I.bool()
	orth_loss_diag = - 2 * Cov.diag().mean()
	orth_loss_offdiag = Cov[off_diag].pow(2).mean()
	orth_loss = orth_loss_offdiag + orth_loss_diag

	loss += orth_loss

	return loss
\end{lstlisting}

\begin{lstlisting}[language=Python, caption=Pytorch code for the contrastive \CL loss ]
def compute_phi_loss(agent, obs, future_obs):

	future_phi = agent.feature_net(future_obs)
	mu = agent.mu_net(obs)
	future_phi = F.normalize(future_phi, dim=1)
	mu = F.normalize(mu, dim=1)
	logits = torch.einsum('sd, td-> st', mu, future_phi)  # batch x batch
	I = torch.eye(*logits.size(), device=logits.device)
	off_diag = ~I.bool()
	logits_off_diag = logits[off_diag].reshape(logits.shape[0], logits.shape[0] - 1)
	loss = - logits.diag() + torch.logsumexp(logits_off_diag, dim=1)
	loss = loss.mean()

	return loss
\end{lstlisting}

\begin{lstlisting}[language=Python, caption=Pytorch code of \ICM loss]
def compute_phi_loss(agent, obs, action, next_obs):

	phi = agent.feature_net(obs)
	next_phi = agent.feature_net(next_obs)
	predicted_action = agent.inverse_dynamic_net(torch.cat([phi, next_phi], dim=-1))
	loss = (action - predicted_action).pow(2).mean()

	return loss
\end{lstlisting}

\begin{lstlisting}[language=Python, caption=Pytorch code for \Transition loss]
def compute_phi_loss(agent, obs, action, next_obs):

	phi = agent.feature_net(obs)
	predicted_next_obs = agent.forward_dynamic_net(torch.cat([phi, action], dim=-1))
	loss = (predicted_next_obs - next_obs).pow(2).mean()

	return loss
\end{lstlisting}

\begin{lstlisting}[language=Python, caption=Pytorch code for \Latent loss]
def compute_phi_loss(agent, obs, action, next_obs):
	phi = agent.feature_net(obs)
	with torch.no_grad():
		next_phi = agent.target_feature_net(next_obs)
	predicted_next_obs = agent.forward_dynamic_net(torch.cat([phi, action], dim=-1))
	loss = (predicted_next_obs - next_phi.detach()).pow(2).mean()

	# update target network
	for param, target_param in zip(agent.feature_net.parameters(), agent.target_feature_net.parameters()):
		target_param.data.copy_(tau * param.data +
							(1 - tau) * target_param.data)
	
	return loss
\end{lstlisting}

\begin{lstlisting}[language=Python, caption=Pytorch code for \AEnc loss]
def compute_phi_loss(agent, obs):

	phi = agent.feature_net(obs)
	predicted_obs = agent.decoder(phi)
	loss = (predicted_obs - obs).pow(2).mean()

	return loss
	
\end{lstlisting}

\begin{lstlisting}[language=Python, caption=Pytorch code for \LRAP loss]
	def compute_phi_loss(agent, obs, action, next_obs):
		
		phi = agent.feature_net(next_obs)
		mu = agent.mu_net(torch.cat([obs, action], dim=1))
		P = torch.einsum("sd, td -> st", mu, phi)
		I = torch.eye(*P.size(), device=P.device)
		off_diag = ~I.bool()
		loss = - 2 * P.diag().mean() + P[off_diag].pow(2).mean()
	
		# compute orthonormality loss
		Cov = torch.matmul(phi, phi.T)
		I = torch.eye(*Cov.size(), device=Cov.device)
		off_diag = ~I.bool()
		orth_loss_diag = - 2 * Cov.diag().mean()
		orth_loss_offdiag = Cov[off_diag].pow(2).mean()
		orth_loss = orth_loss_offdiag + orth_loss_diag
		loss += orth_loss
	
		return loss
	\end{lstlisting}

\begin{lstlisting}[language=Python, caption=Pytorch code for \LRASR loss]
def compute_phi_loss(agent, obs, action, next_obs, discount):
	
	phi = agent.feature_net(next_obs)
	mu = agent.mu_net(obs)
	SR = torch.einsum('sd, td -> st', mu, phi)
	with torch.no_grad():
		target_phi = agent.target_feature_net(next_obs)
		target_mu = agent.target_mu_net(next_obs)
		target_SR = torch.einsum("sd, td -> st", target_mu, target_phi)

	I = torch.eye(*SR.size(), device=SR.device)
	off_diag = ~I.bool()
	loss = - 2 * SR.diag().mean() 
		+ (SR - discount * target_SR.detach())[off_diag].pow(2).mean()

	# compute orthonormality loss
	Cov = torch.matmul(phi, phi.T)
	orth_loss_diag = - 2 * Cov.diag().mean()
	orth_loss_offdiag = Cov[off_diag].pow(2).mean()
	orth_loss = orth_loss_offdiag + orth_loss_diag
	loss += orth_loss

	return loss
\end{lstlisting}

\end{document}